\documentclass[10pt]{article}

\usepackage{graphicx}
\usepackage{authblk}
\usepackage{geometry}
\usepackage{amsmath}
\usepackage{amssymb}
\usepackage{algorithm}
\usepackage{bm}
\usepackage[noend]{algpseudocode}

\newcommand{\abs}[1]{\left|#1\right|}

\usepackage{subfig}
\usepackage{wrapfig}
\usepackage[labelfont=bf, justification=justified]{caption}
\usepackage{xcolor}
\usepackage{textcomp}
\usepackage{booktabs}
\usepackage[english]{babel}
\usepackage[latin1]{inputenc}
\usepackage[gen]{eurosym}
\usepackage{hyperref}
\usepackage[hyphenbreaks]{breakurl}
\usepackage{blindtext}
\usepackage{indentfirst}

\usepackage{tikz}
\usetikzlibrary{shapes.geometric,backgrounds, arrows,calc,fit,positioning-plus,node-families}
\tikzstyle{small_block} = [rectangle, rounded corners, minimum width=2cm, text width=2cm, minimum height=1cm,text centered, draw=black]
\tikzstyle{large_block} = [rectangle, rounded corners, minimum width=2cm, text width=4.5cm, minimum height=1cm,text centered, draw=black]
\tikzstyle{arrow} = [thick,->,>=stealth]

\tikzset{
  basic box/.style = {
    shape = rectangle,
    align = center,
    draw  = #1,
    fill  = #1!5,
    rounded corners}
}
\def\connectwithlongtext#1#2#3#4{ % line options, start, end, text
\draw[#1] 
  let
    \p1 = ($(#2)-(#3)$),
    \n1 = {0.6*veclen(\x1,\y1)}
  in  (#2) -- (#3)
   node[midway, below, align=center, text width=\n1, sloped] 
   {#4};
}
\def\connectwithlongtextabove#1#2#3#4{ % line options, start, end, text
\draw[#1] 
  let
    \p1 = ($(#2)-(#3)$),
    \n1 = {0.6*veclen(\x1,\y1)}
  in  (#2) -- (#3)
   node[midway, above, align=center, text width=\n1, sloped] 
   {#4};
}

\begin{document}

\title{Online Structural Health Monitoring by Model Order Reduction and Deep Learning Algorithms}

\author[1]{Luca Rosafalco} %\footnote{Corresponding author}}
\author[1]{Matteo Torzoni}
\author[2]{Andrea Manzoni}
\author[1]{Stefano Mariani}
\author[1]{Alberto Corigliano}
\affil[1]{Dipartimento di Ingegneria Civile e Ambientale,
Politecnico di Milano, Italy\footnote{\{luca.rosafalco,matteo.torzoni,stefano.mariani,
alberto.corigliano\}@polimi.it}}
\affil[2]{MOX, Dipartimento di Matematica, Politecnico di Milano, Italy\footnote{andrea1.manzoni@polimi.it}}

\date{}
\renewcommand\Affilfont{\small}

\maketitle  

\begin{abstract}
Within a structural health monitoring (SHM) framework, we propose a simulation-based classification strategy to move towards online damage localization. The procedure combines parametric Model Order Reduction (MOR) techniques and Fully Convolutional Networks (FCNs) to analyze raw vibration measurements recorded on the monitored structure. First, a dataset of possible structural responses under varying operational conditions is built through a physics-based model, allowing for a finite set of predefined damage scenarios. Then, the dataset is used for the offline training of the FCN. Because of the extremely large number of model evaluations required by the dataset construction, MOR techniques are employed to reduce the computational burden. The trained classifier is shown to be able to map unseen vibrational recordings, \textit{e.g.} collected on-the-fly from sensors placed on the structure, to the actual damage state, thus providing information concerning the presence and also the location of damage. The proposed strategy has been validated by means of two case studies, concerning a 2D portal frame and a 3D portal frame railway bridge;  MOR  techniques have allowed us to respectively speed up the analyses about $30$ and $420$ times. For both the case studies, after training the classifier has attained an accuracy greater than $85\%$.
\\[18pt]
\noindent {\em Keywords\/}: structural health monitoring, deep learning, reduced order models, fully convolutional networks, damage localization.
\end{abstract}
%\newpage
%%%%%%%%%%%%%%%%%%%%%%%%%%%%%%%%%%%%%%%%%%%%%%%%%%%%%

\section{Introduction}
Physics-based models are derived from first principles expressing the laws of nature, and from accurate and efficient numerical methods for their approximation, such as Finite Elements (FE). Dual to the physics-based paradigm is the data-driven paradigm, according to which laws and patterns governing complex data systems are unveiled by using statistical tools, among which Machine Learning (ML) stands out. Physics-based and data-driven paradigms are not mutually exclusive, even if in the past they were mostly applied separately. Indeed, when dealing with data assimilation, data-driven modelling might be exploited to establish a link between physics-based models and experimental data \cite{art:ML_multi_scale_modelling, art:Virtual_twins}. Structural Health Monitoring (SHM) is one of the research fields in which this combination is more promising \cite{book:Farrar_Worden}. \newpage

SHM aims at detecting, locating and quantifying the inception and propagation of damage in a structure by analysing data acquired through pervasive sensor networks \cite{art:Farrar_Worden_RSTA}. As said, the above paradigms yield two frameworks in SHM: the {\em model-based} and the {\em data-driven} approaches. Model-based approaches deal with the monitoring of civil infrastructures by relying on the update of a numerical model (\textit{e.g.} through Kalman filters \cite{art:Azam_Mariani_1,art:Azam_Mariani_2,art:Azam_3} or optimization procedures \cite{art:Rizzi}). They enable a mechanical intuition of the structural degradation process and the forecasting of the system evolution (prognosis). On the other hand, model-based approaches hardly manage to deal with the great amount of (noisy) data acquired through sensor networks. For this reason, data-driven approaches have become more and more widespread \cite{art:Pozo, art:FW_env_cond}.

Among data-driven approaches, we can distinguish supervised \cite{art:Alireza_2} and unsupervised  \cite{art:Bull,art:Alireza_1,art:Alireza_3, art:Alireza_4,art:Rafei} methods. Supervised methods employ labeled data referring both to the undamaged condition, assumed as baseline, and to the damage scenarios possibly affecting the structure. Unsupervised methods rely only on unlabeled data collected from the undamaged condition. When coming to damage localization and quantification, supervised methods are more powerful, although they feature an obvious drawback: experimental data referring to possible damage conditions of the structure are, a priori, not available.

To cope with this issue, model-based and data-driven approaches are combined by introducing a physics-based model to simulate the effect of damage on the dynamic response of the structure. Specifically, SHM is approached as a classification problem \cite{art:Farrar_1} and different damage scenarios, featuring a range of damage classes, are numerically simulated and used as dataset to train a ML-based classifier. This approach is called Simulation-Based Classification (SBC) \cite{art:Taddei_Patera,art:Bigoni_1,art:Rosafalco_1}. To be effective, the construction of the dataset should take into account the effect of varying operational and environmental conditions. In this respect, Model Order Reduction (MOR) techniques for parametrized systems, such as the Reduced Basis (RB) method \cite{book:RB}, can be exploited to speed up the generation of the dataset, thus replacing the solution of a high-fidelity, Full Order Model (FOM) with a cheaper, yet accurate, approximation obtained through a Reduced Order Model (ROM). The RB method provides a low-dimensional approximation of the set of solutions of the FOM, within a prescribed parameters range. It combines a handful of FOM solutions (or snapshots) computed for a set of parameter values to generate a low dimensional space exploiting, \textit{e.g.}, the Proper Orthogonal Decomposition (POD). 
%By this approach, a very large dynamical system is replaced by a much smaller reduced problem -- whose dimension is related to the number of selected basis functions -- relying on a suitable (Galerkin) projection approach.
After the ROM has been built (offline),  the associated approximation for the input parameters within the range of interest can be obtained (online) in an almost inexpensive way.

In this paper, a ML-based classifier has been designed by using concepts typical of Deep Learning (DL), a branch of ML: the processing of raw data and the classification task are enveloped through the minimization of a single loss function \cite{art:DL} by exploiting a suitable Neural Network (NN) architecture \cite{proc:time_series_base,art:SHMnet}, called Fully Convolutional Network (FCN). This latter is trained to assign labels (or classes) to input data given by vibrational measurements of the monitored structure. The strength of this architecture, already successfully applied in \cite{proc:Rosafalco_1,art:Rosafalco_1}, relies on the repeated convolution operations. It has been proven to be much less resource-demanding than the feed-forward NN one; as it is also tailored to detect the correlation within a time series and also across different time series, it looks appropriate for SHM purposes as correlations are actually induced by the vibrations of the structure excited by the external loads. Alternative architectures have been recently proposed to approximate the solution of partial differential equations, by adopting the energy of the system as a "\emph{natural loss function for a machine learning method}", see \cite{art:Rabczuk20}.

While in the past MOR techniques \cite{art:Taddei_Patera,art:Bigoni_1} and DL \cite{art:SHMnet} have been explored separately, here their use is advantageously combined to exploit the physical knowledge of the system, to minimize the efforts and time required by data processing, but also to efficiently cope with uncertainties (due to, \textit{e.g.}, the operational conditions of the structure). Moreover, by constructing the training dataset and by training the classifier offline, we allow for an extremely efficient online monitoring of the structure, given that the trained NN only needs to apply a fixed sequence of linear mappings and nonlinear (activation) functions to the incoming vibration recordings to obtain the classification outcome.

The assessment of the proposed procedure has been done through two numerical case studies addressing several operative issues, from the influence of the ROM reconstruction error on the classification accuracy, to the effect of different measurement noise levels on the procedure performance. Experimental data, still not investigated in this paper, will be addressed in future works.  

The reminder of the paper is organized as follows. In Sec. \ref{sec:methodology}, we detail our methodology, highlighting the stochastic treatment of the operational and damage conditions faced by the structure, the construction of the (reduced order) physics-based model, and the setting of the FCN. In Secs. \ref{sec:portal_frame} and \ref{sec:bridge} two numerical case studies are discussed, dealing with a two-dimensional portal frame and an integral concrete portal frame railway bridge, respectively. Conclusions, remarks and future developments are finally discussed in Sec. \ref{sec:conclusion}.

\section{Methodology}
\label{sec:methodology}

In the following, we present our methodology. Specifically: in Sec. \ref{sec:dataset_def} we detail the content of the dataset $\mathbf{D}$ used to train and validate the classifier $\mathcal{G}$; in Sec. \ref{sec:str_model_def} we focus on the high-fidelity FOM of the structure; in Sec. \ref{sec:ROM_construction} we show how a ROM is next obtained through a MOR technique; finally, in Sec. \ref{sec:FCN}, we discuss the peculiarities of the employed NN architecture.

\subsection{Dataset definition}
\label{sec:dataset_def}

A set of $N_0$ sensors is exploited to track the vibrational response of the monitored structure \cite{proc:Capellari_1,art:Capellari_3,art:Bigoni_2}. Within the observation interval $(0,T)$, each sensor is assumed to provide  $L$ measurements of the local displacement with fixed sampling rate, all collected in the vector $\mathbf{u}_n \in \mathbb{R}^{L}$ ($n=1,\ldots,N_0$); the same holds if accelerations $\mathbf{\ddot{u}}_n$ are sensed. Here, we consider only displacement measurements, even if (see Sec. \ref{sec:portal_frame}-\ref{sec:bridge}) our methodology can cope with acceleration recordings too. Obviously, both the duration $T$ of the time interval and the sampling rate must be chosen according to the structural frequencies to be handled and to the sensor characteristics.

We call \textit{instance} a set of recordings $\mathbf{U}_i=\left[\mathbf{u}^i_1, \ldots, \mathbf{u}^i_{N_0} \right] \in \mathbb{R}^{L \times N_0}$, $i=1,...,I$, related to the same time interval. The dataset $\mathbf{D}$, used to train and validate the classifier $\mathcal{G}$, is constructed by collecting $I$ instances
\begin{equation}
\mathbf{D}=\lbrace\left(\mathbf{U}_1,g_1\right),\ldots,\left(\mathbf{U}_I,g_I\right) \rbrace,
\label{eq:dataset_def}
\end{equation}
where: $I=I_{tr}+I_{val}$; $g_i$ labels the damage state (if any) in the structure.

The goal of the classifier $\mathcal{G}$ is to build the underlying mapping between $\mathbf{U}_i$ and $g_i$ \cite{book:Bishop}. During the training phase (see Sec. \ref{sec:FCN}), $I_{tr}$ instances are employed by the classifier to {\em learn} how to model this mapping \cite{art:ML_def}, while $I_{val}$ instances  are later used to validate the learning process by verifying that the training data are not simply {\em memorized}. Once trained, the classifier should be able to map an unseen instance $\mathbf{U}_i$, uniquely defined by the damage state $g_i=0,\ldots,G$  ($g_i=0$ refers to the undamaged condition) and by the operational conditions, into the correct damage class $g_i$. Only a finite number $G$ of damage states is allowed for, coherently with the classification framework within which we approach the SHM problem. Both the possible damage states and operational conditions must be determined through a preliminary study, by evaluating the mechanical behavior of the structure.

The vector of parameters $\boldsymbol{\eta}_i \in \text{H} \subset \mathbb{R}^Q$ (\textit{e.g.} acting as load multipliers) is used to describe the operational conditions relevant to the $i-$th instance. We assume that $\boldsymbol{\eta}_i$ does not vary in $\left(0,T\right)$; in other words, a  time independent set of parameters is associated to each instance. Even the damage state $g_i$ is assumed to be frozen within the time interval of interest, coherently with the damage growth typically faced by a civil structure \cite{art:Azam_Mariani_1}. The $q$-th parameter $\eta^i_q$ ($q=1,\ldots,Q$) is sampled from a continuous probability density function (pdf) $\mathcal{P}_q$, preliminarily set (see, \textit{e.g.}, \cite{art:Sudret_1,art:Sudret_2}, where the authors talk about {\em probabilistic input data}). Similarly, the occurrence of the considered damage state is sampled from a discrete pdf $\mathcal{P}_g$. Different sampling strategies, both random (\textit{e.g.} Latin Hypercube \cite{art:LHS}) and quasi-random (\textit{e.g.} Sobol' sequences \cite{art:Sobol_sequence}) can be adopted to explore the parametric space defined by the combination between the parameters governing the operational conditions and the damage states. In this work, a Latin Hypercube Sampling (LHS) has been adopted, as it provided a good compromise between randomness and  coverage of the parameter domain \cite{art:Ubbiali}. The instance $\mathbf{U}_i$, corresponding to the sampled $\lbrace g,\boldsymbol{\eta}\rbrace_i$, is simulated through a numerical model.

To slightly simplify the notation, in the following the index $i$ relevant to the instance will be dropped. Anyhow, it must be remembered that the discussion refers to the $i-$th instance only, and computations must be therefore repeated $I$ times at varying loading and operational conditions.

\subsection{Full Order Model construction}
\label{sec:str_model_def}

Before coming to the details related to the FOM setting, few hypotheses concerning the response of a civil infrastructure under varying operational conditions are discussed. The strains and displacements are assumed to be small and, if not specified otherwise, damping effects are disregarded; see e.g.  \cite{art:Azam_Mariani_1,art:Corigliano_Mariani} for some results regarding the  relevance of damping in the identification of continuously excited structures. Damage is modeled as a localized reduction in stiffness, temporarily frozen in time. Despite the simplicity of this last assumption, which rests on a time scale separation between damage evolution and health assessment, many engineering problems can be tackled as discussed, \textit{e.g.}, in \cite{art:Teughels}.

To describe the behavior of the structure, we rely upon linear elasto-dynamics. By space discretizing the governing equation through Finite Elements (FEs), we obtain the following semi-discretized problem: 
%By introducing a finite element FE space $V_h\left(x\right) \subset V \left(x \right)$ of dimension $M=\text{dim}\left(V_h\left(x\right)\right)$, associated with a fine triangulation of the domain $\Omega$, 
\begin{equation}
\left\{
\begin{array}{ll}
\mathbf{M}\ddot{\mathbf{v}} + \mathbf{K}\left(g\right)\mathbf{v}=\mathbf{f}\left(\boldsymbol{\eta}\right), & \quad t \in (0,T) \\
\mathbf{v}(0) = \mathbf{v}_0\left(\boldsymbol{\eta}\right) & \\
\dot{\mathbf{v}}(0)= \dot{\mathbf{v}}_0\left(\boldsymbol{\eta}\right)   &
\end{array}
\right.
\label{eq:semi-discretized_governing}
\end{equation}
where: $\mathbf{v}=\mathbf{v}\left(t\right) \in \mathbb{R}^M$ is the displacement vector, while $\dot{\mathbf{v}}$ and $\ddot{\mathbf{v}}$ are the corresponding velocity and acceleration vectors; $\mathbf{M} \in \mathbb{R}^{M \times M}$ is the mass matrix; $\mathbf{K}\left(g\right) \in \mathbb{R}^{M \times M}$ is the elastic stiffness matrix; $\mathbf{f}\left(\boldsymbol{\eta}\right)$ is the vector collecting the external loadings. Here $M$ denotes the total number of degrees of freedom (dofs) of the FE space. 

A time discretization of the monitoring window $(0,T)$ is next determined on the basis of the sensor sampling rate. By adopting a suitable time integration scheme (like the generalised-$\alpha$ method \cite{art:alpha_method}), the displacements $\mathbf{v}_l$ (with $l=0,\ldots,L$) related to the sampled $\lbrace g,\boldsymbol{\eta}\rbrace$ are provided. Displacements $\mathbf{v}_l$ are then all collected in $\mathbf{V}=\left[\mathbf{v}_1,\ldots,\mathbf{v}_L\right] \in \mathbb{R}^{M \times L}$ and, through a Boolean matrix $\mathbf{T}\in \mathbb{R}^{N_0 \times M}$, whose $\left(n,m\right)$-th entry is equal to $1$ if and only if the position and orientation in space of the $n$-th sensor and of the $m$-th dof coincide, the corresponding instance
\begin{equation}
\mathbf{U}=\left(\mathbf{T}\mathbf{V}\right)^T,
\label{eq:pick_sensors}
\end{equation}
is obtained. 

\subsection{Reduced Order Model construction}
\label{sec:ROM_construction}

To get a high quality dataset $\mathbf{D}$, the number of required instances may be extremely high. By increasing $I$, we enhance the performance of $\mathcal{G}$ even if, beyond a certain threshold, the gain becomes marginal. To speed up the dataset construction, we propose to adopt parametric MOR techniques \cite{book:RB, proc:Rosafalco_2} as detailed below.

The FOM solution is approximated as $\mathbf{v} \approx \mathbf{W}\mathbf{v}_{R}$, $\mathbf{v}_R \in \mathbb{R}^W$, through a linear combination of $W \ll M$ basis functions $\mathbf{w}_w \in \mathbb{R}^{M}$ (with $w=1,\ldots,W$) collected into the matrix \mbox{$\mathbf{W}=\left[\mathbf{w}_1,\ldots,\mathbf{w}_W \right] \in \mathbb{R}^{M \times W}$}. To determine the ROM solution $\mathbf{v}_{R}$, we enforce the orthogonality between the residual $\mathbf{v}-\mathbf{v}_{R}$ and the subspace $\text{span}\lbrace \mathbf{w}_1,\ldots, \mathbf{w}_W \rbrace$; in other words, we perform a Galerkin projection onto the subspace $\text{span}\lbrace \mathbf{w}_1,\ldots, \mathbf{w}_W \rbrace$. The governing equation of the ROM then becomes
\begin{equation}
\left\{
\begin{array}{ll}
\mathbf{M}_{R}\ddot{\mathbf{v}}_{R} + \mathbf{K}_{R}\left(g\right)\mathbf{v}_{R}=\mathbf{f}_{R}\left(\boldsymbol{\eta}\right) , & \quad t \in (0,T) \\
\mathbf{v}_R(0) = \mathbf{W}^T\mathbf{v}_0\left(\boldsymbol{\eta}\right) & \\
\dot{\mathbf{v}}_R(0)= \mathbf{W}^T\dot{\mathbf{v}}_0\left(\boldsymbol{\eta}\right)   &
\end{array}
\right.
\label{eq:red_semi_discr_gov}
\end{equation}
where
\[
%\begin{equation*}
%\begin{array}
%\]
%\[
\mathbf{M}_{R}   = \mathbf{W}^T \mathbf{M}   \mathbf{W}, \qquad  
%\[
\mathbf{K}_{R} \left( g \right)= \mathbf{W}^T \mathbf{K} \left( g \right) \mathbf{W}, \qquad
%\[
\mathbf{f}_{R} = \mathbf{W}^T \mathbf{f}.
\]

Eq. \eqref{eq:red_semi_discr_gov} is integrated in time to obtain $\mathbf{V}_{R}=\left[\mathbf{v}_{R1},\ldots,\mathbf{v}_{RL}\right]\in \mathbb{R}^{W \times L}$, and then projected back onto the original FOM space to obtain the whole solution $\mathbf{V} \approx \mathbf{W} \mathbf{V}_{R}$.

If the FOM arrays in Eq. \eqref{eq:semi-discretized_governing} exhibit an affine parametric dependency, it is possible to write
\[
%\begin{equation}
%\begin{array}
%\]
%\[
\mathbf{K} \left( g \right)= \sum_{p=1}^{P_k} \psi_p \left(g \right) \mathbf{K}_p, \qquad
%\[
\mathbf{f}\left( \boldsymbol{\eta} \right) = \sum_{p'=1}^{P_f} \psi'_{p'}\left(\boldsymbol{\eta}\right)\mathbf{f}_{p'},
\]
where $\psi_p\left(g\right): \lbrace 0,\ldots, G \rbrace \rightarrow \mathbb{R}$ (with $p=1,\ldots,P_k$) and $\psi'_{p'}\left(\boldsymbol{\eta}\right): \text{H} \rightarrow \mathbb{R}$ (with \mbox{$p'=1,\ldots,P_f$}) are two sets of scalar functions; $\mathbf{K}_p \in \mathbb{R}^{M \times M}$ (with $p=1,\ldots,P_k$) is a set of $g$-independent matrices; $\mathbf{f}_{p'} \in \mathbb{R}^{M}$ (with $p'=1,\ldots,P_f$) is a set of $\boldsymbol{\eta}$-independent vectors.
For the case at hand, affine parametric dependency is built-in in the formulation of the FOM, since parameters governing both the operational conditions and the damage states are taken as constant in time, and piecewise constants over different spatial subdomains, thus yielding the possibility to factor them out of the assembled matrices. Nonaffine parametric dependency, on the other hand, would require the use of suitable {\em hyper-reduction} techniques, to restore an approximate affine parametric dependency, see, e.g. \cite{book:RB}.

Under the assumption of affine parametric dependency, assembling the ROM arrays in Eq. \eqref{eq:red_semi_discr_gov} can be made independent of the FOM dimension $M$ for any $\lbrace g, \boldsymbol{\eta} \rbrace$. Indeed, we have
\begin{equation*}
\begin{array}{l}
\mathbf{K}_R \left( g \right)= \sum_{p=1}^{P_k} \psi_p \left(g \right) \mathbf{W}^T \mathbf{K}_p \mathbf{W}=\sum_{p=1}^{P_k} \psi_p \left(g \right) \mathbf{K}^p_R, \\[0.25cm]
\mathbf{f}_R\left( \boldsymbol{\eta} \right) = \sum_{p'=1}^{P_f} \psi'_{p'}\left(\boldsymbol{\eta}\right) \mathbf{W}^T \mathbf{f}_{p'}= \sum_{p'=1}^{P_f} \psi'_{p'}\left(\boldsymbol{\eta}\right) \mathbf{f}_R^{p'},
\end{array}
\end{equation*}
given that
\[
%\begin{equation*}
%\begin{array}
%\]
%\[
\mathbf{K}^p_R= \mathbf{W}^T \mathbf{K}_p \mathbf{W} \in \mathbb{R}^{W \times W}, \qquad
%\[
\mathbf{f}_R^{p'} = \mathbf{W}^T \mathbf{f}_{p'} \in \mathbb{R}^W,
\]
can be computed and stored once for all, with $\mathbf{K}_R \left( g \right)$, $\mathbf{f}_R\left( \boldsymbol{\eta} \right)$ constructed without repeating the costly assembling operations required by the FE model. 

To set $\mathbf{W}$, a Proper Orthogonal Decomposition (POD) is performed on the matrix \mbox{$\boldsymbol{\mathcal{S}}=\left[\mathbf{v}_1,\ldots,\mathbf{v}_S\right] \in \mathbb{R}^{M \times S}$} collecting $S$ snapshots of the FOM. The collected snapshots must embody both the dependence on $\lbrace g,\boldsymbol{\eta}\rbrace$ and on $t$; details on basis construction in elasto-dynamics are reported in Algorithm \ref{al:POD_param_elastodyn}. The total number of snapshots collected is $S= Y \times X$, where $Y \geq 1+G$ is the number of samples of $\lbrace g,\boldsymbol{\eta}\rbrace$, so that  each damage state is sampled at least once, and  $X$ is the number of samples in time. In order to guide the choice of $Y$ and $X$, we refer, \textit{e.g.} to  \cite{art:Azam_Mariani_2}: here, we only remark that the constraint $X \leq L$ must be always satisfied. Indeed, to speedup the ROM construction there is often the possibility to restrict the snapshot collection to a small portion of the time window of interest, given that enough information on the dynamic evolution of the system is still captured.

POD of matrix $\boldsymbol{\mathcal{S}} \in \mathbb{R}^{M \times S}$ is performed via a singular value decomposition according to
\begin{equation}
\boldsymbol{\mathcal{S}}=\mathbf{P}\mathbf{\Sigma}\mathbf{Z}^T
\end{equation}
where: $\mathbf{P}=\left[\mathbf{w}_1,\ldots, \mathbf{w}_M \right]\in \mathbb{R}^{M \times M}$ is an orthogonal matrix whose columns are the  left singular vectors of $\boldsymbol{\mathcal{S}}$; $\mathbf{Z}=\left[\mathbf{z}_1,\ldots, \mathbf{z}_S \right] \in \mathbb{R}^{S \times S}$ is an orthogonal matrix whose columns are the  right singular vectors of $\boldsymbol{\mathcal{S}}$; $\mathbf{\Sigma} \in \mathbb{R}^{M \times S}$ collects the singular values of $\boldsymbol{\mathcal{S}}$. When $M>S$,
\begin{equation*}
\mathbf{\Sigma}=\begin{bmatrix}
\sigma_1 & 0 & \ldots & 0 \\
0 & \sigma_2 & \ldots & 0 \\
\vdots & \vdots & \ddots &\vdots \\
0 & 0 & \ldots & \sigma_r \\
\vdots & \vdots & \vdots & \vdots \\
0 & 0 & \ldots & 0
\end{bmatrix} ,
\end{equation*}
where $\sigma_1 \geq \sigma_2 \geq \ldots \geq \sigma_r \geq 0$ and $r$ is the rank of $\boldsymbol{\mathcal{S}}$.

The POD bases $\mathbf{W}=\left[\mathbf{w}_1,\ldots,\mathbf{w}_W\right]$ are then obtained by retaining the first $W\leq S$ left singular vectors in $\mathbf{P}$. Among all  possible approximations of $\boldsymbol{\mathcal{S}}$ of rank $W$, $\mathbf{W}$ captures as much energy of $\boldsymbol{\mathcal{S}}$ as possible \cite{book:RB,art:Kerschen2002}. The normalized reconstruction error $\varepsilon$ obtained by retaining the first $W$ modes  can be related to the discarded singular values  as 
\begin{equation}
\varepsilon = \sqrt{\frac{\sum_{s=W+1}^S \sigma_s^2}{\sum_{s=1}^S \sigma_s^2}},
\label{eq:reconstruction_error}
\end{equation}
By prescribing a tolerance $\varepsilon_{tol}$, such that $\varepsilon < \varepsilon_{tol}$, we can automatically set the dimension $W$ of the ROM.

\begin{algorithm}[t!]
\begin{algorithmic}[1]
\State Sample $\boldsymbol{\eta}_1$ via Latin Hypercube Sampling
\State Collect $\boldsymbol{\mathcal{S}}_1=\left[\mathbf{v}\left(g_1, \boldsymbol{\eta}_1, t_1 \right) | \ldots | \mathbf{v}\left(g_1,\boldsymbol{\eta}_1,t_{X} \right) \right]$
\State $\mathbf{W} = \mathbf{POD}\left(\boldsymbol{\mathcal{S}}_1\right)$
\State FOR $\tau=2,\ldots,Y$ DO
\State $\qquad$ Sample $\boldsymbol{\eta}_{\tau}$ via Latin Hypercube Sampling
\State $\qquad$ Collect $\boldsymbol{\mathcal{S}}_{\tau}=\left[\mathbf{v}\left(g_{\tau},\boldsymbol{\eta}_{\tau},t_1\right) | \ldots | \mathbf{v}\left(g_{\tau},\boldsymbol{\eta}_{\tau}, t_{X}\right) \right]$
\State $\qquad \mathbf{W}_{\tau} = \mathbf{POD}\left(\boldsymbol{\mathcal{S}}_{\tau}\right)$
\State $\qquad \boldsymbol{\mathcal{S}}=\left[\mathbf{W} | \mathbf{W}_{\tau} \right]$
\State $\qquad \mathbf{W} = \mathbf{POD} \left(\boldsymbol{\mathcal{S}} \right)$
\State END FOR
\end{algorithmic}
\caption{POD bases determination in elasto-dynamics, as proposed in \cite{proc:Rosafalco_1}.}
\label{al:POD_param_elastodyn}
\end{algorithm}

\subsection{Fully Convolutional Networks}
\label{sec:FCN}

Once $\mathbf{D}$ has been constructed according to Eq. \eqref{eq:dataset_def}, $I_{tr}$ instances are used to train the classifier $\mathcal{G}$. First, $\mathcal{G}$ performs a series of (nonlinear) mappings through basis functions ruled by tunable weights $\mathbf{\Omega}$ \cite{book:Bishop}, whose overall effect is to make the damage classes linearly separable \cite{book:Haykin}. In this way, a linear mapping, ruled by a weight matrix $\mathbf{\Theta}$ and followed by a softmax function, proves sufficient to perform the classification task. In details, denoting by  $\boldsymbol{\vartheta}  =({\vartheta}_0, \ldots, {\vartheta}_G)^T\in \mathbb{R}^{G+1}$ the outcome of the linear mapping, the softmax function computes a vector $\boldsymbol{\varrho} =(\varrho_0, \ldots, \varrho_G)^T\in \left[0,1\right]^{G+1}$, with
\begin{equation}
\varrho_g = \frac{\text{e}^{{\vartheta}_g}}{\sum_{j=0}^G \text{e}^{{\vartheta}_j}} \quad g=0,\ldots,G ,
\end{equation}
denoting the probability by which the input $\mathbf{U}$ is assigned to the $g$-th damage class. The classification task is then performed by selecting the class $g$ corresponding to the highest $\varrho_g$ value.

During the training stage, the classification error is quantified by a loss function $C$, assumed to be the cross entropy
\begin{equation}
C\left({\scriptstyle \boldsymbol{\mathcal{Z}}},\boldsymbol{\varrho}\right)=-\sum_{g=0}^G {\scriptstyle \mathcal{Z}}_g \text{log}\left(\varrho_g\right),
\end{equation}
where ${\scriptstyle \mathcal{Z}}_g \in \lbrace 0, 1 \rbrace$ is the confidence with which the $g$-th damage class should be assigned to $\mathbf{U}$; ${\scriptstyle \boldsymbol{\mathcal{Z}}}$ is the vector that collects the confidence values. The training consists in minimizing $C$ by tuning $\mathbf{\Omega}$ and $\mathbf{\Theta}$ through an iterative procedure; Adam \cite{art:Adam}, a first-order stochastic gradient descend algorithm, is employed with this aim. At each iteration, a certain number of instances, called mini-batch, are  analyzed simultaneously. In the forthcoming example sections, we have employed mini-batches containing $B=16$ 	instances, but in the following we assume that each mini-batch counts just one of them, in order to simplify the notation. At the end of the training, the $I_{tr}$ instances are processed a number of times named {\em epochs}. 

\begin{figure}[h!]
\includegraphics[width=1\textwidth]{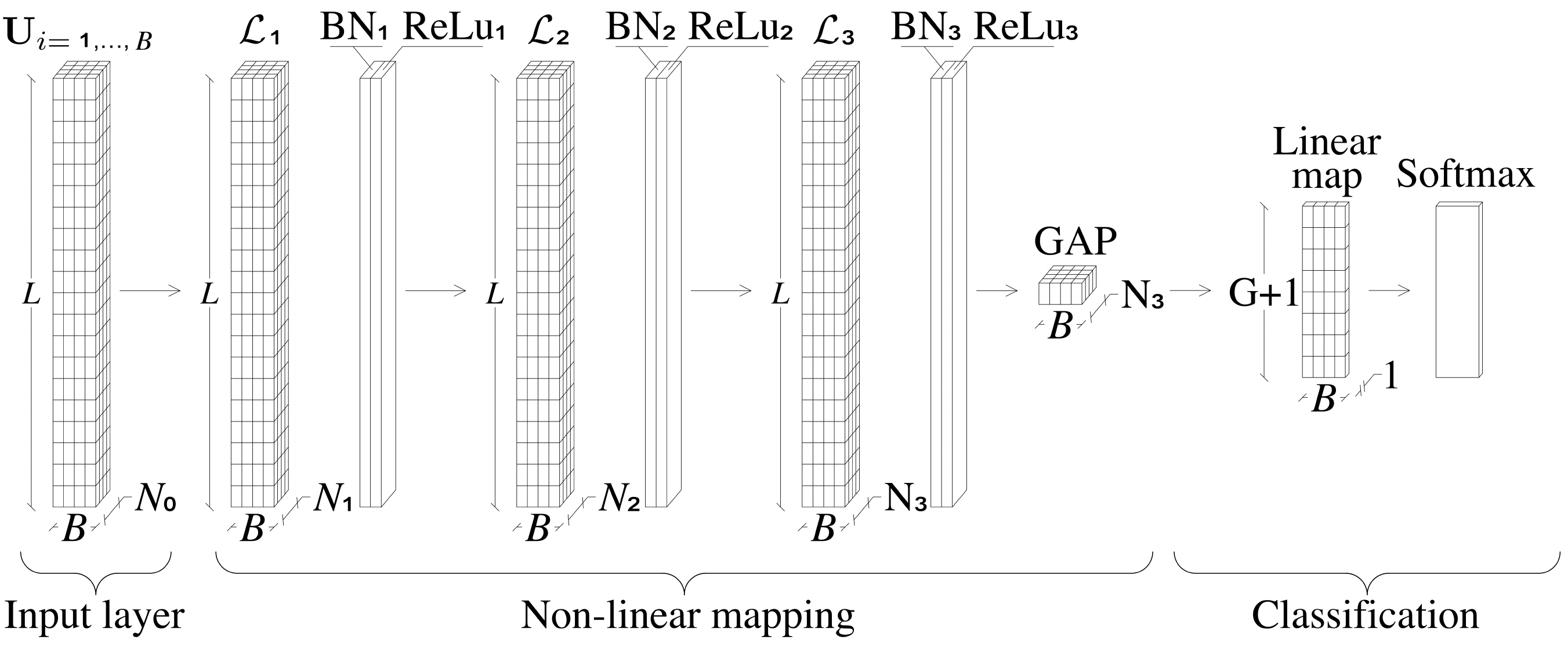}
\caption{{FCN architecture.}\label{fig:FCN}}
\end{figure}

The nonlinear mappings ruled by $\mathbf{\Omega}$ and the final classification ruled by $\mathbf{\Theta}$ have been performed employing a NN called Fully Convolutional Network (FCN), resembling the one proposed in \cite{proc:time_series_base}. The chosen NN architecture, depicted in Fig. \ref{fig:FCN}, can analyze multivariate time series, so that each channel $\mathbf{u}_n$ is not treated separately, and correlations between different channels can be exploited to improve the classifier effectiveness. The functioning of each block depicted in Fig. \ref{fig:FCN} is detailed in the following.

A NN is a computational algorithm that assembles basic units called neurons \cite{book:Haykin}. Each neuron computes a scalar output \footnotesize $\mathcal{Y}$ \normalsize by operating first a linear mapping of the input \footnotesize $\boldsymbol{\mathcal{U}}$ \normalsize $\in \mathbb{R}^{\lambda}$ (which reads $\mathbf{u}_n$ for the first layer of the NN, see below) through a weight vector $\boldsymbol{\omega}\in \mathbb{R}^{\lambda}$ and a bias term $\beta$, and by using next a nonlinear activation function $\zeta$ according to
\begin{equation*}
{\scriptstyle \mathcal{Y}}= \zeta\left(\boldsymbol{\omega} \cdot {\scriptstyle \boldsymbol{\mathcal{U}}} + \beta \right) .
\end{equation*}
Typical choices for the activation function are the hyperbolic tangent or the Rectified Linear Unit (ReLU). Thanks to the use of activation functions, NNs can perform tasks that go beyond the ones accomplished by a sequence of linear projections \cite{book:Bishop}.  If $\lambda$ neurons simultaneously deal with \footnotesize $\boldsymbol{\mathcal{U}}$ \normalsize, the output of the transformation is a vector \footnotesize $\boldsymbol{\mathcal{Y}}$ \normalsize$\in \mathbb{R}^{\lambda}$: this collection of $\lambda$ neurons is called layer. In a DL framework, more layers are usually stacked in order to form a deep architecture. Several layer typologies exist, differing in terms of the way in which the neurons are connected to the inputs: the layer typology described so far is called fully-connected.

In our NN architecture, convolutional layers are exploited. Convolutional layers are widespread in computer vision \cite{art:CNN_review_image} and they are gaining attention in signal processing too \cite{art:multivariate_LSTM_FCN,art:Wavenet}. In particular, three convolutional layers $\mathcal{L}_k$ (with $k=1,2,3$) have been employed, each one together with a Batch Normalization (BN) $\mathcal{B}_k$ and a ReLU activation $\mathcal{R}_k$. We call convolutional block the computational unit made up by $\mathcal{L}_k$, $\mathcal{B}_k$ and $\mathcal{R}_k$. By putting three convolutional blocks in sequence and by applying a Global Average Pooling (GAP) \cite{proc:glob_pooling} to their output, the first part of the NN is constructed.

We now detail how the convolutional units and the GAP work. The adopted notation holds for the first convolutional block and, for this reason, the input is denoted by \mbox{$\mathbf{U}=$ $\left[\mathbf{u}_1,\ldots,\mathbf{u}_{N_0} \right]$}; the same reasoning holds for the second and third convolutional blocks, for which the inputs are the outputs of the first ($\mathbf{\bar{Y}}^1 \in \mathbb{R}^{L \times N_1}$) and second convolutional blocks ($\mathbf{\bar{Y}}^2 \in \mathbb{R}^{L \times N_2}$), respectively.

The output $\mathbf{Y}^1=\left[\mathbf{y}^1_1,\ldots,\mathbf{y}^1_{N_1}\right] \in \mathbb{R}^{L \times N_1}$ of $\mathcal{L}_1$ is computed as
\begin{equation}
\mathbf{y}^1_{b}= \sum_{n=1}^{N_0}\boldsymbol{\omega}_b^{1n} \ast \mathbf{u}_n \quad b=1,\ldots,N_1,
\end{equation}
where: $\ast: \left( \mathbb{R}^{H_{1}} \times \mathbb{R}^{L} \right) \rightarrow \mathbb{R}^{L}$ is the discrete convolution operator \cite{proc:SENN}; $\mathbf{\Omega}^1_b=\left[\boldsymbol{\omega}^{11}_b, \ldots, \boldsymbol{\omega}^{1N_0}_b\right]$ $\in \mathbb{R}^{H_1 \times N_0}$ are the weights, called filter kernels, applied to $\mathbf{u}_n$; $\mathbf{\Omega}^1=\left[\mathbf{\Omega}^1_1, \ldots, \mathbf{\Omega}^1_{N_1} \right] \in \mathbb{R}^{H_1 \times N_0 \times N_1}$ is the overall weights set of $\mathcal{L}_1$. Bias terms are omitted to simplify the notation.

As a second step, the BN zero-centers and normalizes $\mathbf{y}^1_b$, in order to address the issue of the vanishing/exploding gradient \cite{art:BN} that frequently affects the training of NNs in DL.

Finally, the adopted ReLU activation function reads
\begin{equation}
\bar{y}^{1l}_b=\mathcal{R}_1\left(\mathcal{B}_1\left(y^{1l}_b\right)\right)=\text{max}\left(0,\mathcal{B}_1\left(y^{1l}_b\right)\right) \quad l=1,\ldots,L ,
\end{equation}
where $y^{1l}_b$ is the $l$-th entry of $\mathbf{y}^1_b$.

Both the BN and the ReLU activation function do not involve any tunable parameter, so that the outcome of the first part of the NN is ruled just by $\mathbf{\Omega}=\left[\mathbf{\Omega}^1 | \mathbf{\Omega}^2 | \mathbf{\Omega}^3 \right]$, where $\mathbf{\Omega}^1$, $\mathbf{\Omega}^2$ and $\mathbf{\Omega}^3$ are the weights employed by $\mathcal{L}_1$, $\mathcal{L}_2$ and $\mathcal{L}_3$, respectively. Each channel $\mathbf{\bar{y}}^3_{b}$, output of the third computational block, is still shaped as a time series of length $L$, but does not represent a displacement time history anymore, as it becomes a feature of $\mathbf{U}$. A feature is an optimized representation of $\mathbf{U}$, optimality being meant in the sense of the minimization of the loss function $C$. Also the GAP does not use any tunable parameter: the input $\mathbf{\bar{Y}}^3=\left[\mathbf{\bar{y}}^3_1, \ldots, \mathbf{\bar{y}}^3_{N_3}\right] \in \mathbb{R}^{L \times N_3}$ is handled to compute an average value for each channel $\mathbf{\bar{y}}^3_b$ as
\begin{equation}
\text{GAP}\left(\mathbf{\bar{y}}^3_b\right) =\frac{1}{L}\sum_{l=1}^L \bar{y}^{3l}_b  \quad b=1,\ldots,N_3 .
\end{equation}
The GAP outcomes are synthetic description of the channel contents, highly informative for the classification task.

The implementation of the FCN architecture has been developed making use of the Keras API, based on Tensorflow. The hyperparameters featured by the NN (\textit{e.g.} the number of filters $N_k$) and controlling the training (\textit{e.g.} the number of epochs) have been initially set according to \cite{proc:time_series_base,art:Rosafalco_1}. A further hyperparameter tuning has been carried out through the repeated evaluation of the classification accuracy of $\mathcal{G}$ on the case studies discussed next. Specific attention has been paid to avoid overfitting of the training data: in Tab. \ref{tab:hyperparameters}  the hyperparameters values employed for the two case studies are reported.

\begin{table}[h!]
      \centering
       \[
		\begin{array}{ccc}
		\toprule
		 & \mbox{Portal frame} & \mbox{Integral bridge} \\
		\midrule
		N_0 &  6 &  4 \\
		N_1, N_2, N_3 & 16, 32, 16 & 16, 32, 16 \\
		H_1, H_2, H_3 &  8,  5,  3 &  8,  5,  3\\
		B   & 16 & 16 \\
		I   & 10,000 & 10,000 \\
		\text{n}^o\text{ epochs} & 500 & 1000 \\
		\bottomrule
		\end{array}
		\]
    \caption{FCN hyperparameters.\label{tab:hyperparameters}} 
\end{table}

\begin{figure}[t!]
\begin{centering}
\begin{tikzpicture}[node distance=2cm]
\node (FOM) at (-1, 0) [small_block] {FOM};
\node (eval_dam) at (2, 5) [large_block] {evaluation of \\ damage scenarios $g$ \\ (defining discrete pdf $\mathcal{P}_g$ \\ for $g=0,\ldots,G$)};
\node (eval_cond) [large_block, right of=eval_dam, xshift=3cm] {evaluation of \\ operational conditions $\boldsymbol{\eta}$ \\ (defining continuous pdf $\mathcal{P}_q$ \\ for $q=1,\ldots,Q$)};
\begin{scope}[on background layer]
\node[fit = (eval_dam)(eval_cond), basic box = black] (eval) {};
\end{scope}
\node (ROM) [small_block] at (4, 0) {ROM};
\node (D) [large_block] at (8.5, 0) {$\mathbf{D}=\lbrace \left(\mathbf{U}_1,g_1 \right), \ldots, \left(\mathbf{U}_I,g_I \right) \rbrace$ \\ assembling};
\node (train_G) [small_block] at (6.2, -3.2) {train $\mathcal{G}$};
\node (validate_G) [small_block] at (10.8, -3.2) {validate $\mathcal{G}$};
\connectwithlongtext{->}{FOM}{ROM}{POD bases determination in elasto-dynamics};
\connectwithlongtext{->}{ROM}{D}{ };
\connectwithlongtextabove{->}{eval}{FOM}{Sample $\lbrace g, \boldsymbol{\eta} \rbrace_{\tau}$ with $\tau=1,\ldots,Y$};
\connectwithlongtextabove{->}{eval}{D}{Sample $\lbrace g, \boldsymbol{\eta} \rbrace_i$ with $i=1,\ldots,I$};
\connectwithlongtextabove{->}{D}{train_G}{$\mathbf{U}_i$ with $i=1,\ldots,I_{tr}$};
\connectwithlongtextabove{->}{D}{validate_G}{$\mathbf{U}_i$ with $i=1,\ldots,I_{val}$};
\end{tikzpicture}
\caption{{Methodology flowchart.}\label{fig:meth}}
\end{centering}
\end{figure}
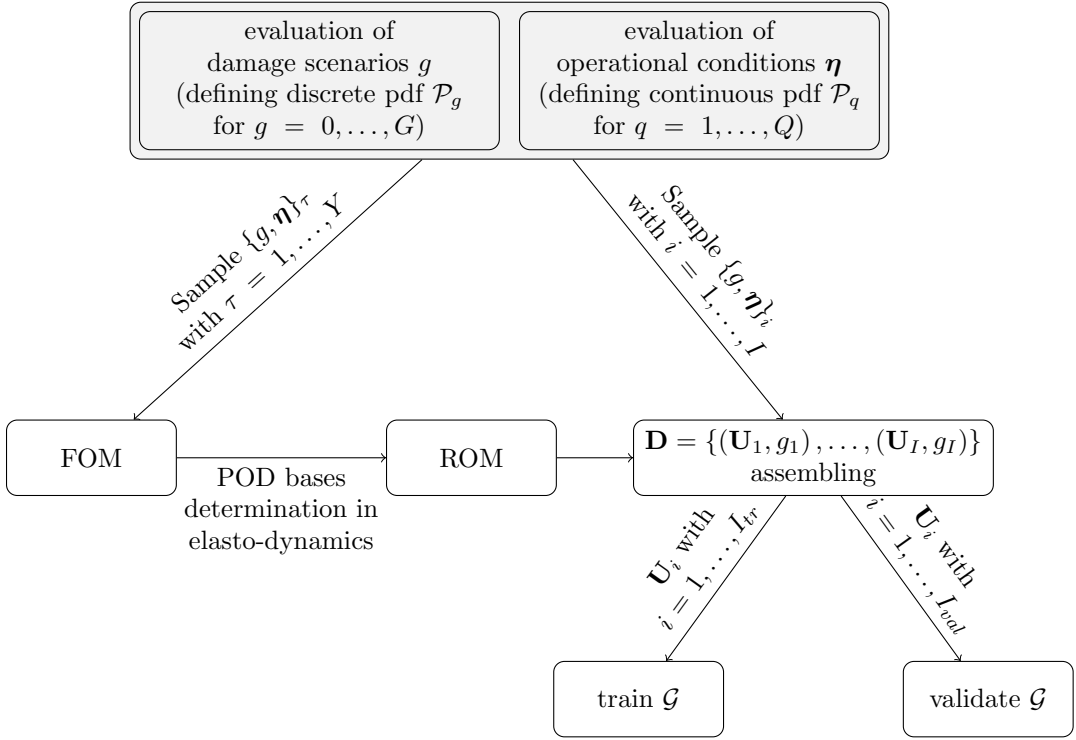

In this section, we have discussed the elements constituting the proposed methodology: the definition of a stochastically parametrized FOM of the monitored structure; the determination of a ROM keeping track of the system parametric and temporal dependence; the construction of a dataset $\mathbf{D}$; the training of a FCN-based classifier $\mathcal{G}$. To further clarify the connection between these different steps, Fig. \ref{fig:meth} is reported.

\vspace{-0.25cm}
 
\section{Case study 1 - Portal frame}
\label{sec:portal_frame}

The proposed methodology has been first  used for a two-dimensional, single-storey frame subjected to a dynamic load. The purpose of this analysis is to verify the impact on the classifier performance of the ROM handling a variable damage level and of different values of the Signal to Noise Ratio (SNR) characterizing the sensor accuracy.

\subsection{Portal frame - FOM}

\begin{figure}[h!]
%\begin{center}
\centerline{
\includegraphics[width=0.5\textwidth]{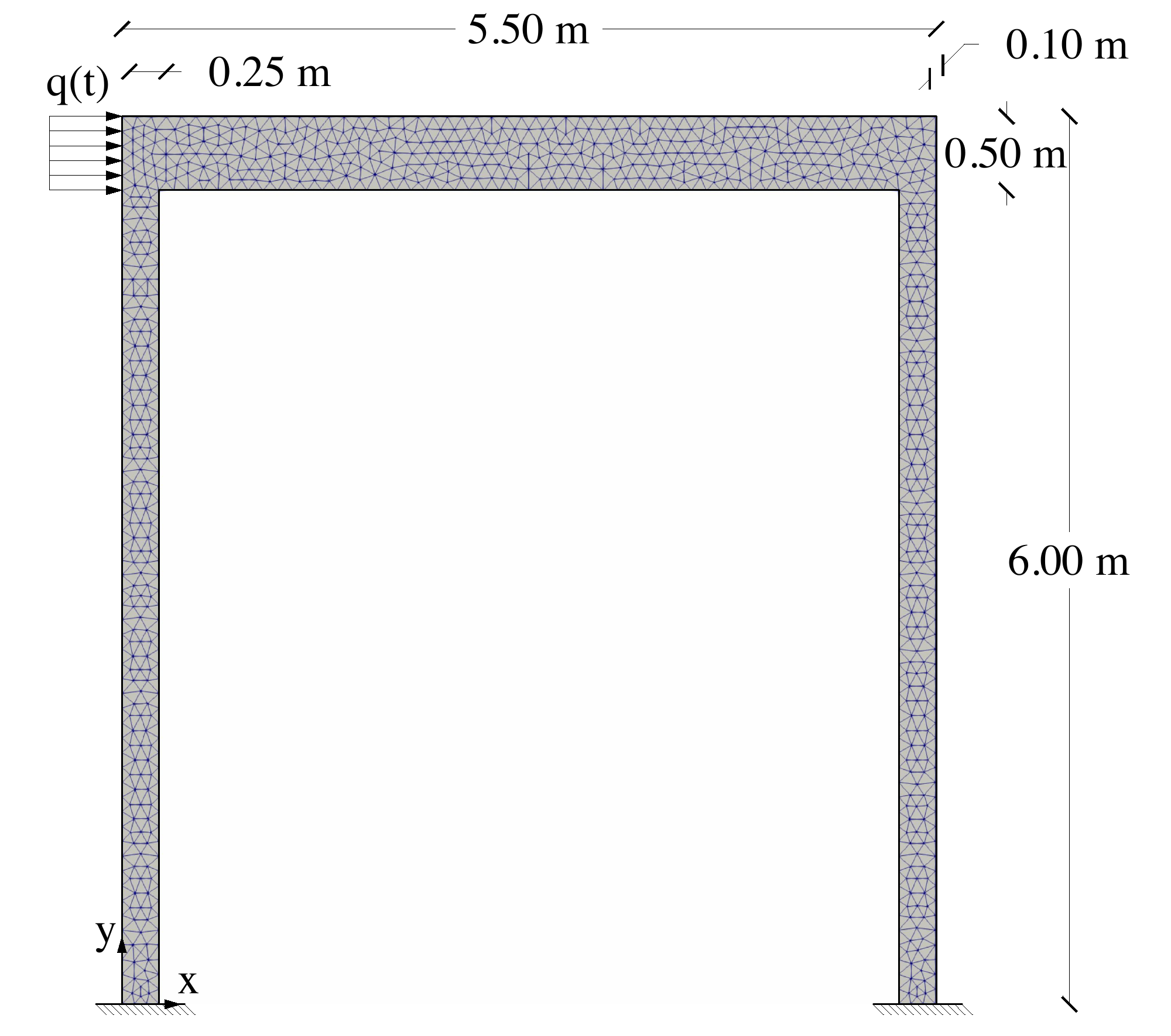}
\includegraphics[width=0.5\textwidth]{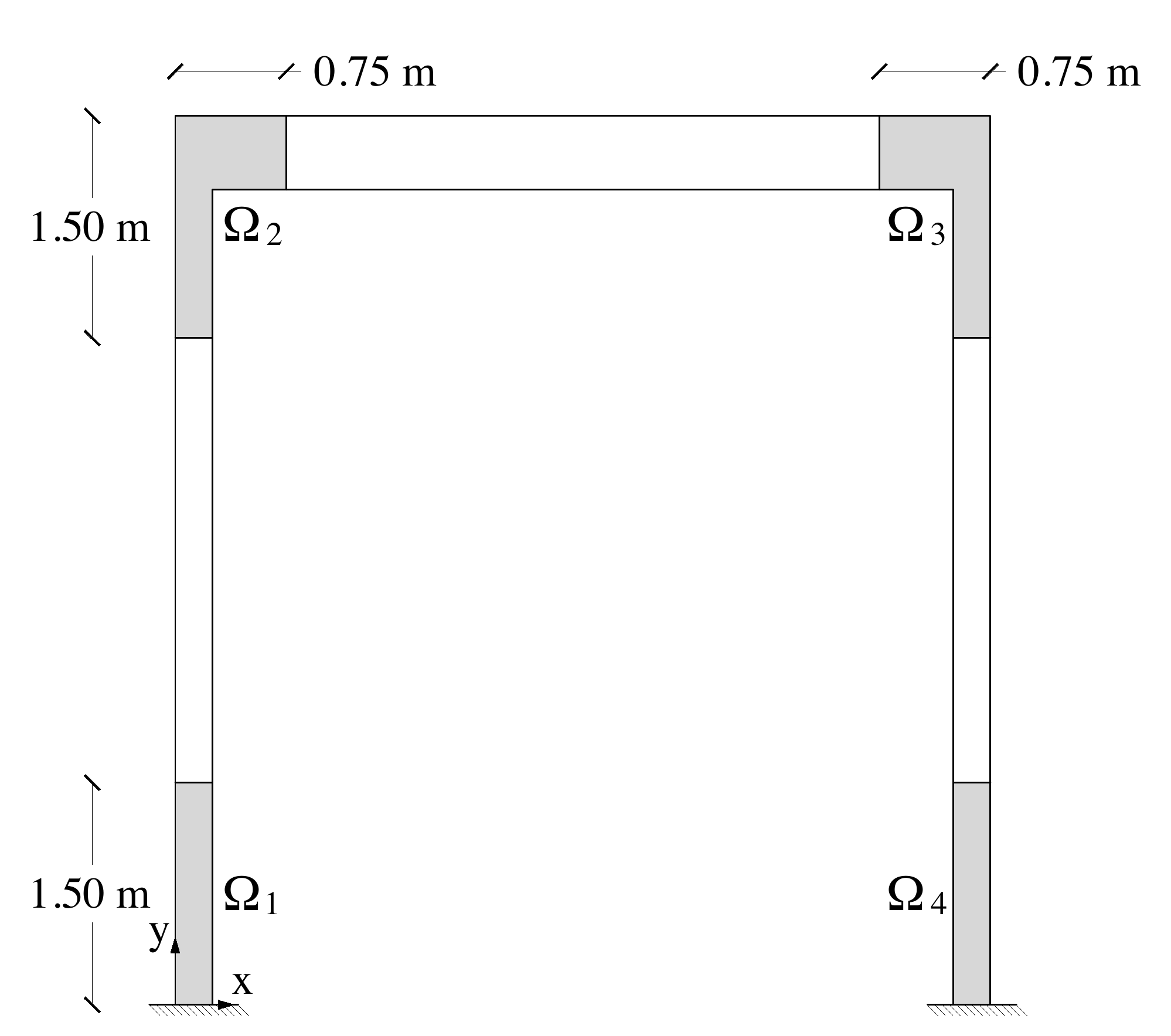}
}
\caption{{Portal frame: (left) loading and space discretization;  (right) considered damage scenarios in the subdomains $\Omega_1,\ldots,\Omega_4$.}\label{fig:fig1}}
%\caption{{C $\ \ $}\label{fig:fig2}}
%\end{center}
\end{figure}

The two-dimensional portal frame depicted in Fig. \ref{fig:fig1} has been numerically modeled via a FE discretization consisting of $1882$ constant strain triangles (CSTs) and $1884$ dofs. Time discretization has been performed by partitioning $(0,T)$ into subintervals of size $5 \cdot 10^{-3}$ s. The structural thickness has been assumed to be $0.1 \text{ m}$, so that a plane stress condition has been adopted. The structure has been assumed to be made of concrete, whose mechanical properties are: Young's modulus $E=30 \text{ GPa}$, Poisson's ratio $\nu= 0.2$, density $\rho=2500 \text{ kg}/ \text{m}^{3}$. The structure has been excited by a distributed load, acting on the left column in correspondence of the deck, varying in time as $q\left(t\right)=A\abs{\sin{\left(2\pi ft\right)}}$. The load amplitude $A$ and the frequency $f$ of the load have been modeled as random variables, with uniform pdfs $\mathcal{U}_A\left(10,50\right) \text{ kPa}$ and $\mathcal{U}_f\left(50,95\right) \text{Hz}$. 

In Fig. \ref{fig:fig1}, the considered damage scenarios are shown on the right. Each relevant structural state $g \in \lbrace1,2,3,4\rbrace$ is linked to a damage of the corresponding subdomain $\Omega_{1},\dots,\Omega_{4}$, while the undamaged scenario is given by $g=0$. If not stated otherwise, damaged and undamaged scenarios have been assumed to have equal probability to be encountered during the monitoring stage, therefore a discrete uniform pdf $\mathcal{U}_g\left(0,\dots,4\right)$ has been assumed for $g$. The damage level $\delta$, which represents the stiffness reduction applied to the considered subdomain, has been modeled as a continuous random variable with uniform pdf $\mathcal{U}_{\delta}\left(2\%,25\%\right)$. Therefore, for the current analysis the parametric dependence has involved $\boldsymbol{\eta}=\lbrace A,f,\delta\rbrace$.
As load amplitude $A$, load frequency $f$ and damage level $\delta$ may vary continuously, the adopted pdfs  $\mathcal{U}_{A}$, $\mathcal{U}_{f}$ and $\mathcal{U}_{\delta}$ are continuous too; on the other hand, since the damage scenario can only take values in the discrete set $\lbrace 0,1,2,3,4\rbrace$, a discrete pdf  $\mathcal{U}_g$ has been adopted.

The dataset $\mathbf{D}$ has been built by collecting instances, together with the corresponding labels, obtained for parameters sampled via LHS from the parametric space spanned by $g$ and $\boldsymbol{\eta}$. In Tab. \ref{tab:TB_1} the first eight vibration frequencies and relevant periods of vibration of the model are listed. The monitoring system consists of $N_{0}=6$ sensors, recording either the horizontal or vertical accelerations  as depicted in Fig. \ref{fig:fig11}. The signals have been recorded with a sampling frequency of $200 \text{ Hz}$, allowing to properly account for the first seven structural frequencies without incurring in aliasing. Each numerical simulation covers $1\text{ s}$ of duration: therefore, in the monitoring interval, each record includes $L=200$ samples. \smallskip

\begin{minipage}{\textwidth}
\begin{minipage}{0.34\textwidth}
\begin{table}[H]
      \centering
      \footnotesize
       \[
		\begin{array}{ccc}
		\toprule
	\text{Mode} & f_{Num} \left[\text{Hz}\right] & \text{Period} \left[\text{s}\right] \\
	\midrule
		1 & 4.02  & 0.2488 \\
		2 & 24.18 & 0.0413 \\
		3 & 31.31 & 0.0319 \\
		4 & 36.71 & 0.0272 \\
		5 & 79.44 & 0.0126 \\
		6 & 80.89 & 0.0124 \\
		7 & 96.70 & 0.0103 \\
		8 & 128.66 & 7.77 \cdot 10^{-3} \\
		\bottomrule
		\end{array}
		\]
\captionsetup{width=.9\linewidth}
\caption{{Portal frame: structural frequencies.}\label{tab:TB_1}}
\end{table}
\end{minipage}
\hspace{0.05cm}
\begin{minipage}{0.64\textwidth}
\begin{figure}[H]
\includegraphics[width=0.9\textwidth]{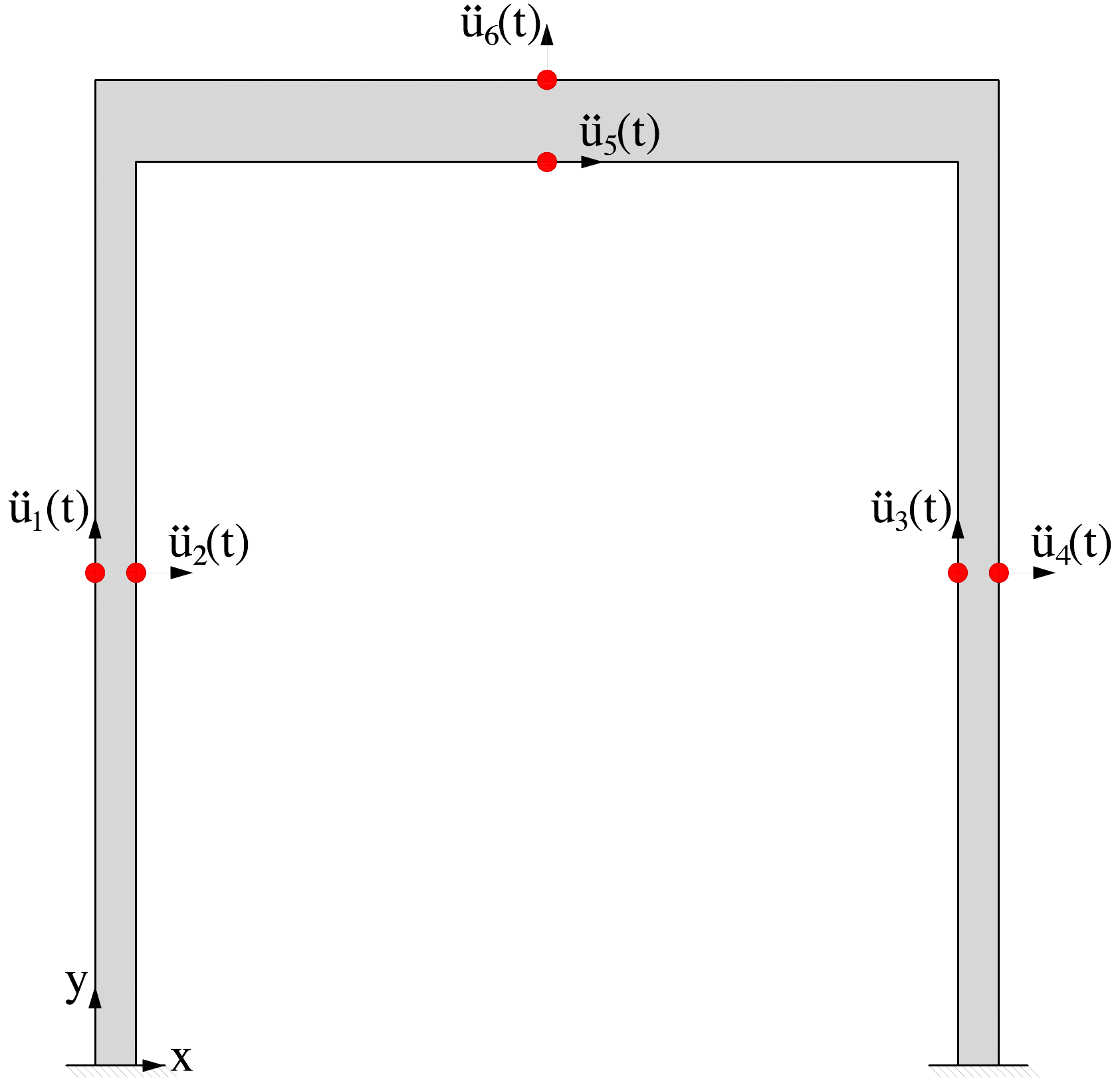}
\captionsetup{width=.9\linewidth}
\caption{{Portal frame: sensor system arrangement.}\label{fig:fig11}}
\end{figure}
\end{minipage}
\end{minipage}

\subsection{Portal frame - ROM}
To build the ROM, the snapshots have been collected for different values of $\lbrace g,\boldsymbol{\eta}\rbrace$. At this stage, no noise has been added to corrupt the model outcomes. The number of samples of $\lbrace g,\boldsymbol{\eta}\rbrace$ has been fixed to $Y=200$; the number of samples in time has been instead fixed to $X=100$ for time windows of $0.5 \text{ s}$. The total number of collected snapshots therefore amounts to $S=200\times 100=20000$.
  
Including the damage level $\delta$ inside $\boldsymbol{\eta}$ proved necessary in order to identify the presence of either a minimal or a moderate structural damage. Indeed, it has been observed that a classifier trained for a fixed damage level does not work properly in recognizing structural states characterized by a different level of damage. As an example, Tab. \ref{tab:TB_2} provides the performance, given in terms of classification accuracy, of a classifier $\mathcal{G}$ trained for $\delta=25\%$ in recognizing instances characterized by different values of $\delta$. The performance drops while moving away from the training value,  due to major difficulties in recognizing a lower damage level, and to the fact that the NN has  only been trained to distinguish between the conditions characterized by either $\delta=25\%$ or $\delta=0\%$.

The damage level $\delta$ also shows a close relationship with the number $W$ of POD bases: Tab. \ref{tab:TB_3} shows how $W$ tends to reduce, as $\delta$ decreases. This happens because the smaller the ROM dimension $W$ by which the prescribed error tolerance $\varepsilon=10^{-4}$ is achieved, the more similar are the scenarios to be described. This also means that the ROM is much more prone to fail when modelling structural states characterized by smaller value of $\delta$, because a smaller number of POD bases hardly catches the effect of small damages, as confirmed by the classification performance obtained with different classifiers trained and tested for an assigned value of $\delta$, see Tab. \ref{tab:TB_4}.
Even if not reported here for the sake of brevity, such accuracy has turned out to be not affected by the structural damping: the same analyses have been run by allowing for a Rayleigh damping featuring a ratio of 5\% for the first two structural modes, with no variations with respect to the values reported in Tab. \ref{tab:TB_4}.

\begin{figure}[h!]
\captionsetup[subfigure]{justification=centering}
%\begin{framed}
\centering
\vspace{-0.4 cm}
\subfloat[]{\includegraphics[scale=0.27]{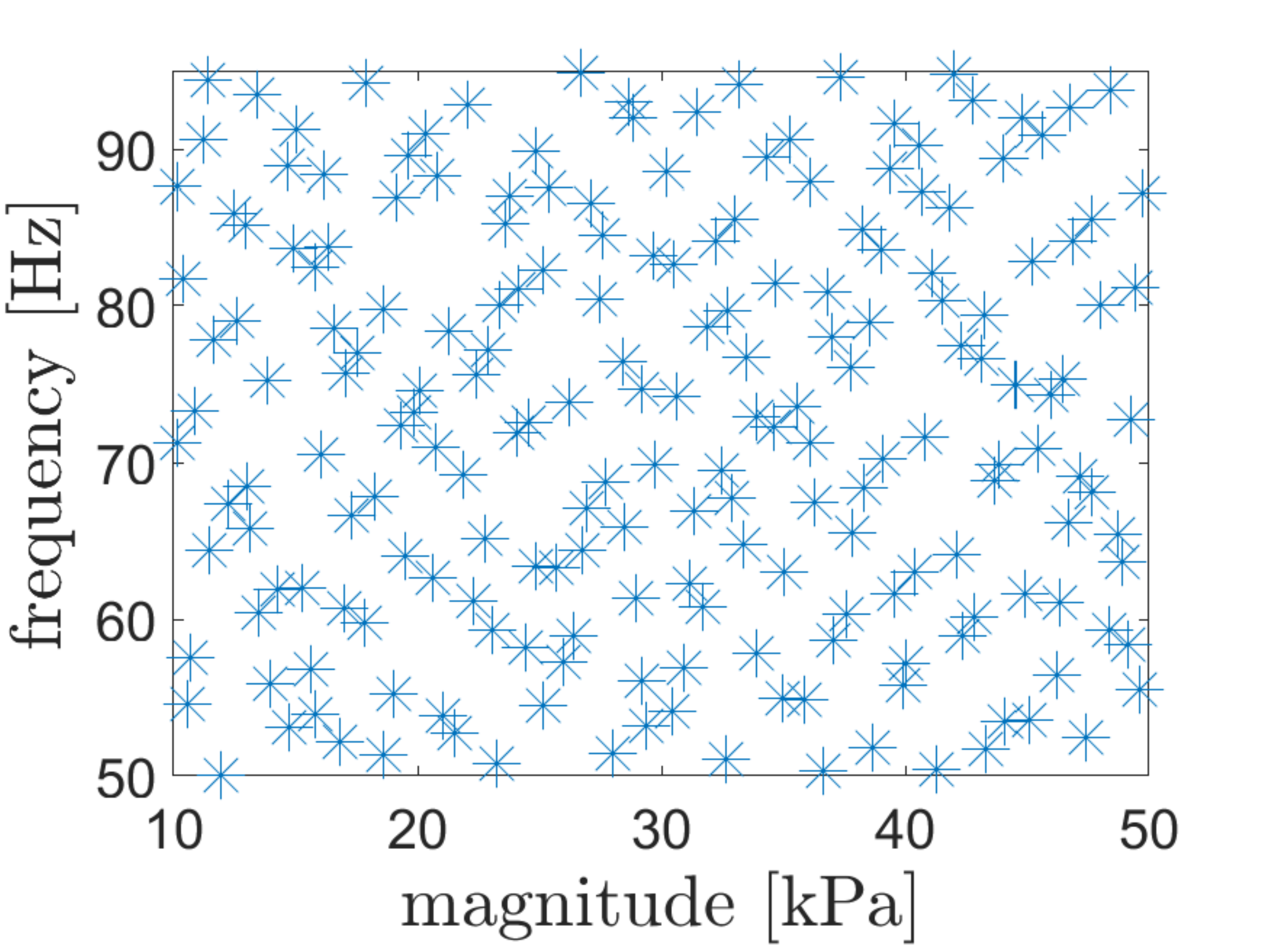}} $~$
\subfloat[]{\includegraphics[scale=0.27]{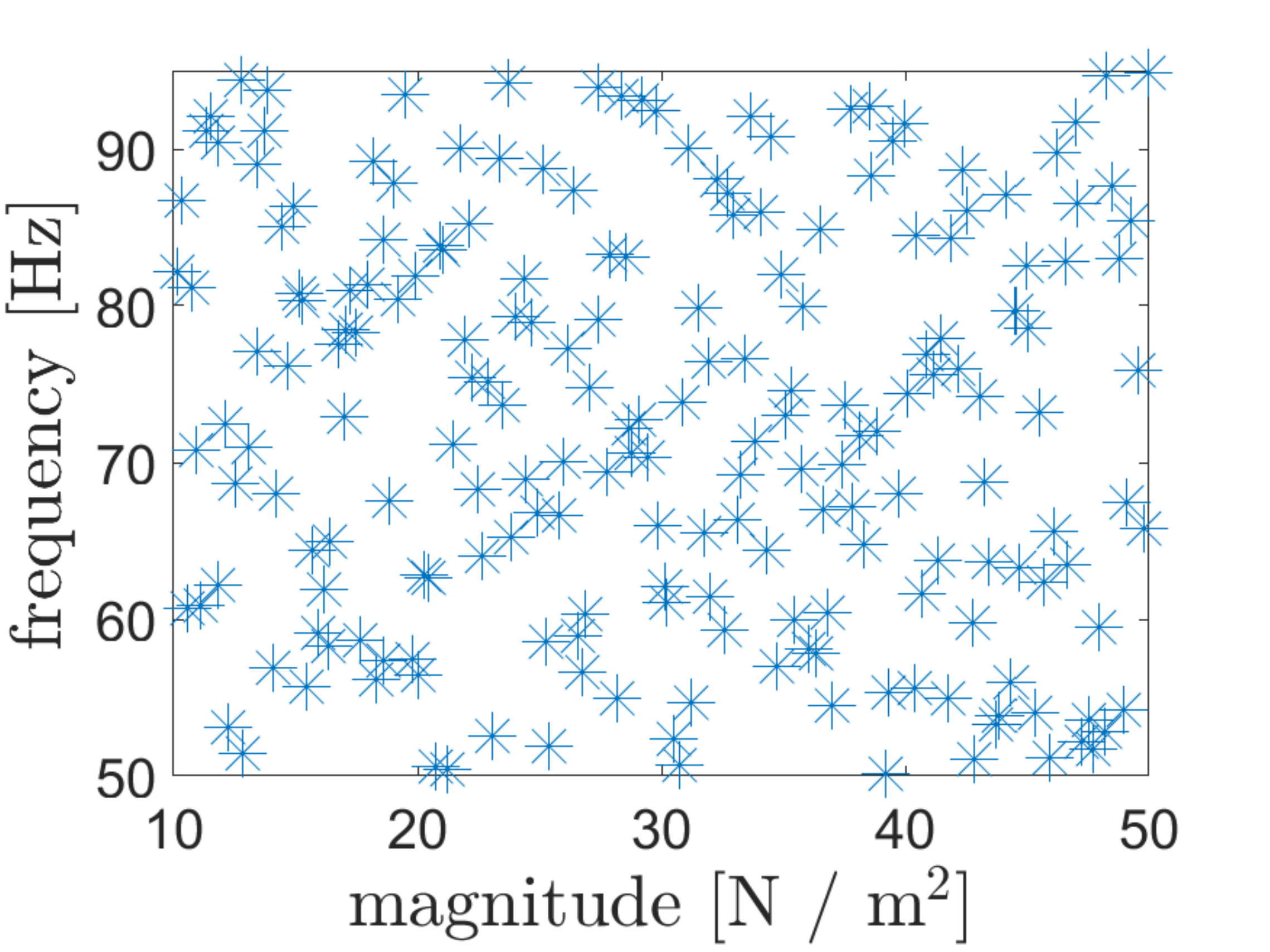}} \\
\subfloat[]{\includegraphics[scale=0.27]{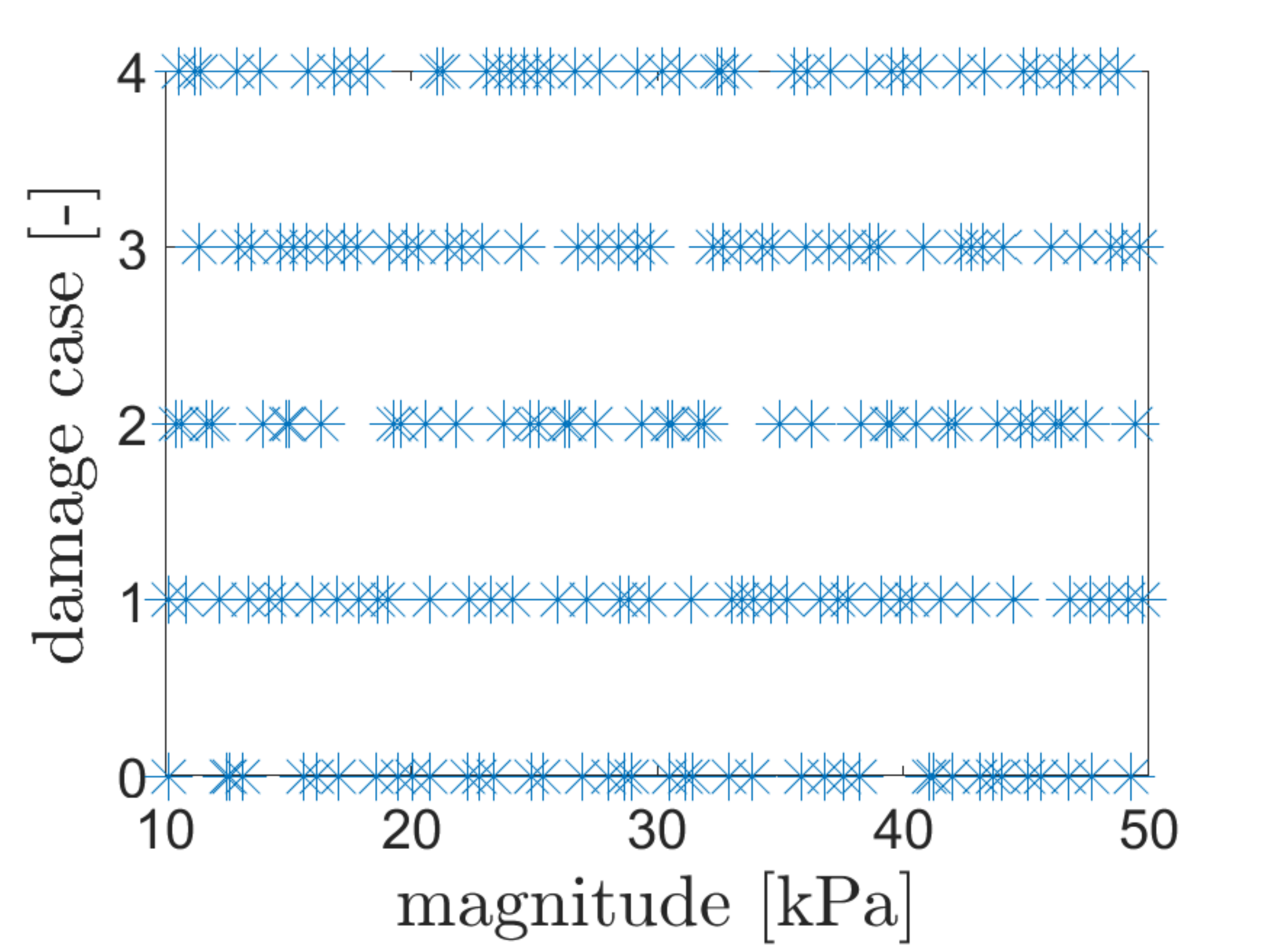}} $~$
\subfloat[]{\includegraphics[scale=0.27]{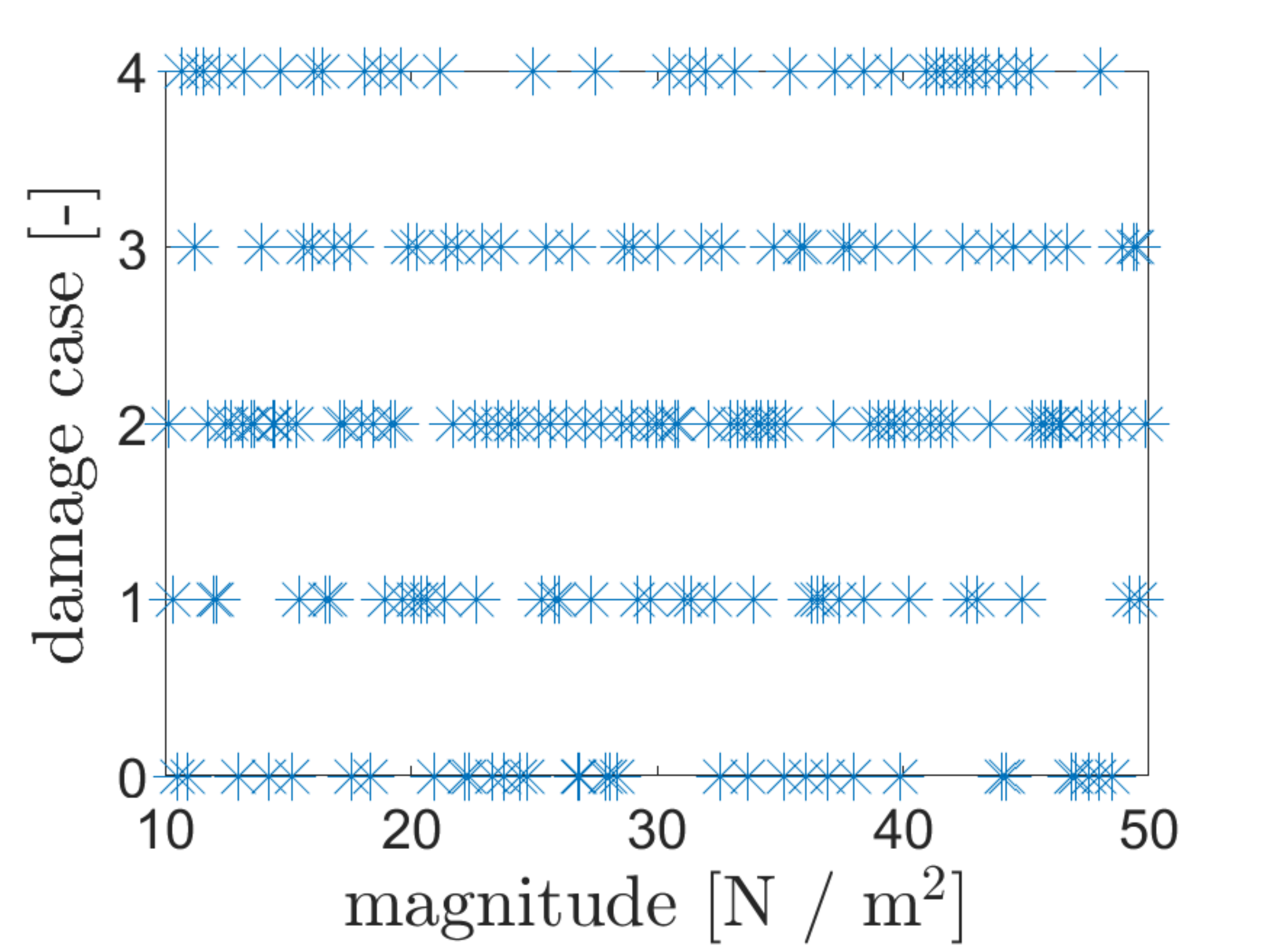}} \\
\subfloat[]{\includegraphics[scale=0.27]{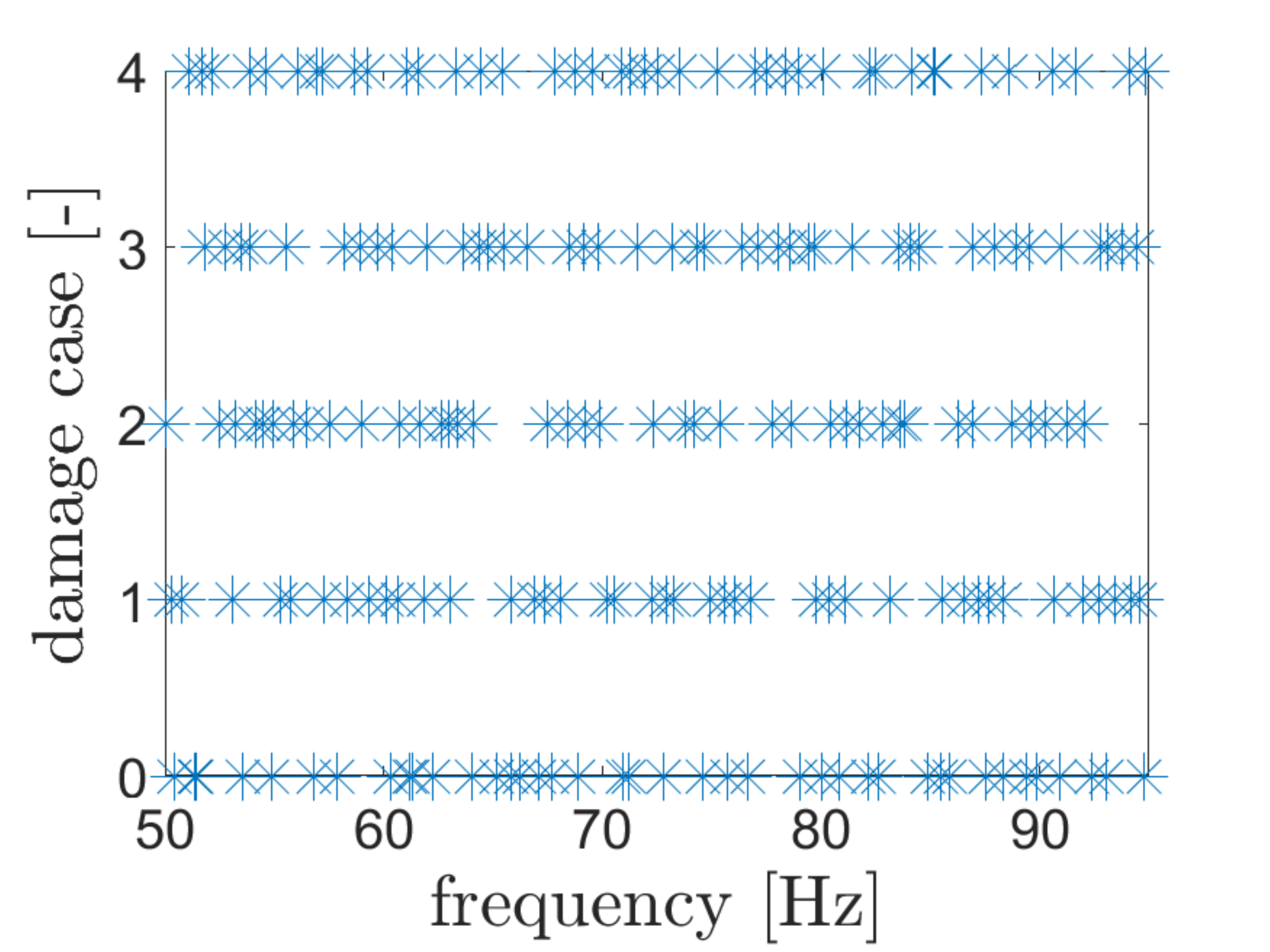}} $~$
\subfloat[]{\includegraphics[scale=0.27]{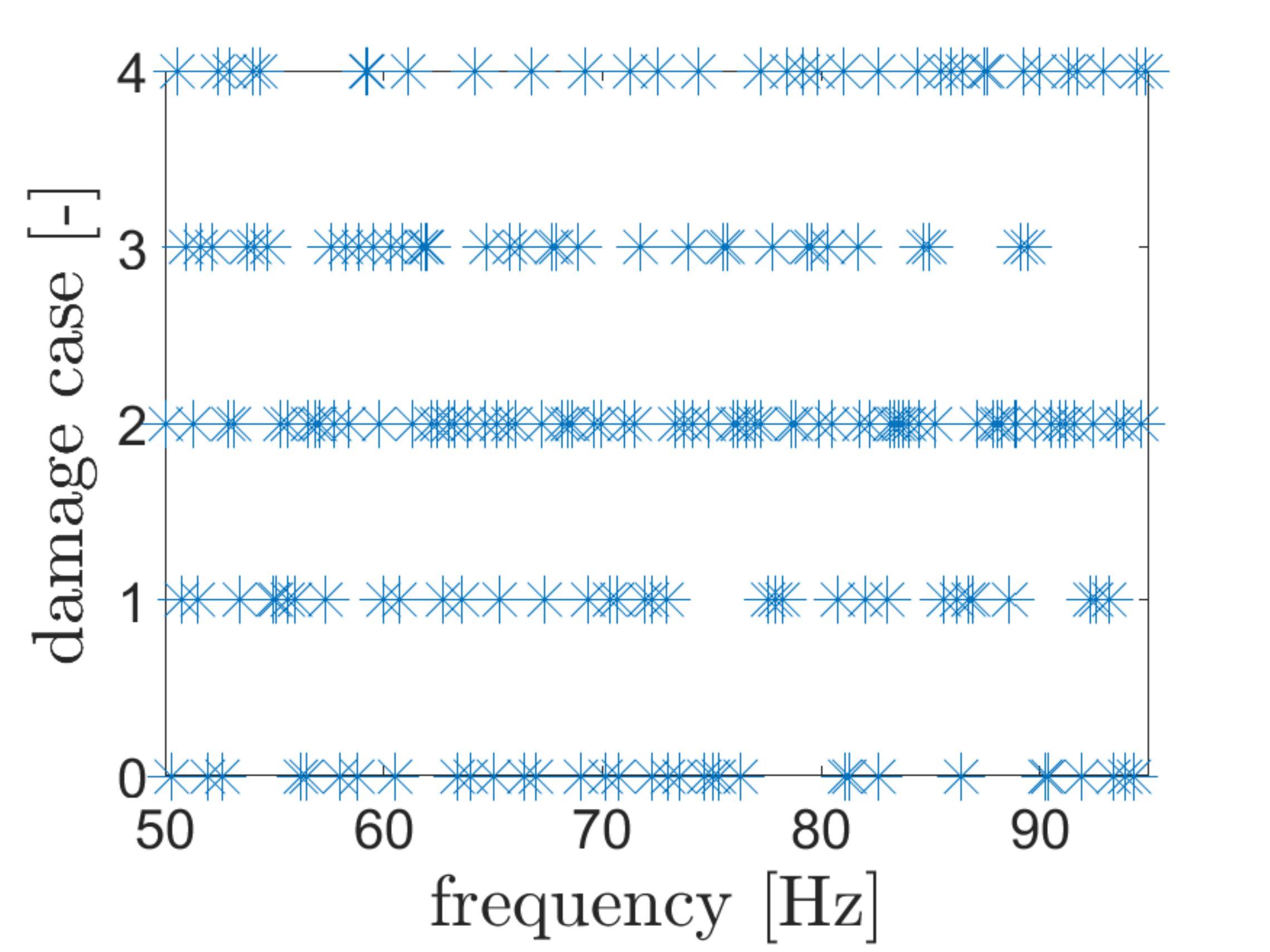}} \\
\caption{\small Portal frame: sampled values of load amplitude $A$, load frequency $f$ and damage scenario $g$ in: (left column) case (a), featuring a uniform pdf $\mathcal{U}_g$ over all the damage scenarios; (right column) case (b), featuring a probability of scenario $g=2$ to occur twice the others.\label{fig:maps_new}}
%\end{framed}
\end{figure}

Additional results have been obtained by considering the discrete pdf $\mathcal{U}_g$ relevant to the damage scenarios to be not uniform. This may be traced back to a former inspection of the structure, to ascertain if a specific damage state can occur more likely than others (maybe due to some defects in the initial state), or to a global sensitivity analysis to provide insights into the links between input loading and probability to incept a specific damage pattern. Datasets can be generated handling different  $\mathcal{U}_g$; results are here discussed for a case featuring a probability of damage scenario $g=2$ to occur, twice the others. The load amplitude and frequency have been instead extracted from the same pdfs $\mathcal{U}_{A}$ and $\mathcal{U}_{f}$ defined before. Maps of the sampled values are reported in Fig. \ref{fig:maps_new}, in terms of projections onto the planes $A-f$, $A-g$ and $f-g$, where it can be easily recognized that the damage scenario $g=2$ has a higher probability testified by the denser distribution of the samples at varying magnitude and frequency of the load. For $\delta=25\%$, the accuracy in classification relevant to case (a) has been the already considered 100\% reported in Tab. \ref{tab:TB_4}; the same accuracy has been obtained for case (b). This outcome testifies that the proposed method is robust against improper assumptions regarding the probability of the different damage scenarios to occur.

\begin{table}[h!]
\centering
\begin{minipage}[t]{0.27\textwidth}
\centering
      \footnotesize
       \[
		\begin{array}{cc}
		\toprule
	\delta \left[\%\right] & \text{Accuracy}\left[\%\right] \\
	\midrule
		25 & 100 \\
		20 & 90 \\
		15 & 82 \\
		10 & 36 \\
		5 & 22 \\
		2 & 22 \\
		\bottomrule
		\end{array}
		\]
\captionsetup{}
\caption{{Portal frame: accuracy performance obtained with a classifier trained for $\delta=25\%$, in recognizing structural states characterized by different values of the damage level $\delta$.}\label{tab:TB_2}}
\end{minipage}
\hspace{0,5cm}
\begin{minipage}[t]{0.27\textwidth}
      \centering
      \footnotesize
       \[
		\begin{array}{cc}
		\toprule
	\delta \left[\%\right] & \text{POD bases} \\
	\midrule
		25 & 65 \\
		20 & 64 \\
		15 & 58 \\
		10 & 56 \\
		5 & 53 \\
		2 & 47 \\
		\bottomrule
		\end{array}
		\]
\caption{{Portal frame: number $W$ of POD bases at varying damage level $\delta$.}\label{tab:TB_3}}
\end{minipage}
\hspace{0,5cm}
\begin{minipage}[t]{0.27\textwidth}
\centering
      \footnotesize
       \[
		\begin{array}{cc}
		\toprule
	\delta \left[\%\right] & \text{Accuracy}\left[\%\right] \\
	\midrule
		25 & 100 \\
		20 & 100 \\
		15 & 96 \\
		10 & 92 \\
		5 & 84 \\
		2 & 62 \\
		\bottomrule
		\end{array}
		\]
\caption{{Portal frame: accuracy performance obtained with classifiers trained and tested for several values of the damage level $\delta$.}\label{tab:TB_4}}
\end{minipage}
\end{table}

\begin{figure}[h!]
\centering
\begin{minipage}[t]{0.48\textwidth}
\centering
\begin{tikzpicture}
  \node (POD_time_graph_1)  {\includegraphics[scale=0.35]{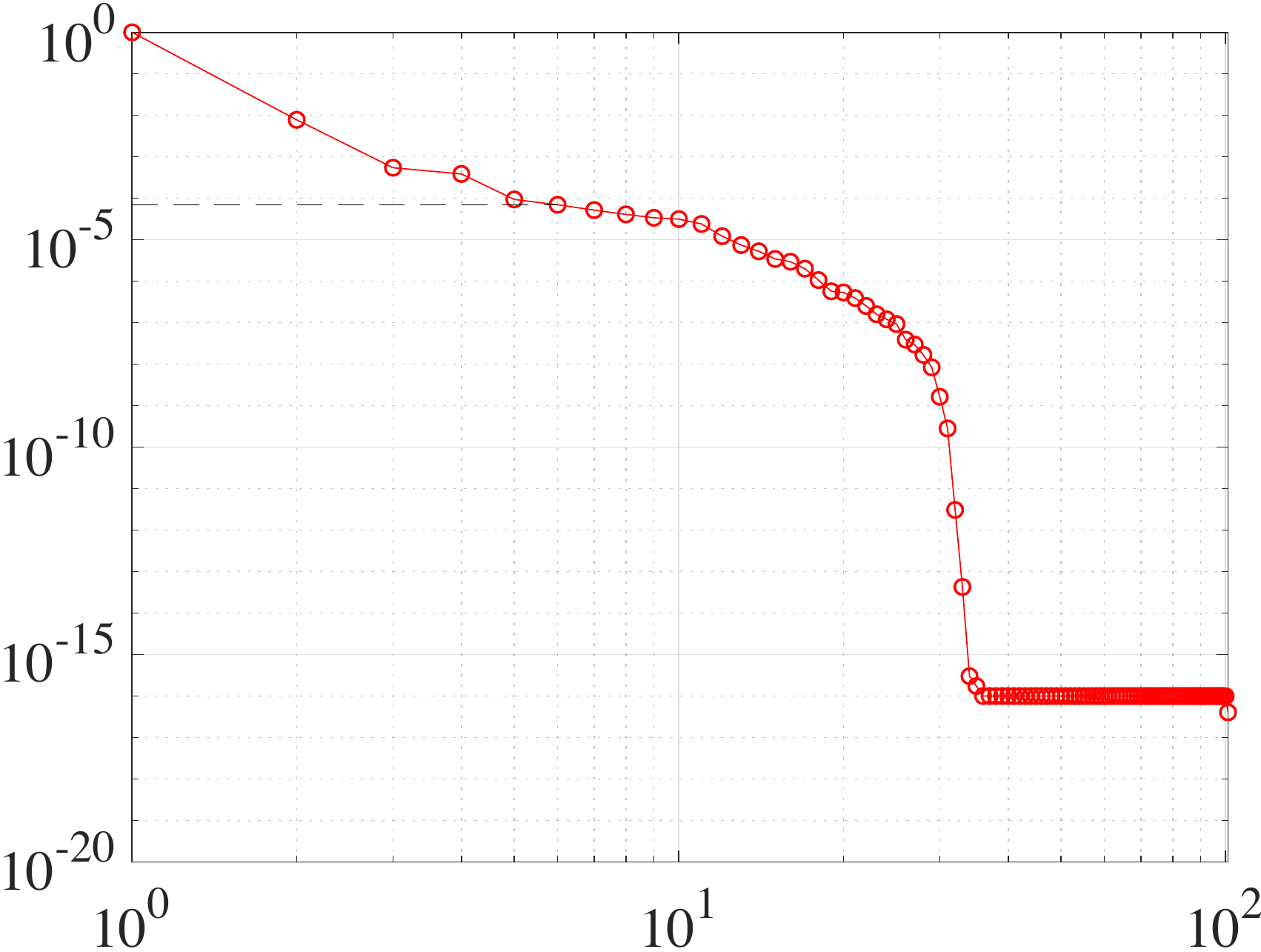}};
  \node[below=of POD_time_graph_1, node distance=0cm, yshift=1cm,font=\color{black}] {Number of snapshots};
  \node[left=of POD_time_graph_1, node distance=0cm, rotate=90, anchor=center,yshift=-0.9cm,font=\color{black}] {$\sigma^{T}_s / \sigma^{T}_1$};
 \end{tikzpicture}
\captionsetup{width=.9\linewidth}
\caption{{Portal frame: POD in time. Descent of the singular values $\sigma^{T}_s$ normalized with respect to $\sigma^{T}_1$.}\label{fig:fig12}}
\end{minipage} 
\hspace{0.05cm}
\begin{minipage}[t]{0.48\textwidth}
\centering
\begin{tikzpicture}
  \node (POD_param_graph_1)  {\includegraphics[scale=0.35]{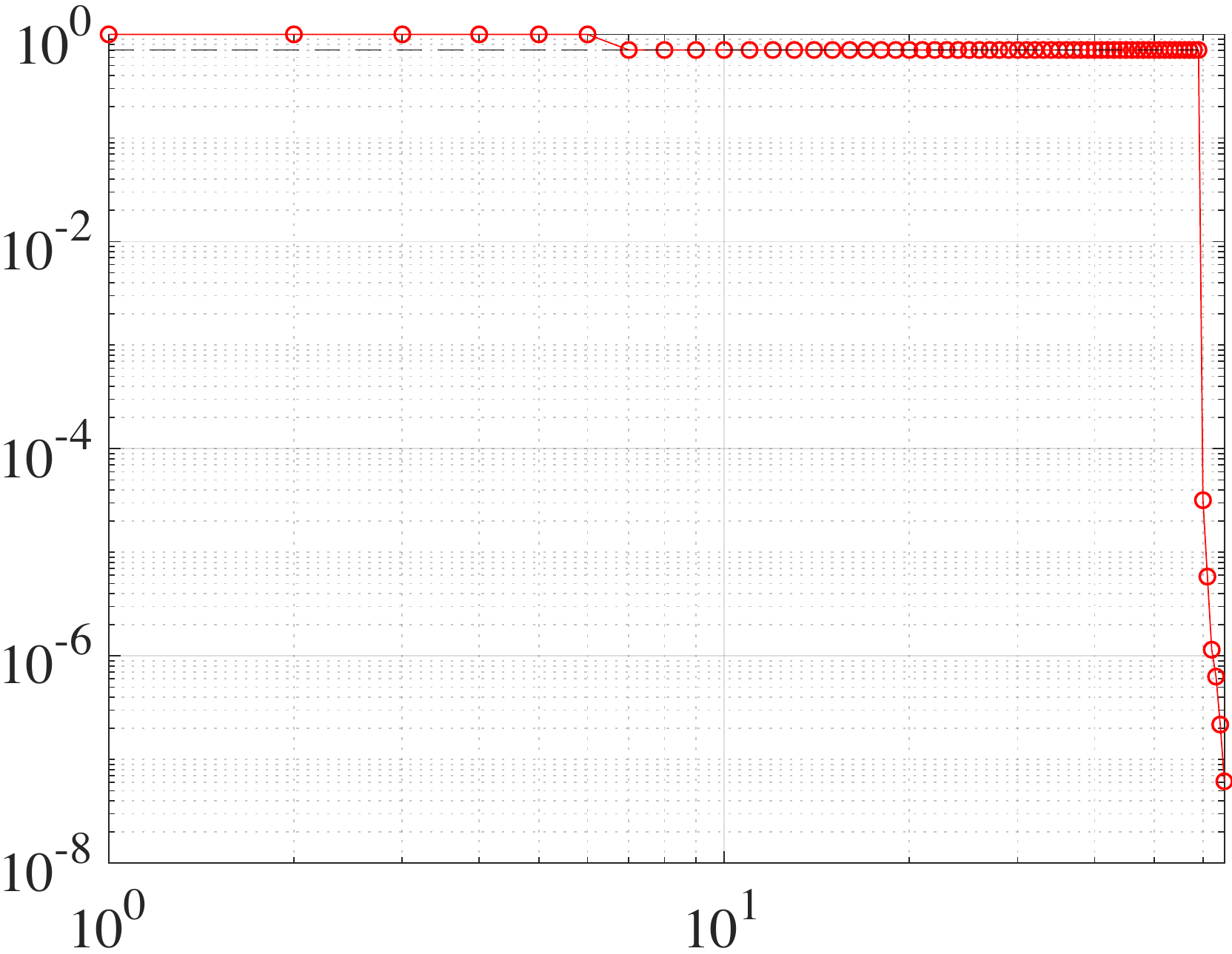}};
  \node[below=of POD_param_graph_1, node distance=0cm, yshift=1cm,font=\color{black}] {Number of snapshots};
  \node[left=of POD_param_graph_1, node distance=0cm, rotate=90, anchor=center,yshift=-0.9cm,font=\color{black}] {$\sigma^{g\boldsymbol{\eta}}_s /\sigma^{g\boldsymbol{\eta}}_1$};
 \end{tikzpicture}
\captionsetup{width=.9\linewidth}
\caption{{Portal frame: POD over parameters. Descent of the singular values $\sigma^{g\boldsymbol{\eta}}_s$ normalised with respect to $\sigma^{g\boldsymbol{\eta}}_1$.}\label{fig:fig13}}
\end{minipage}
\end{figure}

\begin{figure}[h!]
\captionsetup[subfigure]{justification=centering}
%\begin{framed}
\centering
\vspace{-0.4 cm}
\subfloat[[$1^{\circ}$ POD basis.][\label{fig:fig14}]{\includegraphics[scale=0.07]{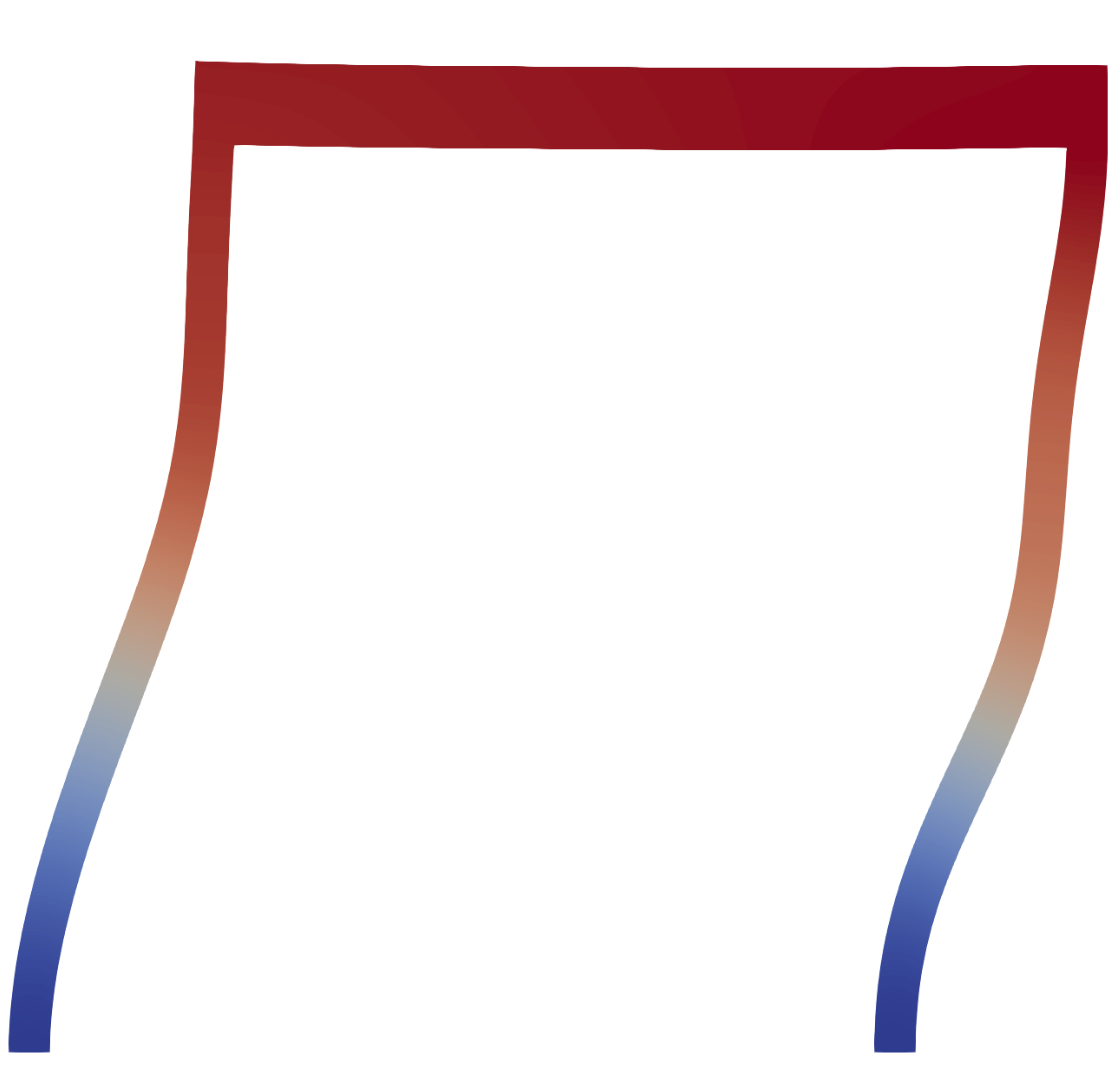}} $~$
\subfloat[[$2^{\circ}$ POD basis.][\label{fig:fig15}]{\includegraphics[scale=0.07]{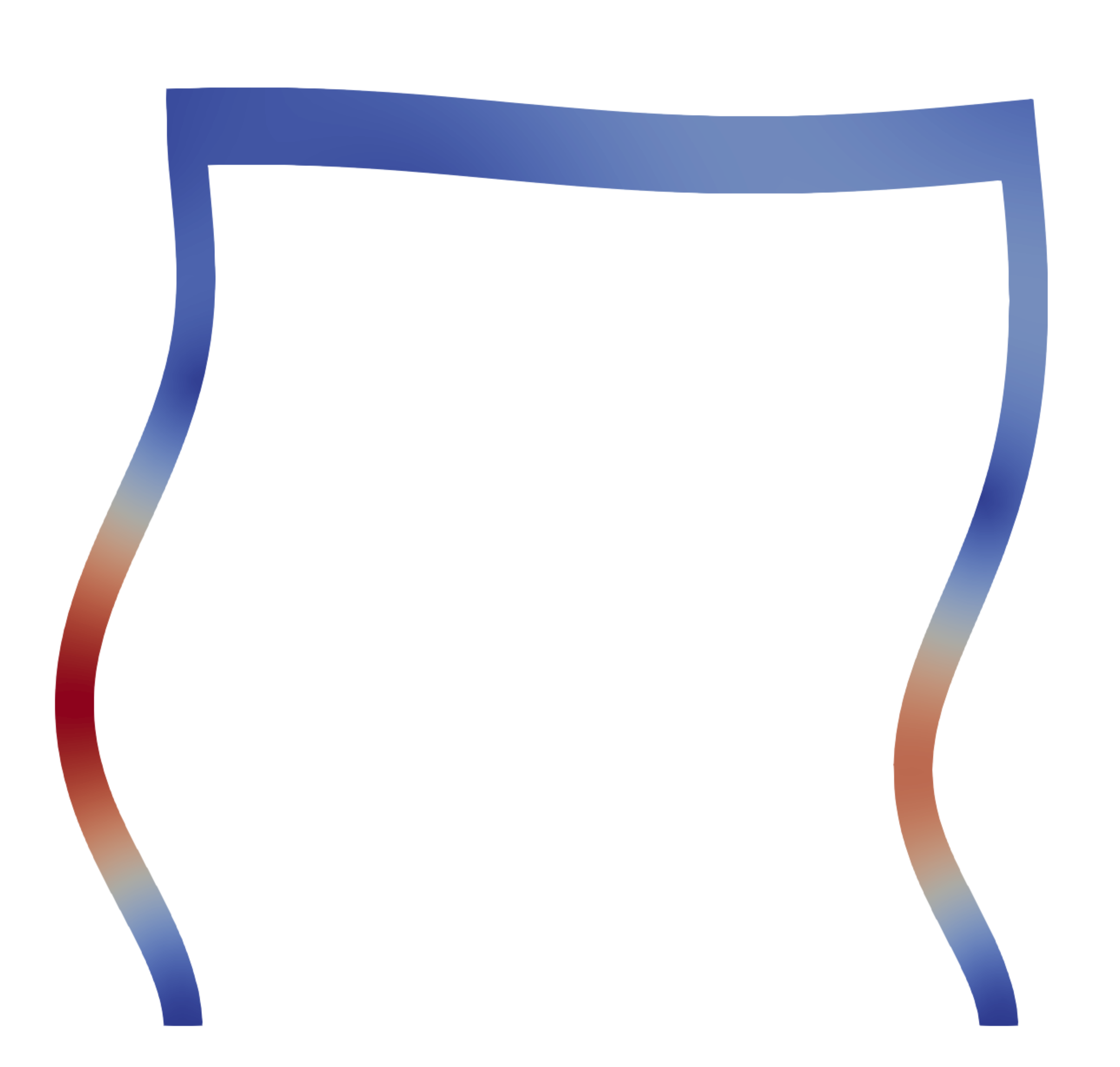}}$~$
\subfloat[[$3^{\circ}$ POD basis.][\label{fig:fig16}]{\includegraphics[scale=0.07]{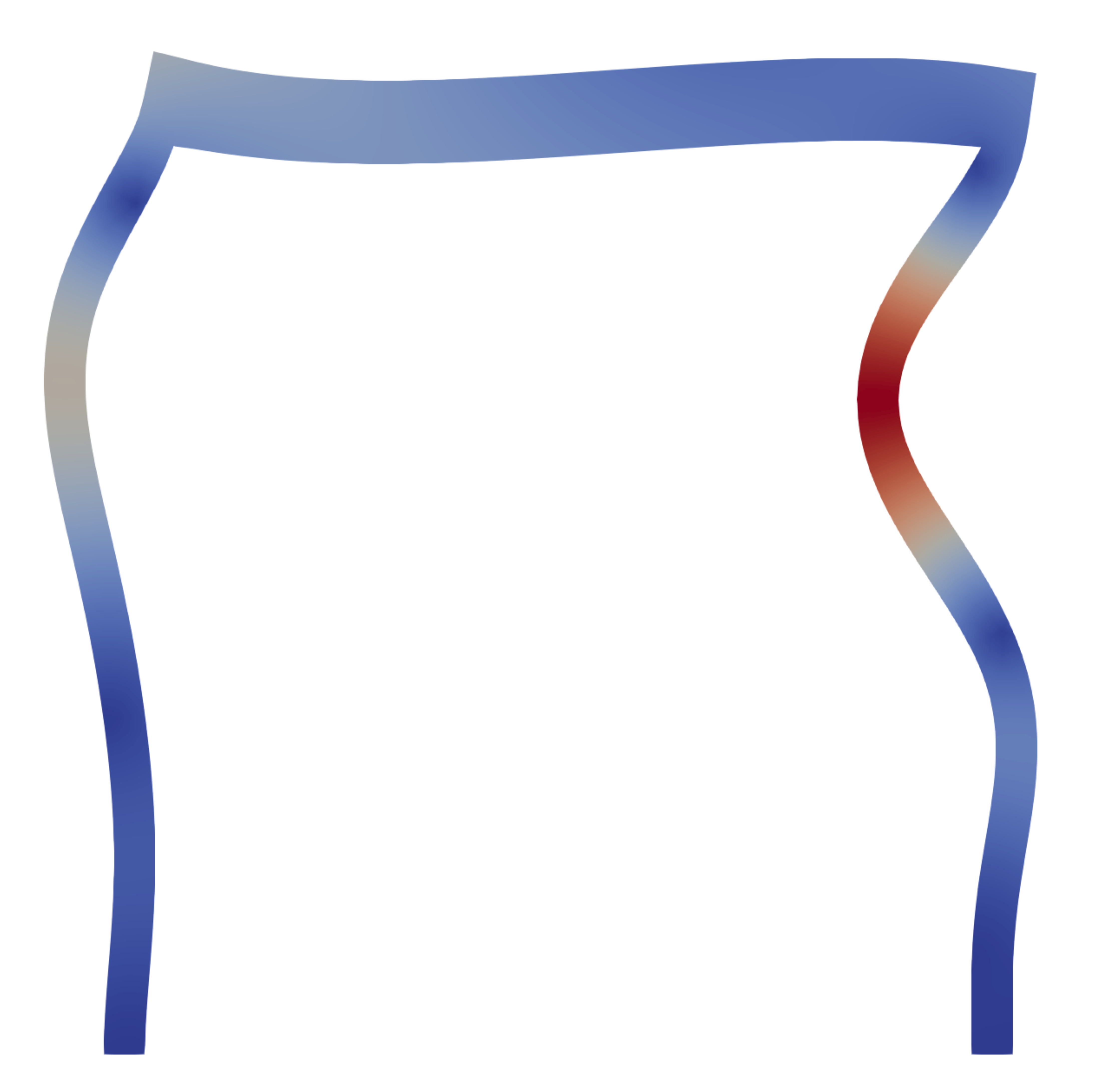}}$~$
\subfloat[[$4^{\circ}$ POD basis.][\label{fig:fig17}]{\includegraphics[scale=0.07]{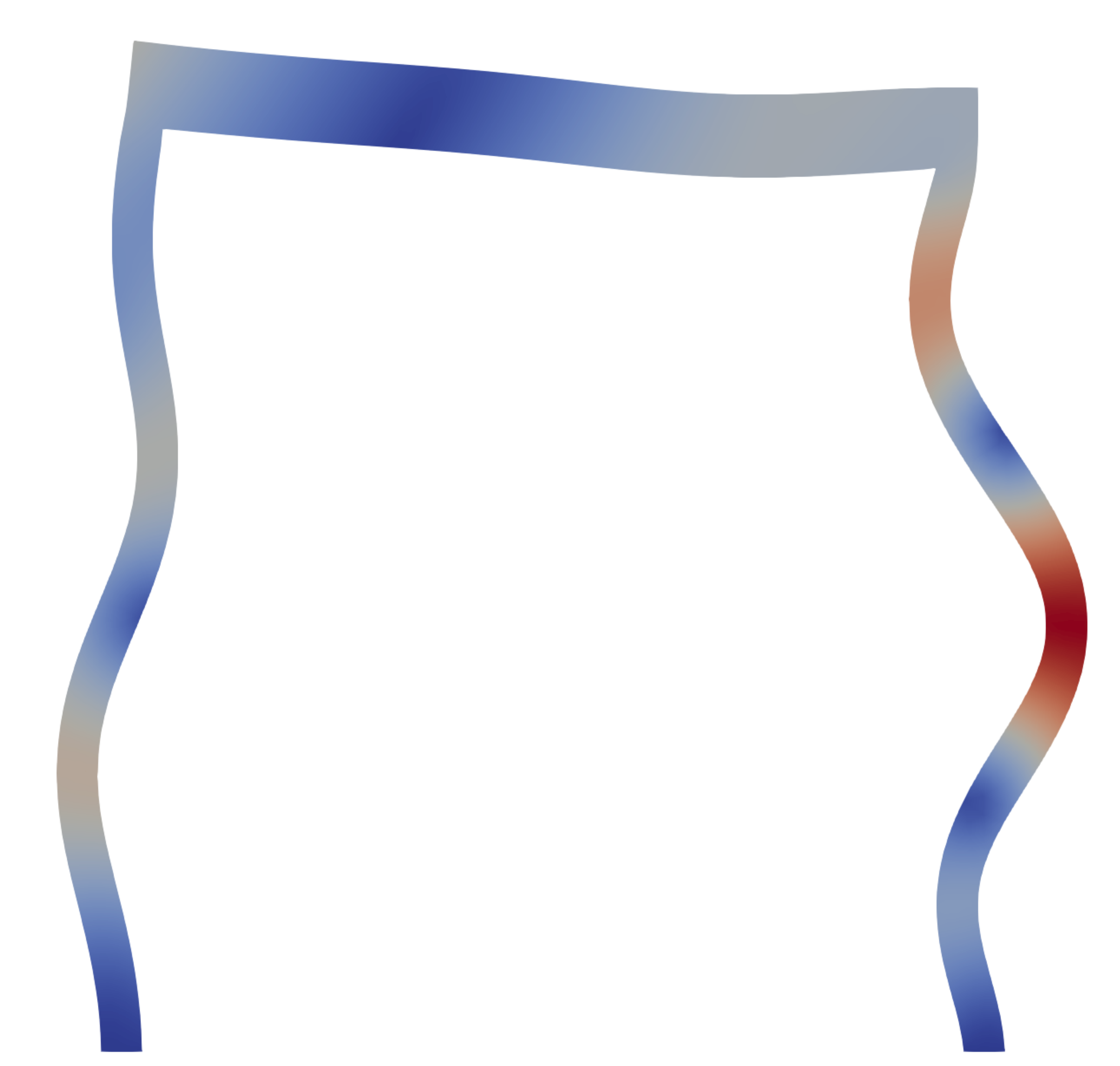}}\\
\subfloat[[$5^{\circ}$ POD basis.][\label{fig:fig18}]{\includegraphics[scale=0.07]{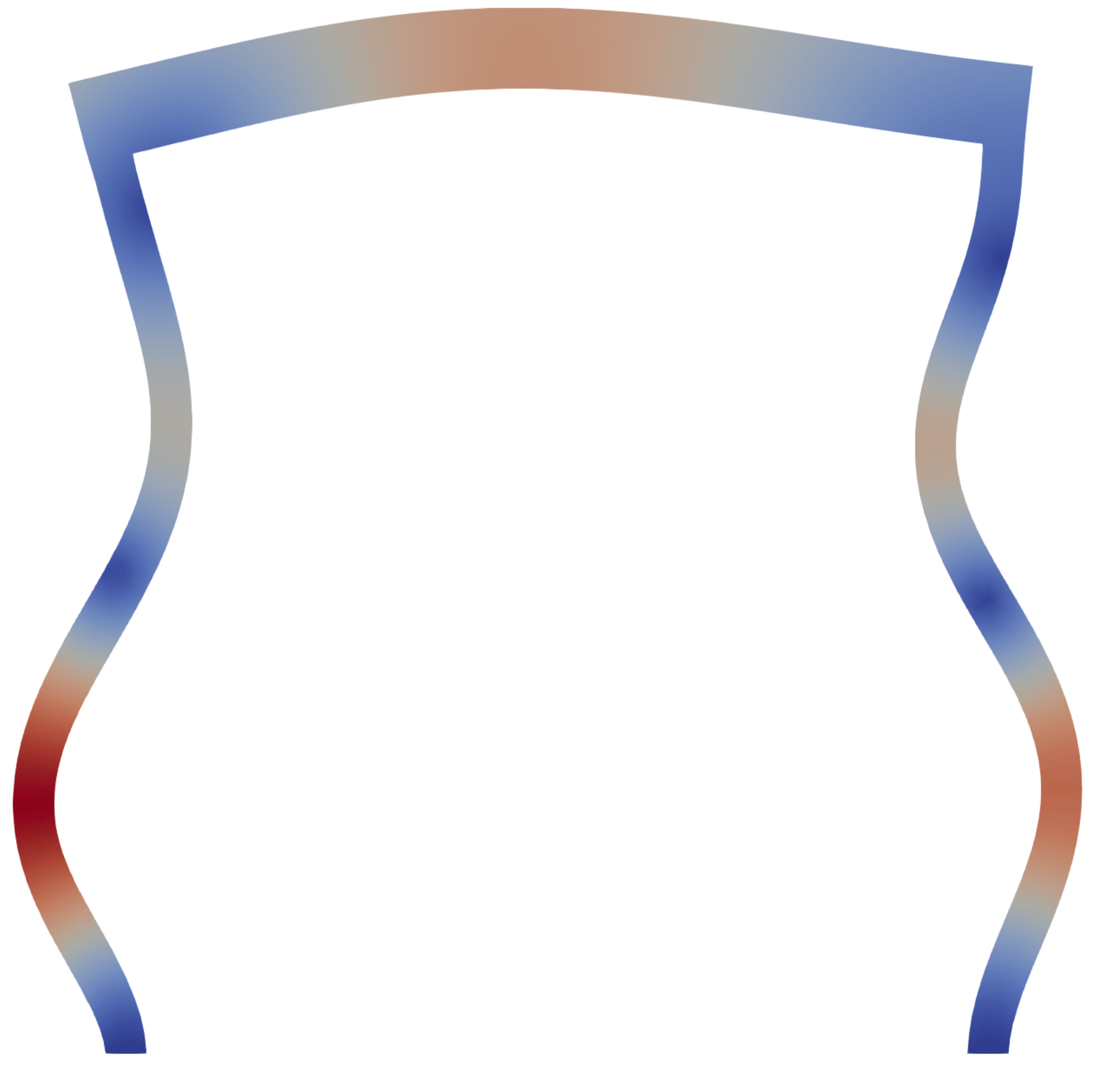}} $~$
\subfloat[[$6^{\circ}$ POD basis.][\label{fig:fig19}]{\includegraphics[scale=0.07]{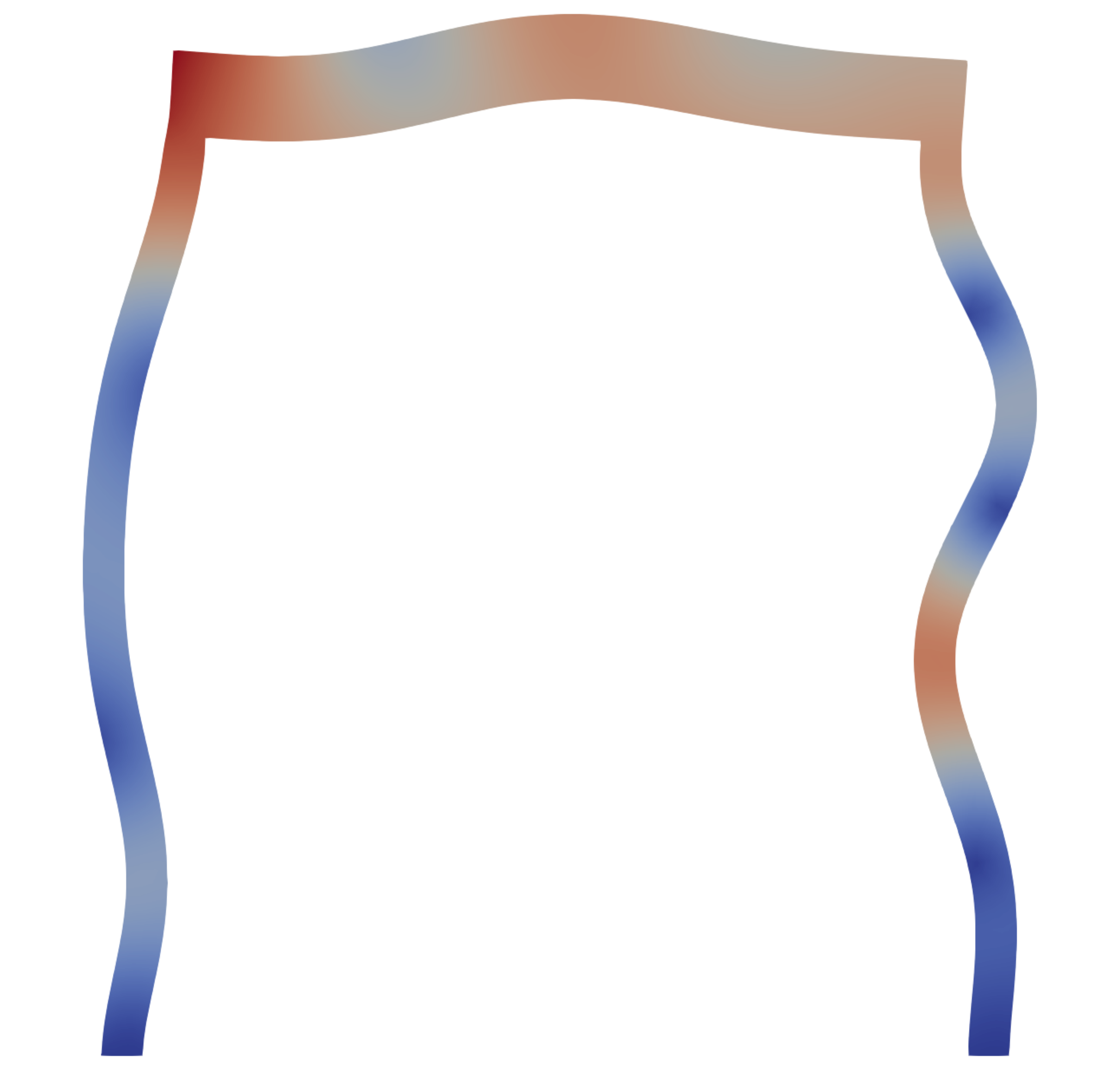}}$~$
\subfloat[[$7^{\circ}$ POD basis.][\label{fig:fig20}]{\includegraphics[scale=0.07]{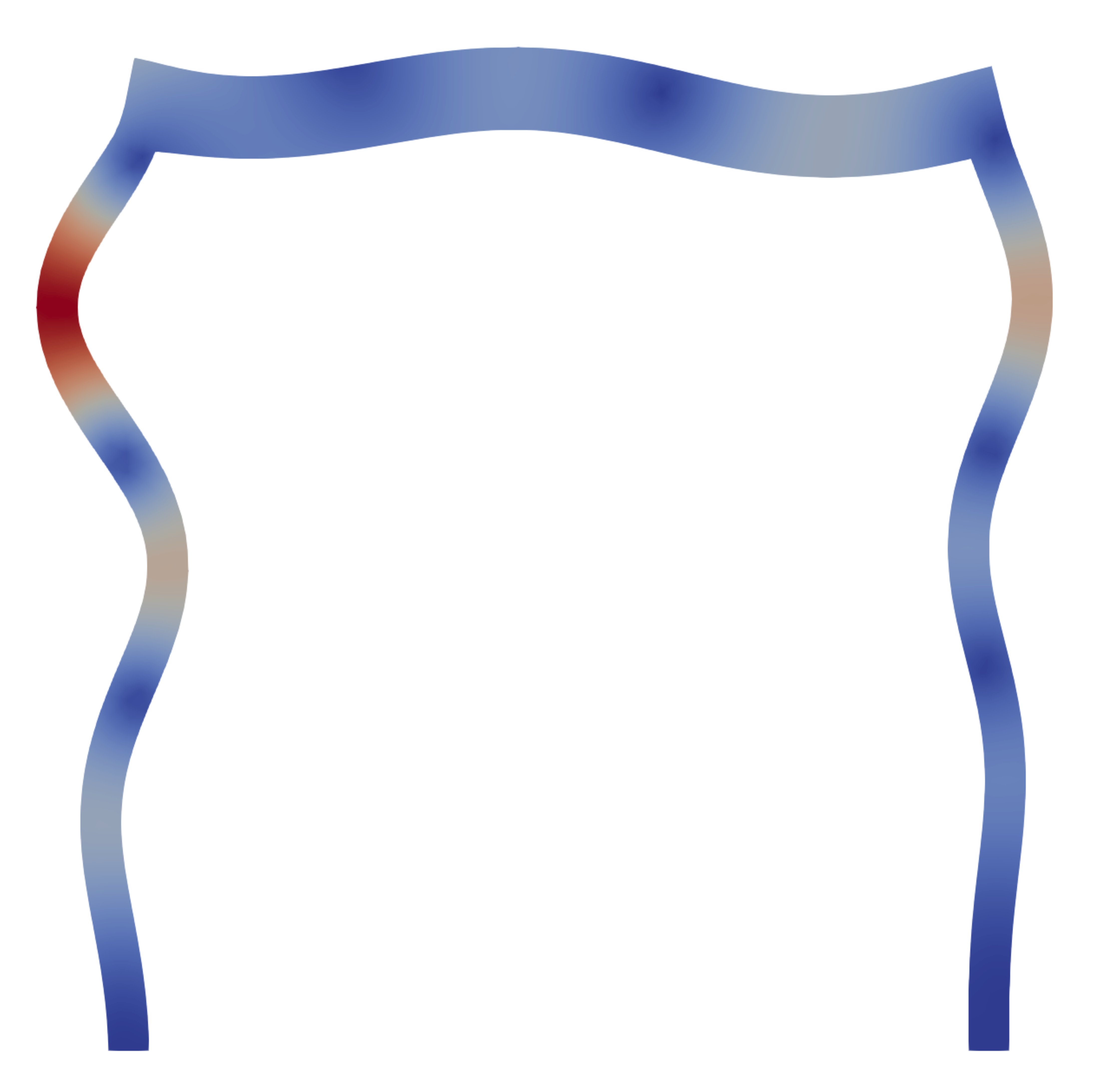}}$~$
\subfloat[[$8^{\circ}$ POD basis.][\label{fig:fig21}]{\includegraphics[scale=0.07]{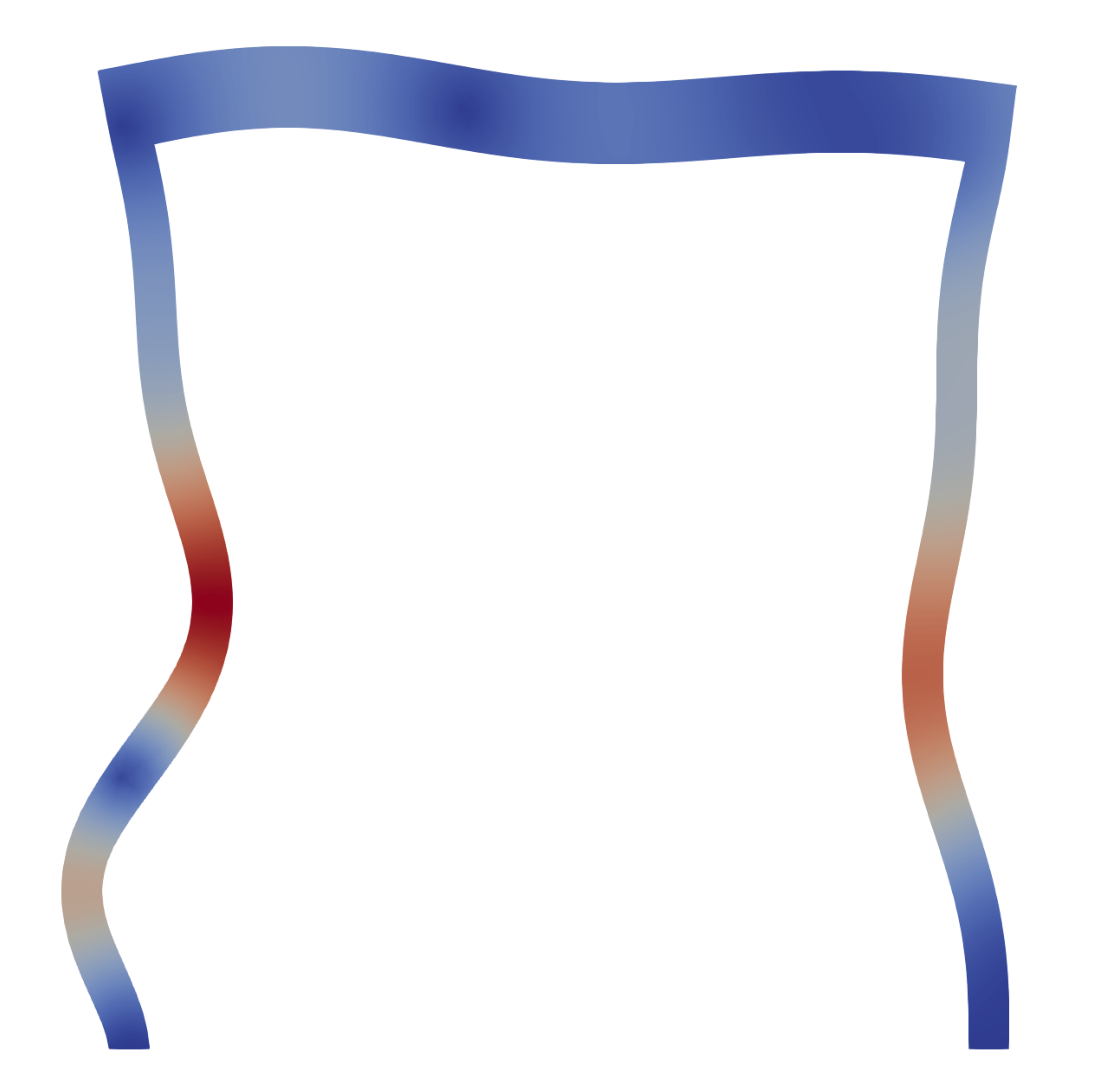}}\\
\caption{\small Portal frame: POD bases.\label{fig:tr_vld_1_error_vs_frequency_A}}
%\end{framed}
\end{figure}

The ROM has been built according to the procedure described in Algorithm \eqref{al:POD_param_elastodyn}. In \mbox{Figs. \ref{fig:fig12}-\ref{fig:fig13}}, the normalized singular values $\sigma^{T}_s/\sigma^{T}_1$ and $\sigma^{g\boldsymbol{\eta}}_s /\sigma^{g\boldsymbol{\eta}}_1$, with their typical descendant behavior, are reported for the POD in time and over the parametric space, respectively. The dashed horizontal line in the graph refers to the last selected POD basis, thus the correspondent abscissa identifies the overall number of selected POD bases ensuring a reconstruction error below the prescribed error tolerance $\varepsilon<\varepsilon_{tol}=10^{-4}$. An excellent approximation capacity has been achieved by relying upon $59$ POD bases only, instead of the original $1884$ dofs. In Fig. \ref{fig:tr_vld_1_error_vs_frequency_A}  the first 8 POD bases are reported:  higher order POD bases feature more complex shapes, useful to simulate the effect of localized damages. Due to the mentioned reduction in the number of the dofs from the FOM to the ROM, the computing time required by each simulation has decreased from $45 \text{ s}$ to $1.5 \text{ s}$, with a speedup of $30$ (computations have been run on a PC featuring an Intel (R) Core\texttrademark, i5 CPU @ 2.6 GHz and 8 GB RAM).

\begin{figure}[h!]
\centering
\includegraphics[width=0.7\textwidth]{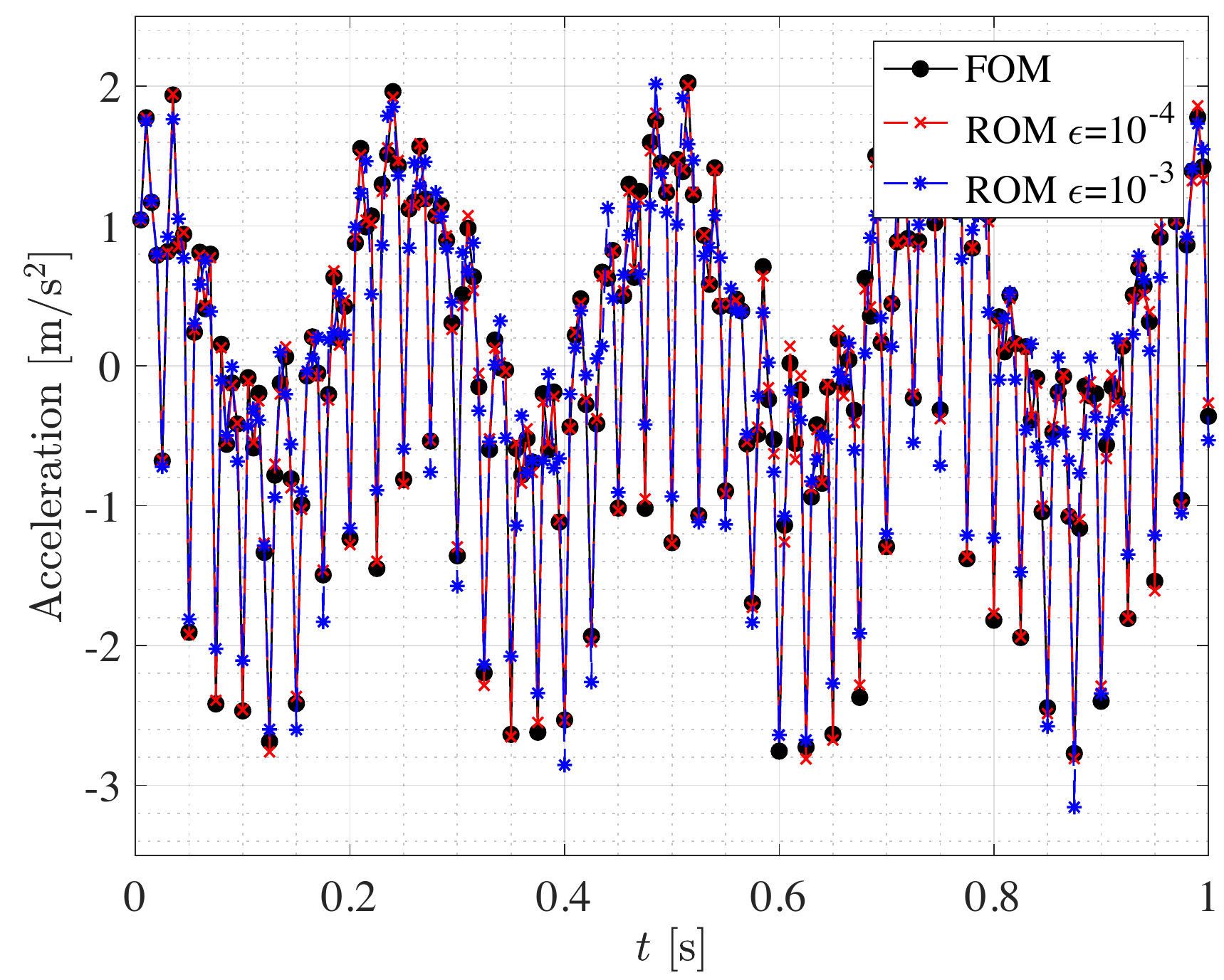}
\captionsetup{width=.8\linewidth}
\caption{{Portal frame: comparison between acceleration time histories generated through the FOM and the ROM for different values of  $\varepsilon_{tol}$.}\label{fig:fig22}}
\end{figure}

Since acceleration measurements have been exploited for SHM purposes, we have kept $\varepsilon_{tol}$ small also considering that  $\varepsilon$ has been based on the reconstruction of the displacement field.  As an example, in Fig. \ref{fig:fig22} the FOM and ROM  time histories are compared for the (unobserved) horizontal acceleration at the top-right corner of the frame, for different values of $\varepsilon_{tol}$ and for $g=0$ (undamaged state), $A=30 \text{ kPa}$, $f=80 \text{ Hz}$, $\delta=0$. The ROM solution relevant to $\varepsilon_{tol}=10^{-3}$ has turned out to be excessively inaccurate, so that $\varepsilon_{tol}=10^{-4}$ has been adopted for the construction of $\mathbf{D}$. 

	\bigskip
	
\subsection{Portal frame - Classification outcomes}
According to the NN hyperparameters setting discussed in Sec. \ref{sec:FCN}, $\mathcal{G}$ has been trained and validated on $I_{tr}+I_{val}=10,000$ instances with ratio $75:25$, for $500$ epochs. A further discussion on the employed number of filters $N_1$, $N_2$ and $N_3$ will be provided in the following.

\begin{figure}[H]
\centering
\begin{minipage}[c]{0.4\textwidth}
\centering
\includegraphics[width=1.1\textwidth]{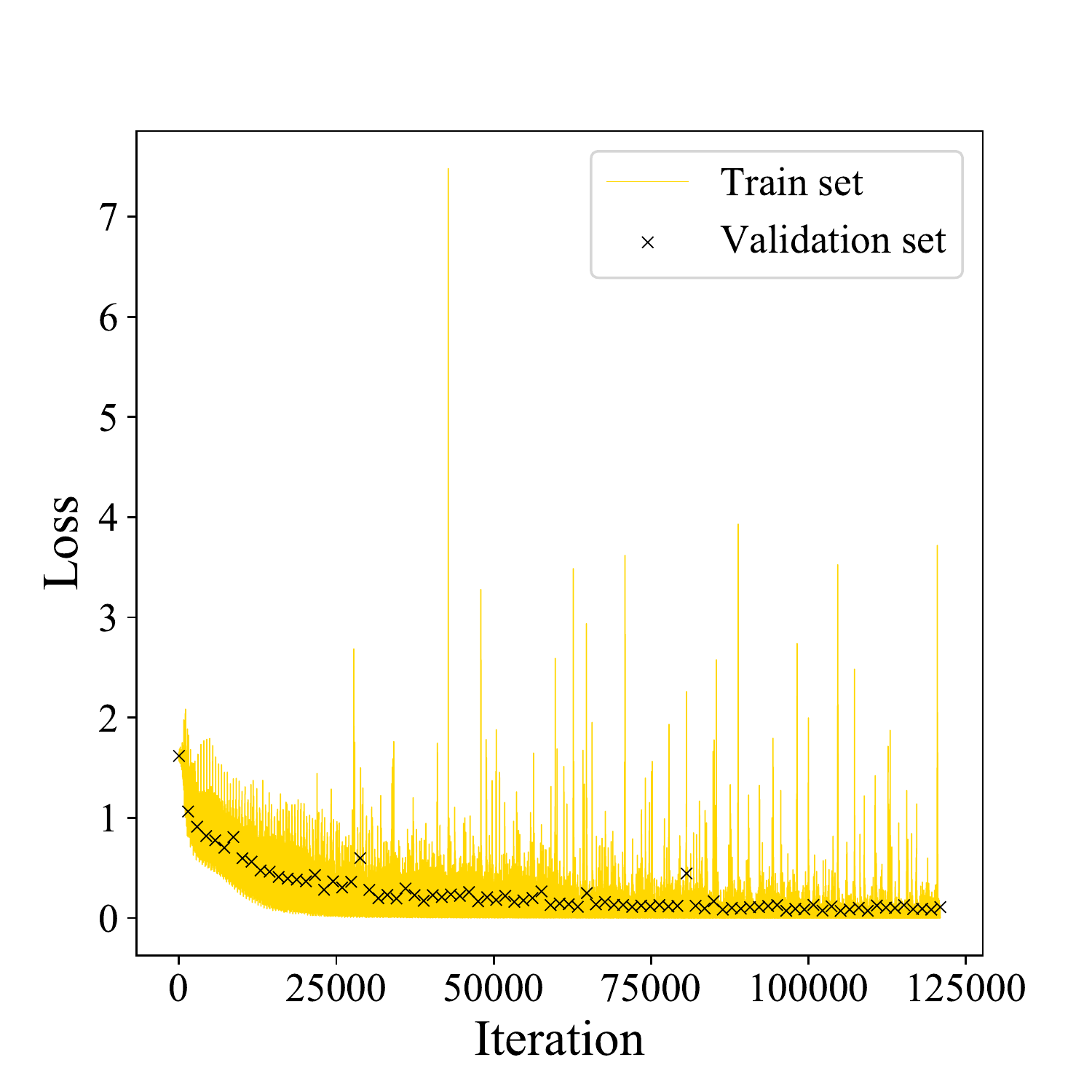}
\captionsetup{width=\linewidth}
\vspace{-0.8cm}
\caption{{Portal frame: FCN training. Loss function evolution on the training and validation sets for the noise-free case.}\label{fig:fig23}}
\end{minipage}
\hspace{1.cm}
\begin{minipage}[c]{0.4\textwidth}
\centering
\includegraphics[width=1.1\textwidth]{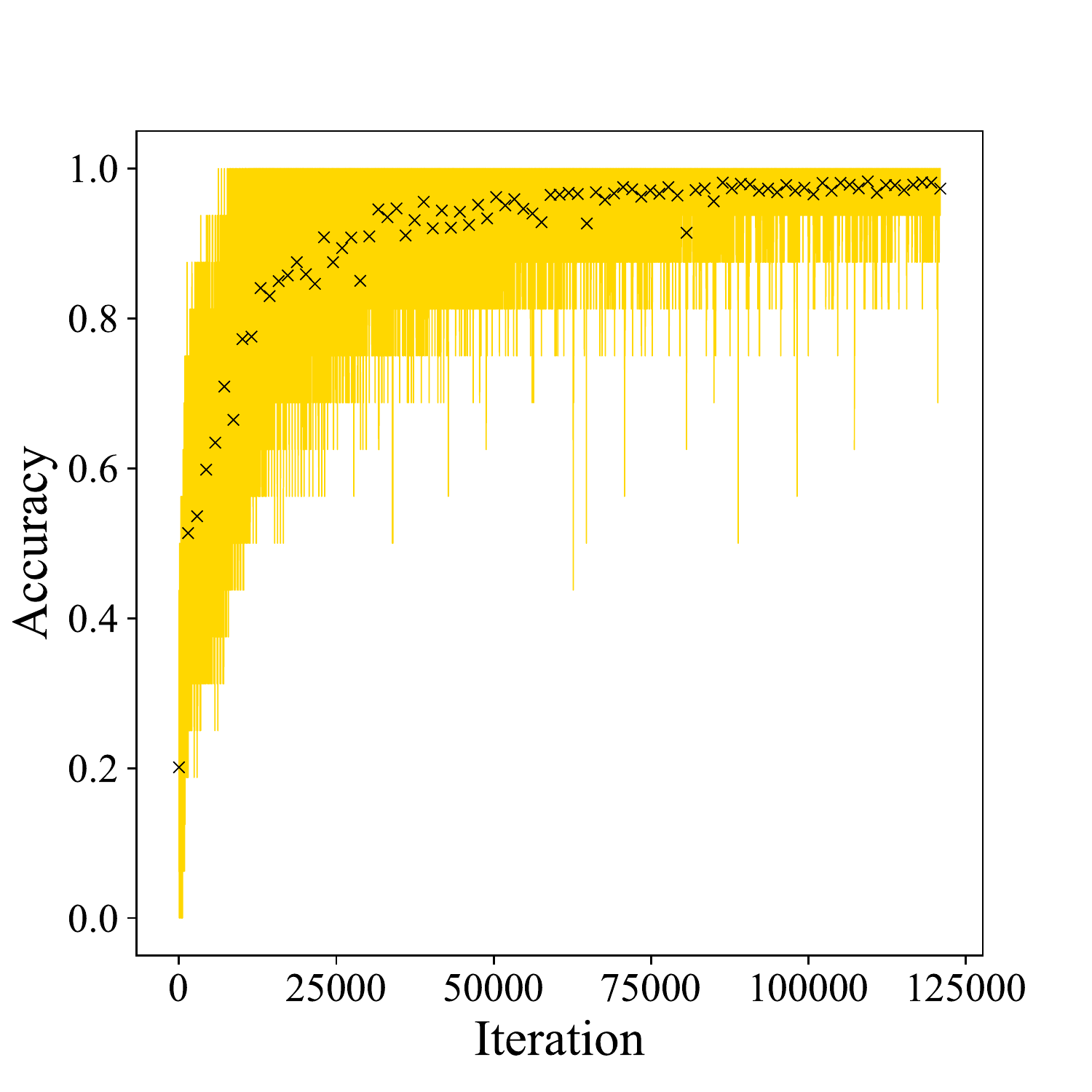}
\captionsetup{width=1\linewidth}
\vspace{-0.8cm}
\caption{{Portal frame: FCN training. Accuracy function evolution on the training and validation sets for the noise-free case.}\label{fig:fig24}}
\end{minipage}
\end{figure}

For the noise-free case, Figs. \ref{fig:fig23} and \ref{fig:fig24} show the evolutions of the loss and of the accuracy  during training, respectively. The iteration number, determined by the number of epochs and by the dimension of the mini-batches, accounts for the number of times the NN weights are modified during the training process. As expected, the first portion of training shows the highest gains in terms of classification accuracy. The spikes, observable both in the loss and accuracy graphs, are due to the different classification performances obtained on different mini-batches. At the end of the training, we have obtained  classification accuracies of  $98.91\%$ and $97.68\%$, respectively for the training and validation sets.

The generalization capabilities of $\mathcal{G}$ have been evaluated against a test set made of $200$ pseudo-experimental instances generated through the FOM. The classification test results are summarized by the confusion matrix in Fig. \ref{fig:fig25}. Even for this test, no noise effects have been taken into account. The classification task has been carried out with a global accuracy of $84.5\%$, but two different sources of error can be highlighted. First, $40\%$ of test instances featuring the scenario $g=2$ have been misclassified as $g=1$; with a smaller frequency, $22.5\%$, the same happens for $g=4$ and $g=3$. This is due to the similar influence of those damage scenarios on the structural response, by virtue of the structural layout and of the applied load. Second, $15\%$ of test instances featuring $g=0$ have been misclassified as well. This is due to the variability of $\delta$, as smaller values of $\delta$ provide the additional issues discussed before: an increased difficulty to distinguish a damaged scenario from an undamaged one; a major difficulty for the ROM to properly describe the damaged scenarios. The other way around, none of the damaged scenarios has been misclassified as undamaged. These effects of $\delta$ have been observed for a relevant number of tests, carried out with the same classifier, by processing different test sets characterized by a fixed damage level $\delta$; corresponding outcomes are summarized in Tab. \ref{tab:TB_5}.

\begin{minipage}[c]{\textwidth}
\centering
\begin{minipage}{0.4\textwidth}

\begin{figure}[H]
\includegraphics[width=\textwidth]{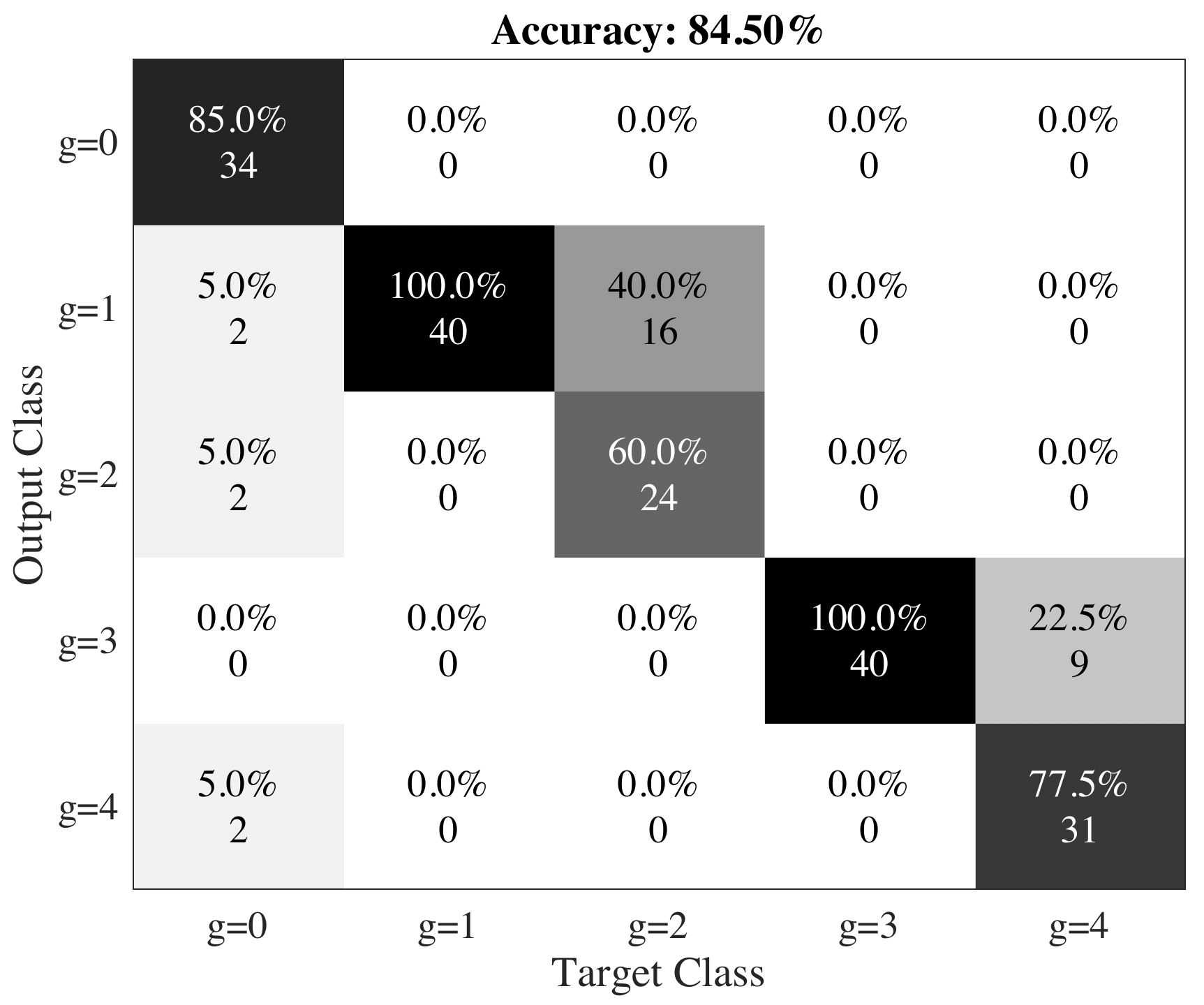}
\captionsetup{width=1\linewidth}
\caption{{Portal frame: testing of the FCN. Confusion matrix for the noise-free case.}\label{fig:fig25}}
\end{figure}
\end{minipage}
\hspace{2cm}
\begin{minipage}{0.26\textwidth}
 \begin{table}[H]
      \centering
      \footnotesize
       \[
		\begin{array}{cc}
		\toprule
	\delta \left[\%\right] & \text{Accuracy}\left[\%\right] \\
	\midrule
		25 & 66 \\
		20 & 78 \\
		15 & 94 \\
		10 & 94 \\
		5 & 78 \\
		2 & 60 \\
		\bottomrule
		\end{array}
		\]
\captionsetup{width=1.2\linewidth}
\caption{{Portal frame: classification accuracy in recognizing structural states characterized by a fixed damage level $\delta$ for the noise-free case.}\label{tab:TB_5}}
\end{table}
\end{minipage}
\hspace{1,2cm}
\end{minipage}\vspace{0.6cm}

To assess the impact of the noise level on the performance of $\mathcal{G}$, the classifier has been next trained and tested on datasets featuring $\text{SNR}=100,50$ and $20$. The chosen SNR values are representative of the self-noise of accelerometers usually employed in the monitoring of civil structures \cite{manual:Safran,manual:ST}. A certain SNR, indicator of the noise level affecting the vibration signals, can be obtained by adding a white noise to the synthetic structural recordings. Despite the relative simplicity of this procedure, the white noise allows to accurately mimic the signal perturbation affecting the measurements of real-life sensors, among which also micro-electro mechanical accelerometers \cite{proc:SN_Alessandro,art:SN_Evans}.

To get a more clear picture on how SNR affects the classification accuracy together with the ROM complexity, three ROMs, each of them featuring a specific $\varepsilon_{tol} = 10^{-3}$, $10^{-4}$ and $10^{-5}$, have been used for the generation of $\mathbf{D}$. The classification outcomes are reported in Tab. \ref{tab:acc_SNR_eps} for two different choices of the NN filters, $N_1=8$, $N_2=16$ and $N_3=8$ in Tab.\ref{tab:filters1_SNR_eps} and $N_1=16$, $N_2=32$ and $N_3=16$ in Tab. \ref{tab:filters2_SNR_eps}, respectively. The performance of $\mathcal{G}$ has been evaluated against the aforementioned three FOM test sets, with the results obtained after training the classifier for each one of the nine combinations of $\varepsilon_{tol}$ and SNR levels. From the results in Tab. \ref{tab:acc_SNR_eps}, it can be observed that both $\varepsilon_{tol}$ and SNR have a relevant impact on the accuracy of $\mathcal{G}$. Lower performances follow lower SNRs, especially when combined with large values of $\varepsilon_{tol}$. Nevertheless, by setting $\varepsilon_{tol}$ small enough (e.g. to $\varepsilon_{tol}=10^{-4}$), a global accuracy larger than $75 \%$ can be always attained. By enhancing the approximation capacity of the ROM, the classifier thus gets more robust against the noise.

\begin{figure}[h!]
\captionsetup[subfigure]{justification=centering}
\centering
\vspace{-0.2 cm}
\subfloat[SNR$=100$; ${\varepsilon}_{tol} =10^{-5}$\label{fig:Conf_ROM5_SNR100}]{\includegraphics[width=0.32\textwidth]{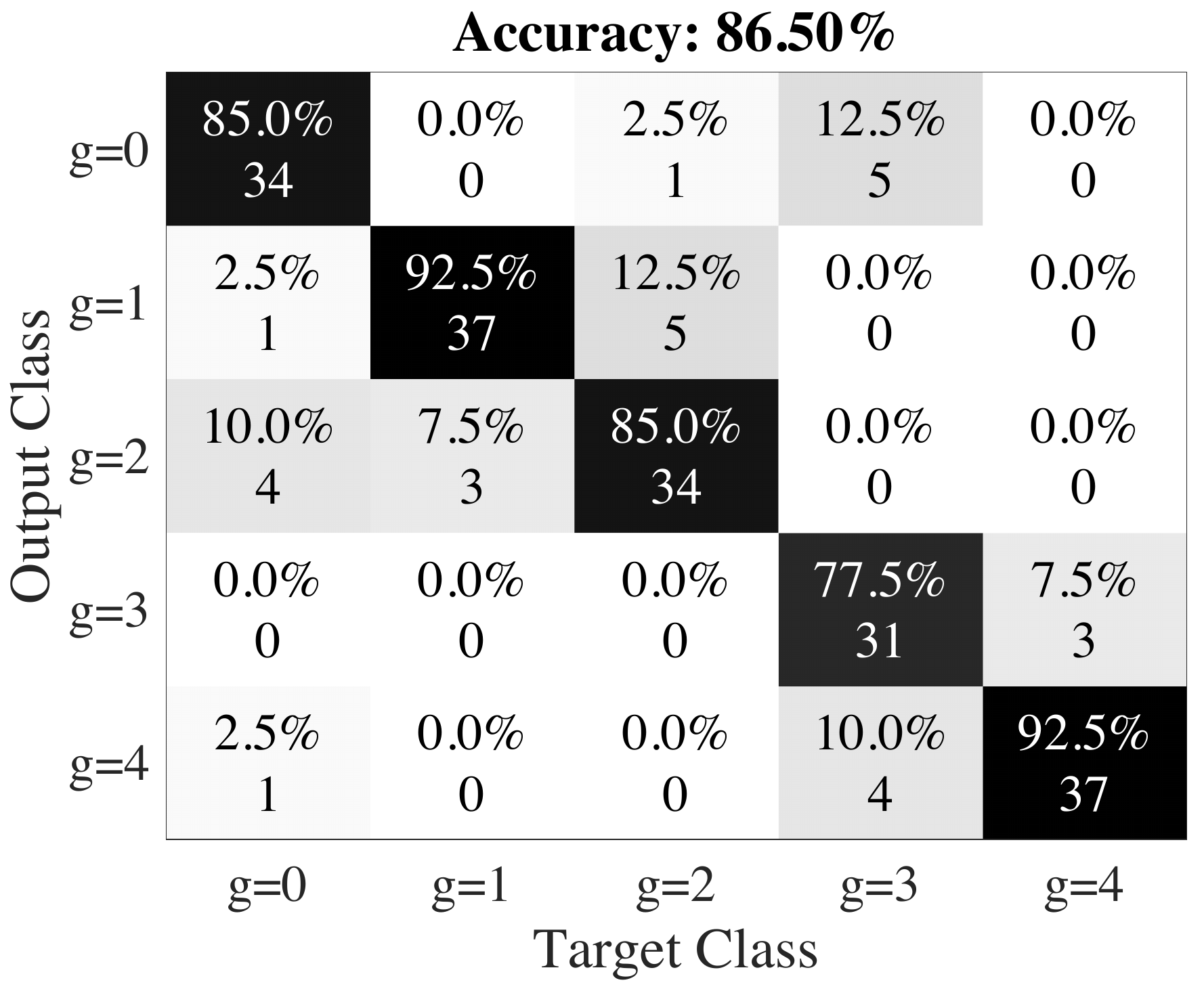}}
\subfloat[SNR$=50$; ${\varepsilon}_{tol} = 10^{-5}$\label{fig:Conf_ROM5_SNR50}]{\includegraphics[width=0.32\textwidth]{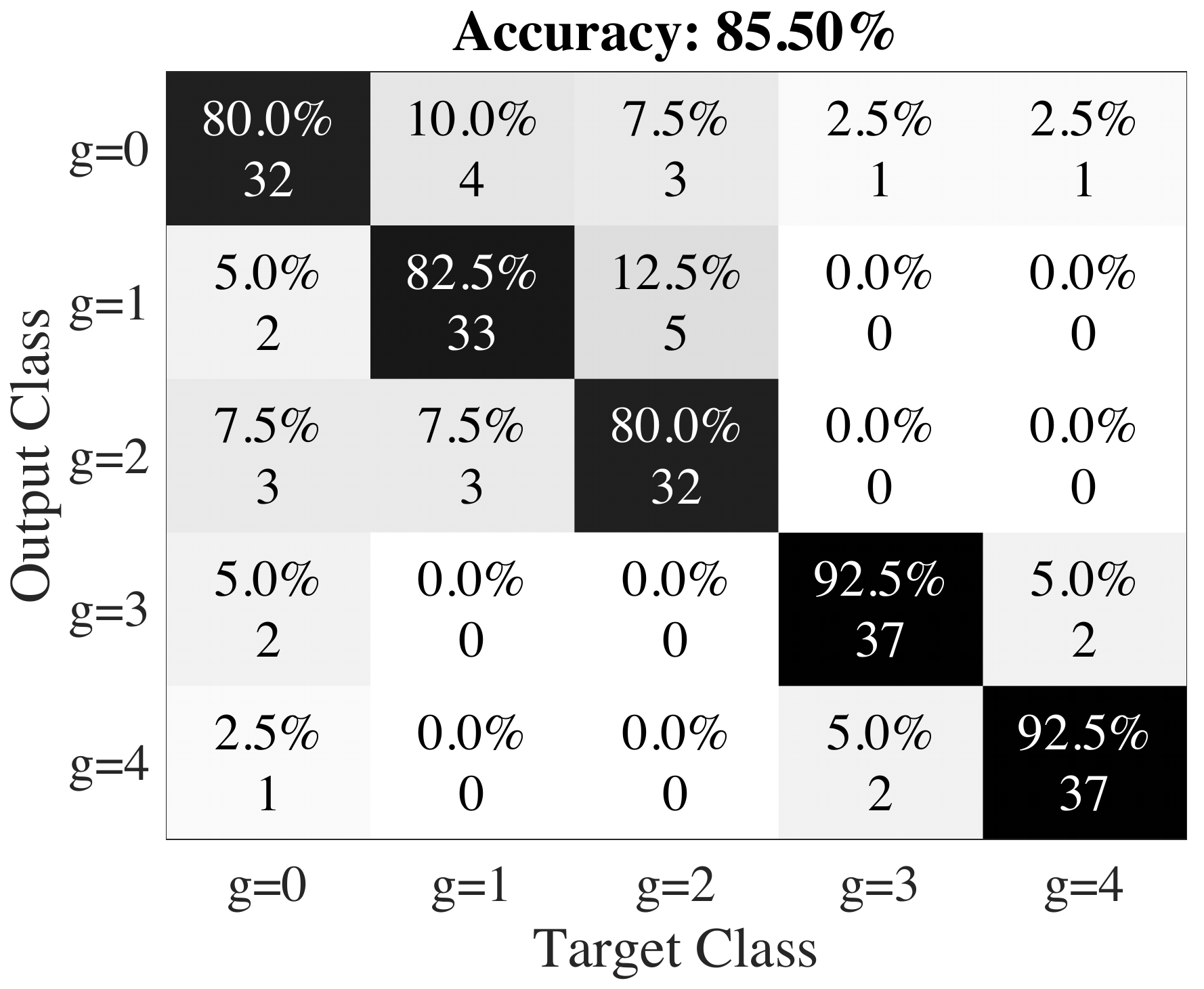}}
\subfloat[SNR$=20$; ${\varepsilon}_{tol} =10^{-5}$\label{fig:Conf_ROM5_SNR20}]{\includegraphics[width=0.32\textwidth]{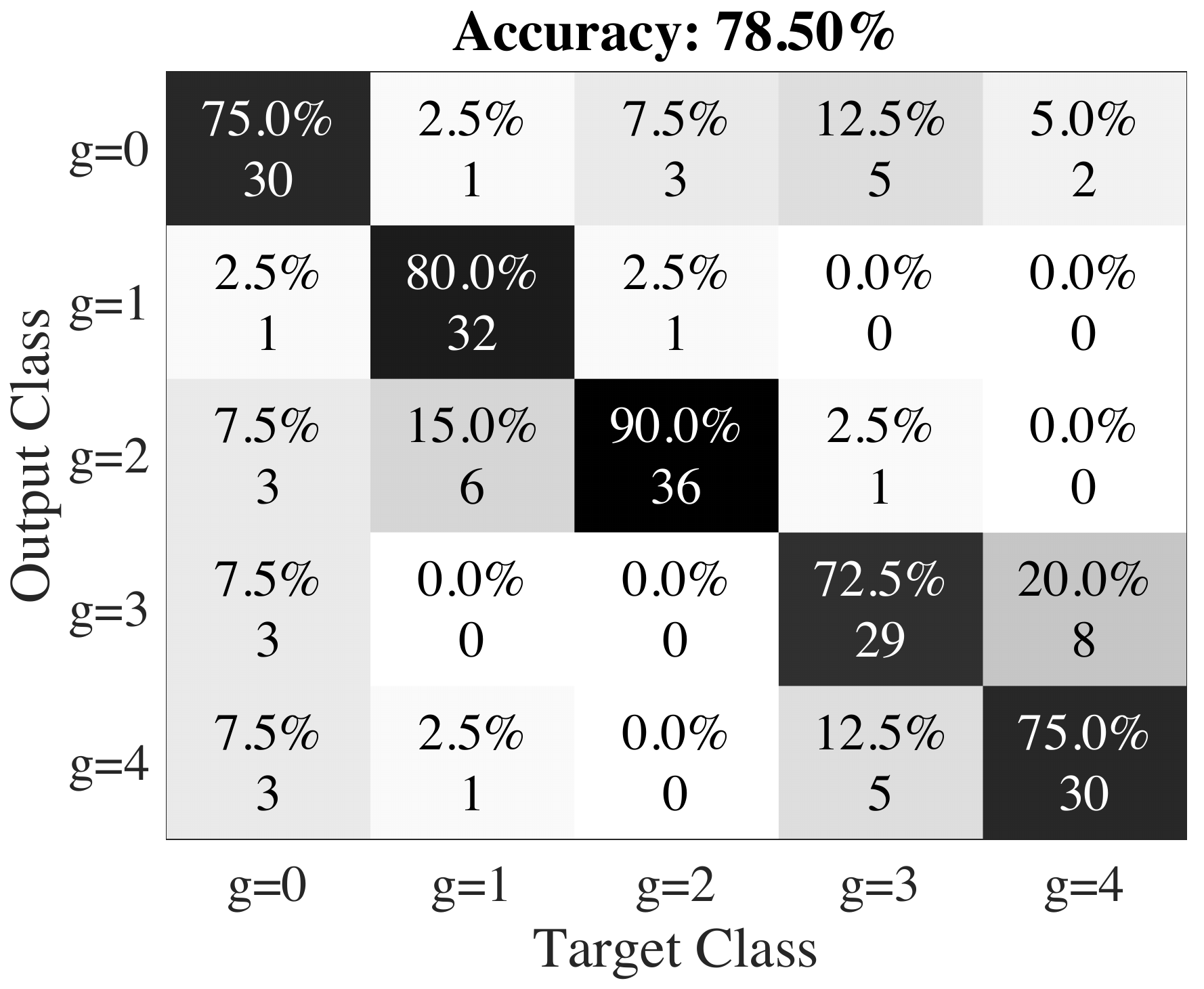}} \\
\vspace{-0.2 cm}
\subfloat[SNR$=100$; ${\varepsilon}_{tol} =10^{-4}$\label{fig:Conf_ROM4_SNR100}]{\includegraphics[width=0.32\textwidth]{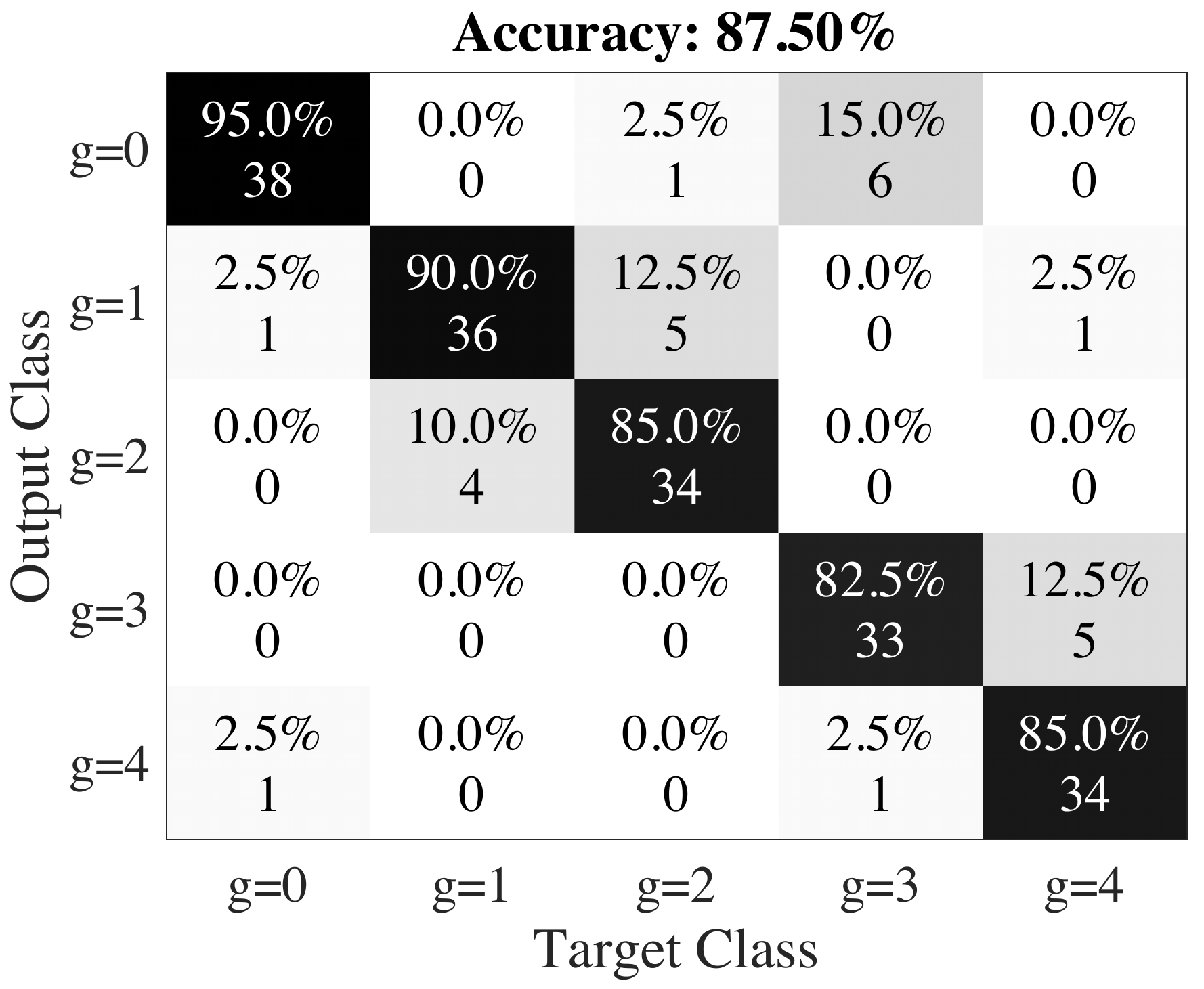}}
\subfloat[SNR$=50$; ${\varepsilon}_{tol} =10^{-4}$\label{fig:Conf_ROM4_SNR50}]{\includegraphics[width=0.32\textwidth]{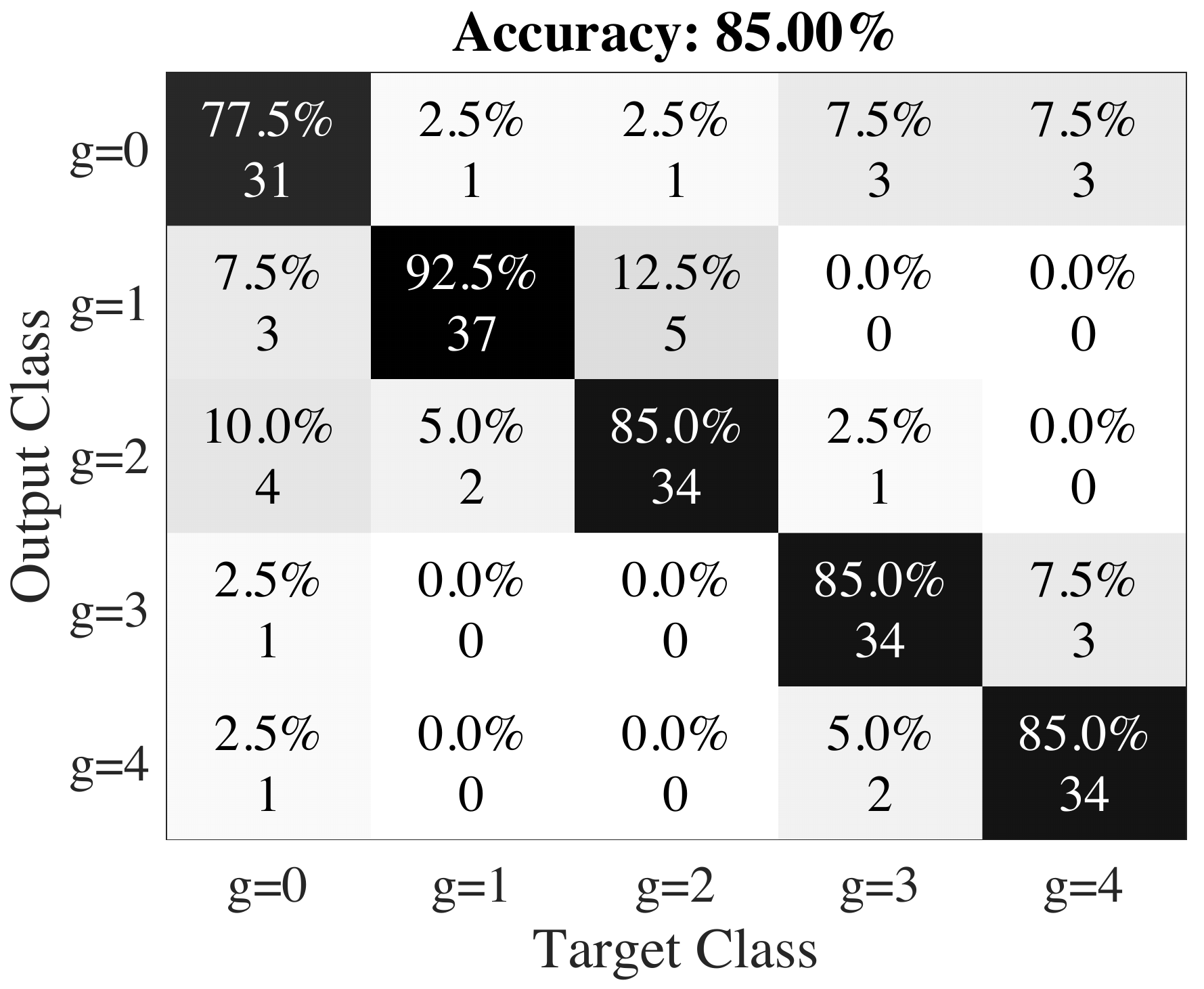}}
\subfloat[SNR$=20$; ${\varepsilon}_{tol} =10^{-4}$\label{fig:Conf_ROM4_SNR20}]{\includegraphics[width=0.32\textwidth]{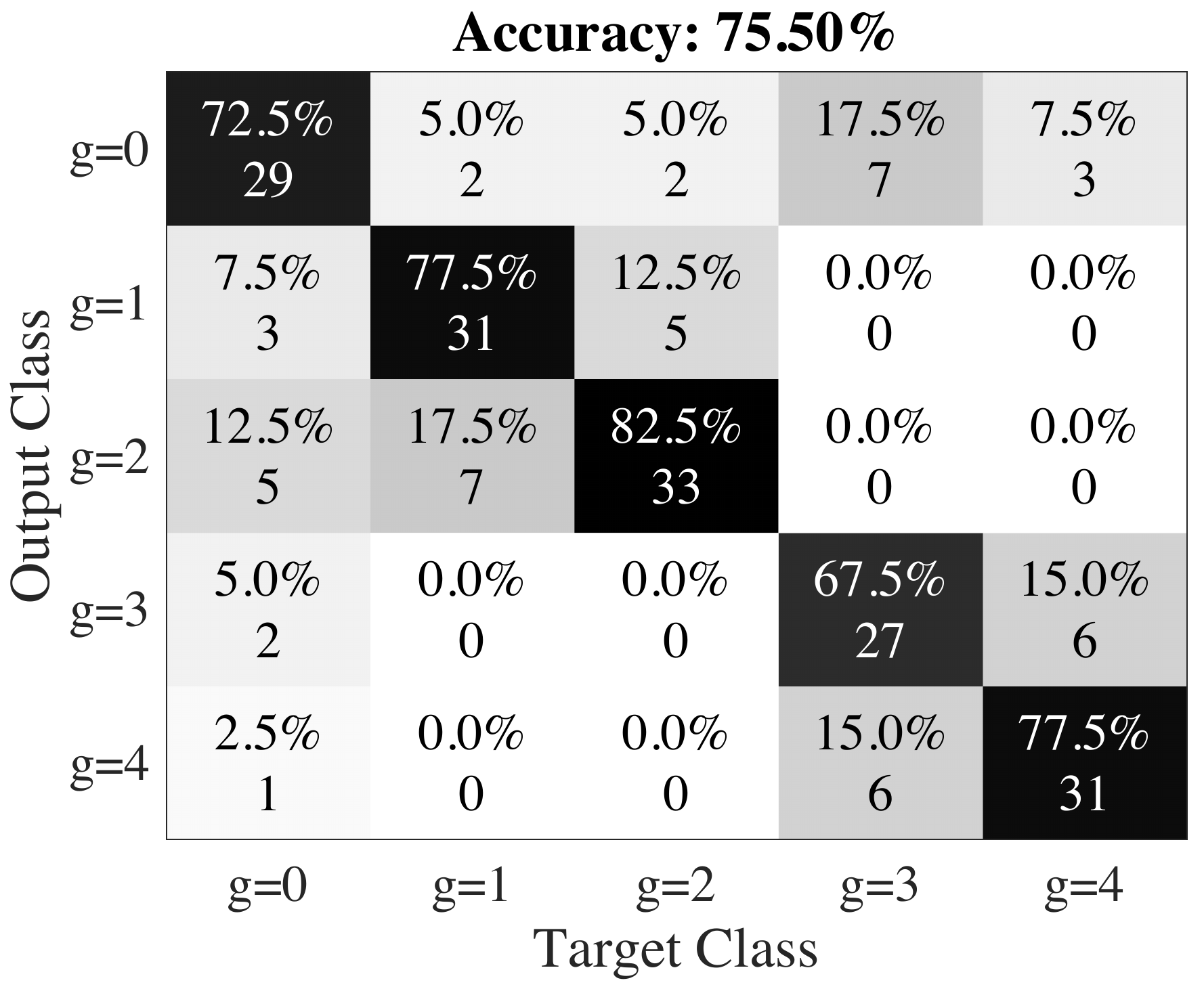}} \\
\vspace{-0.2 cm}
\subfloat[SNR$=100$; ${\varepsilon}_{tol} =10^{-3}$\label{fig:Conf_ROM3_SNR100}]{\includegraphics[width=0.32\textwidth]{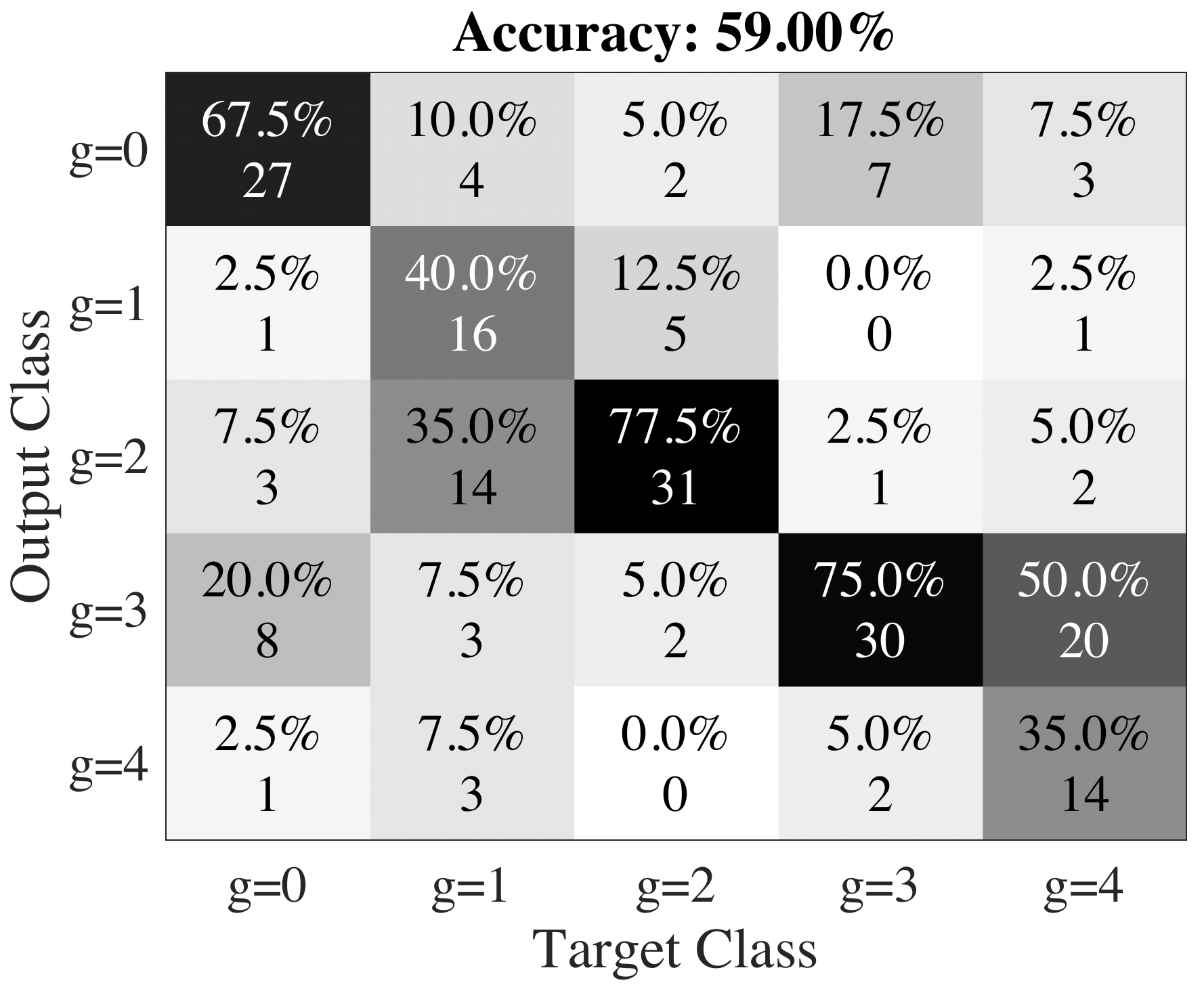}}
\subfloat[SNR$=50$; ${\varepsilon}_{tol} =10^{-3}$\label{fig:Conf_ROM3_SNR50}]{\includegraphics[width=0.32\textwidth]{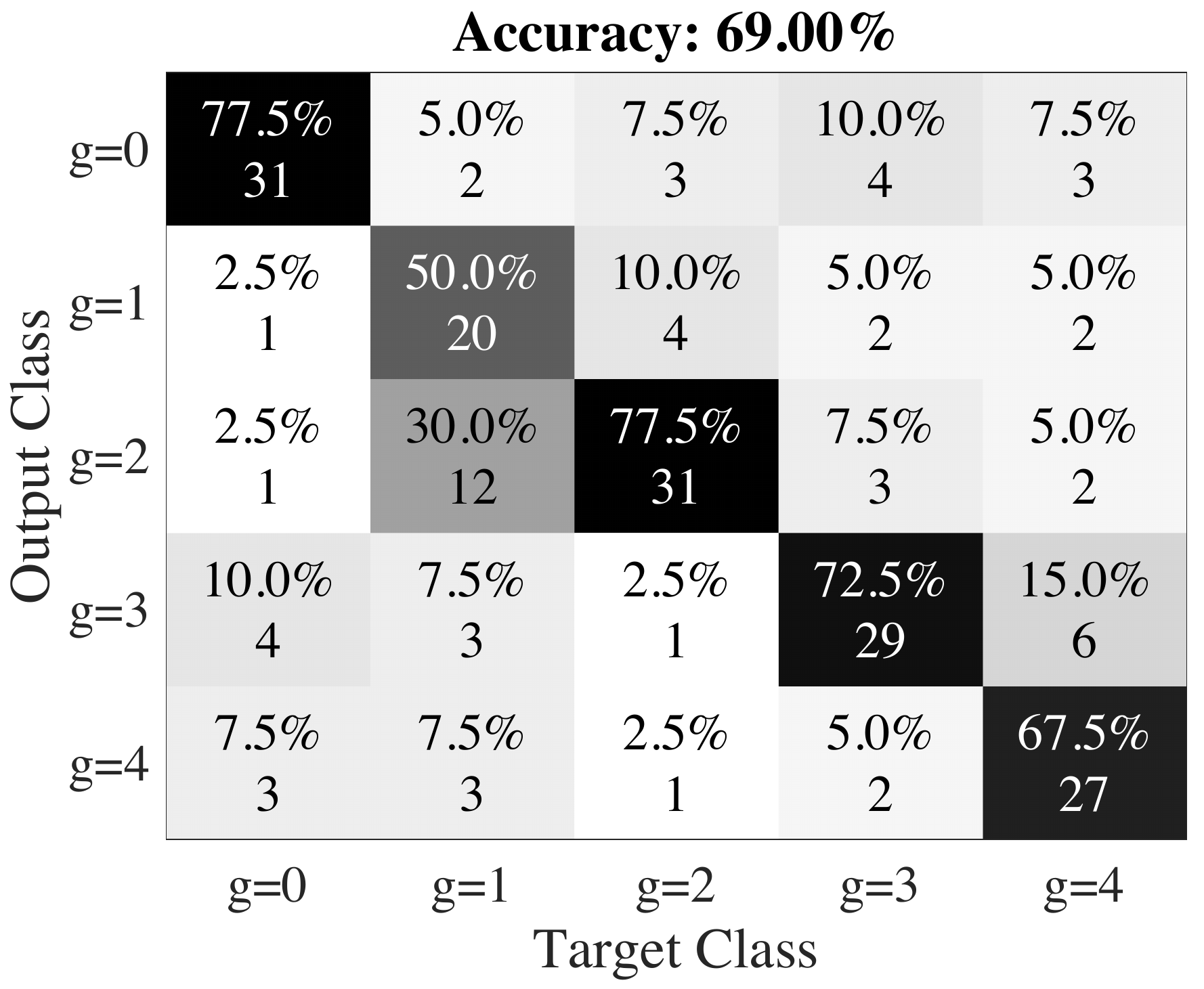}}
\subfloat[SNR$=20$; ${\varepsilon}_{tol} =10^{-3}$\label{fig:Conf_ROM3_SNR20}]{\includegraphics[width=0.32\textwidth]{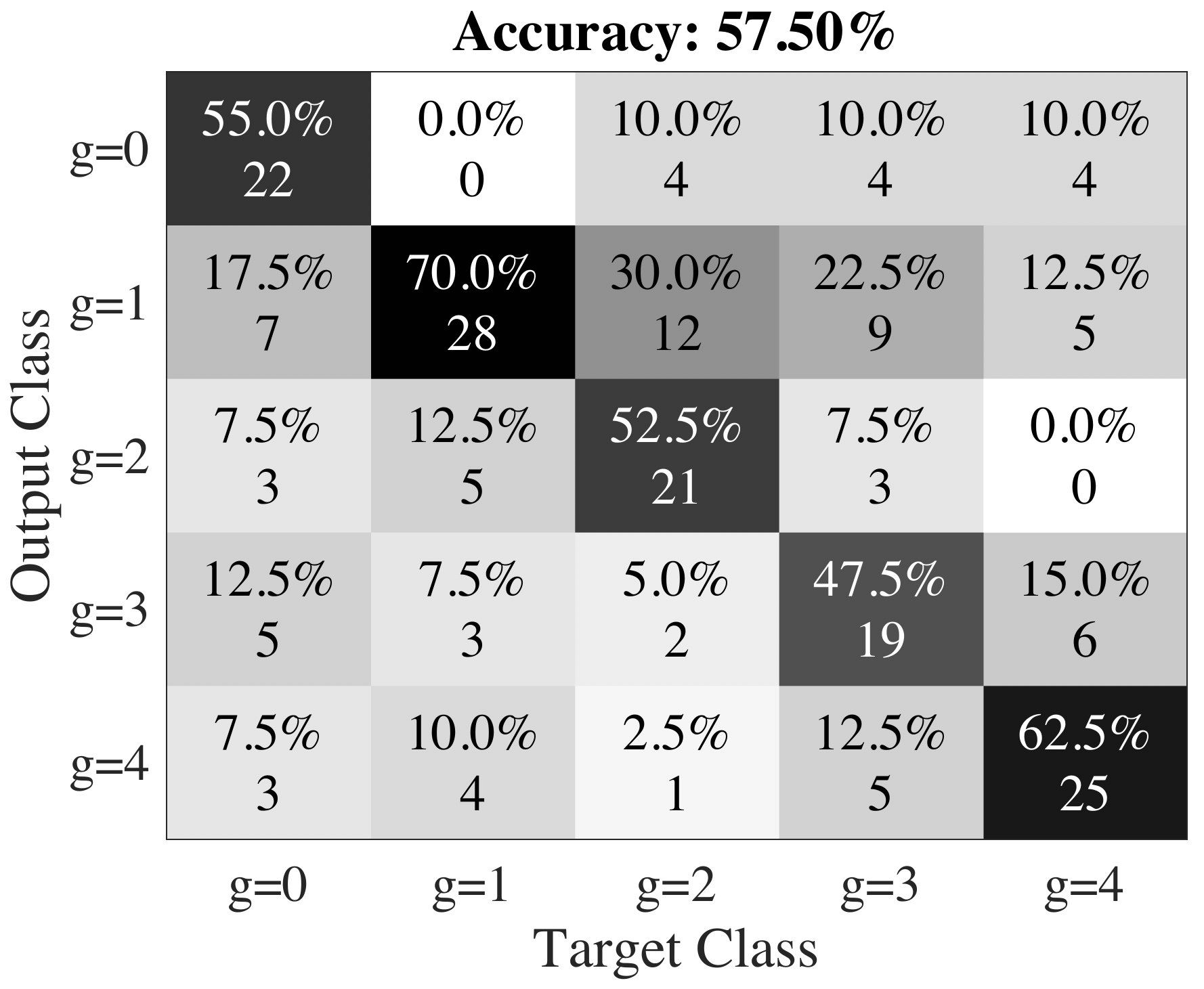}} \\
\smallskip
\caption{Portal frame: FCN testing. Confusion matrices at varying values of ${\varepsilon}_{tol}$ and SNR.\label{fig:conf_matrix_SNR_eps}}
\end{figure}

A further insight into the effects of $\varepsilon_{tol}$ and of SNR on the classification accuracy can be gained by looking at the confusion matrices reported in Fig. \ref{fig:conf_matrix_SNR_eps}. The two main sources of errors highlighted before, \textit{i.e.} the misclassification between $g=1$ and $g=2$, and between $g=3$ and $g=4$, are encountered even under different SNR levels. Smaller values of SNR worsen the classification performance, even without showing a clear path to penalize some specific damage scenarios.

\begin{table}[h!]
      \centering
      \footnotesize
       \[
		\begin{array}{c|ccc|ccc|ccc}
		\toprule
		\multicolumn{1}{c}{ } & \multicolumn{3}{c}{\textbf{training set}} & \multicolumn{3}{c}{\textbf{validation set}} & \multicolumn{3}{c}{\textbf{test set}}\\
		\midrule
{\varepsilon}_{tol} \setminus \textbf{SNR} & \mathbf{100} & \mathbf{50} & \mathbf{20} & \mathbf{100} & \mathbf{50} & \mathbf{20} & \mathbf{100} & \mathbf{50} & \mathbf{20} \\
	\midrule
		\mathbf{10^{-5}} &  87.9 \% & 86.3 \% & 79.3 \% &  81.0 \% & 80.7 \% & 72.2 \% &  79.5 \% & 79.5 \% & 71.5 \% \\
		\mathbf{10^{-4}} & 87.3 \% & 86.6 \% & 80.1 \% & 82.2 \% & 79.8 \% & 72.6 \% & 81.5 \% & 81.5 \% & 68.5 \% \\
		\mathbf{10^{-3}} & 90.0 \% & 88.0 \% & 84.2 \% & 85.2 \% & 82.2 \% & 76.0 \% & 64.0 \% & 59.5 \% & 62.5 \% \\
		\bottomrule
		\end{array}
		\]
		\vspace{-1.5\baselineskip}
		\subfloat[\label{tab:filters1_SNR_eps} $N_1=8$, $N_2=16$, $N_3=8$.]{\hspace{\linewidth}}
		\centering
      \footnotesize
       \[
		\begin{array}{c|ccc|ccc|ccc}
		\toprule
		\multicolumn{1}{c}{ } & \multicolumn{3}{c}{\textbf{training set}} & \multicolumn{3}{c}{\textbf{validation set}} & \multicolumn{3}{c}{\textbf{test set}}\\
		\midrule
{\varepsilon}_{tol}  \setminus \textbf{SNR} & \mathbf{100} & \mathbf{50} & \mathbf{20} & \mathbf{100} & \mathbf{50} & \mathbf{20} & \mathbf{100} & \mathbf{50} & \mathbf{20} \\
	\midrule
		\mathbf{10^{-5}} &  96.3 \% & 94.5 \% & 92.7 \% &  87.3 \% & 85.0 \% & 78.7 \% &  86.5 \% & 85.5 \% & 78.5 \% \\
		\mathbf{10^{-4}} & 96.1 \% & 94.9 \% & 93.3 \% & 87.6 \% & 84.2 \% & 79.5 \% & 87.5 \% & 85.0 \% & 75.5 \% \\
		\mathbf{10^{-3}} & 93.8 \% & 95.9 \% & 95.0 \% & 86.7 \% & 85.1 \% & 79.8 \% & 59.0 \% & 69.0 \% & 57.5 \% \\
		\bottomrule
		\end{array}
		\]
		\vspace{-1.5\baselineskip}
		\subfloat[\label{tab:filters2_SNR_eps} $N_1=16$, $N_2=32$, $N_3=16$.]{\hspace{\linewidth}}
		\captionsetup{width=.84\linewidth}
		\caption{Classification performance, in terms of global accuracy on the training, validation and test sets, as affected by the values of SNR and $\varepsilon_{tol}$.  \label{tab:acc_SNR_eps}} 
\end{table}	

Tab. \ref{tab:acc_SNR_eps} has been reported to clarify the criteria used to set the number of filters $N_1$, $N_2$ and $N_3$. By comparing the classification accuracy on the training and validation sets, the classifier employing $N_1=16$, $N_2=32$ and $N_3=16$ seems to slightly overfit the training data. Indeed, there is a disparity between the higher performance on the training set and the lower performance on the validation set. This tendency is even more evident when datasets featuring high SNR values are considered, due to the combined effect of the high number of NN weights and of the greater uninformative content induced by noise. The combination of these factors does not lead the NN to filter noise, because the NN has enough weights to keep memory of it. Similar outcomes are reported for the training and  test sets. By lowering the number of filters to $N_1=8$, $N_2=16$ and $N_3=8$, that is by halving the overall number of weights in $\mathbf{\Omega}$, it is possible to reduce this tendency; on the other hand, the accuracy of the NN is reduced. In light of this, we have chosen an architecture employing $N_1=16$, $N_2=32$ and $N_3=16$ to carry out the analysis, even if it  slightly overfits the data.

\begin{figure}[h!]
\begin{center}
\includegraphics[width=0.5\textwidth]{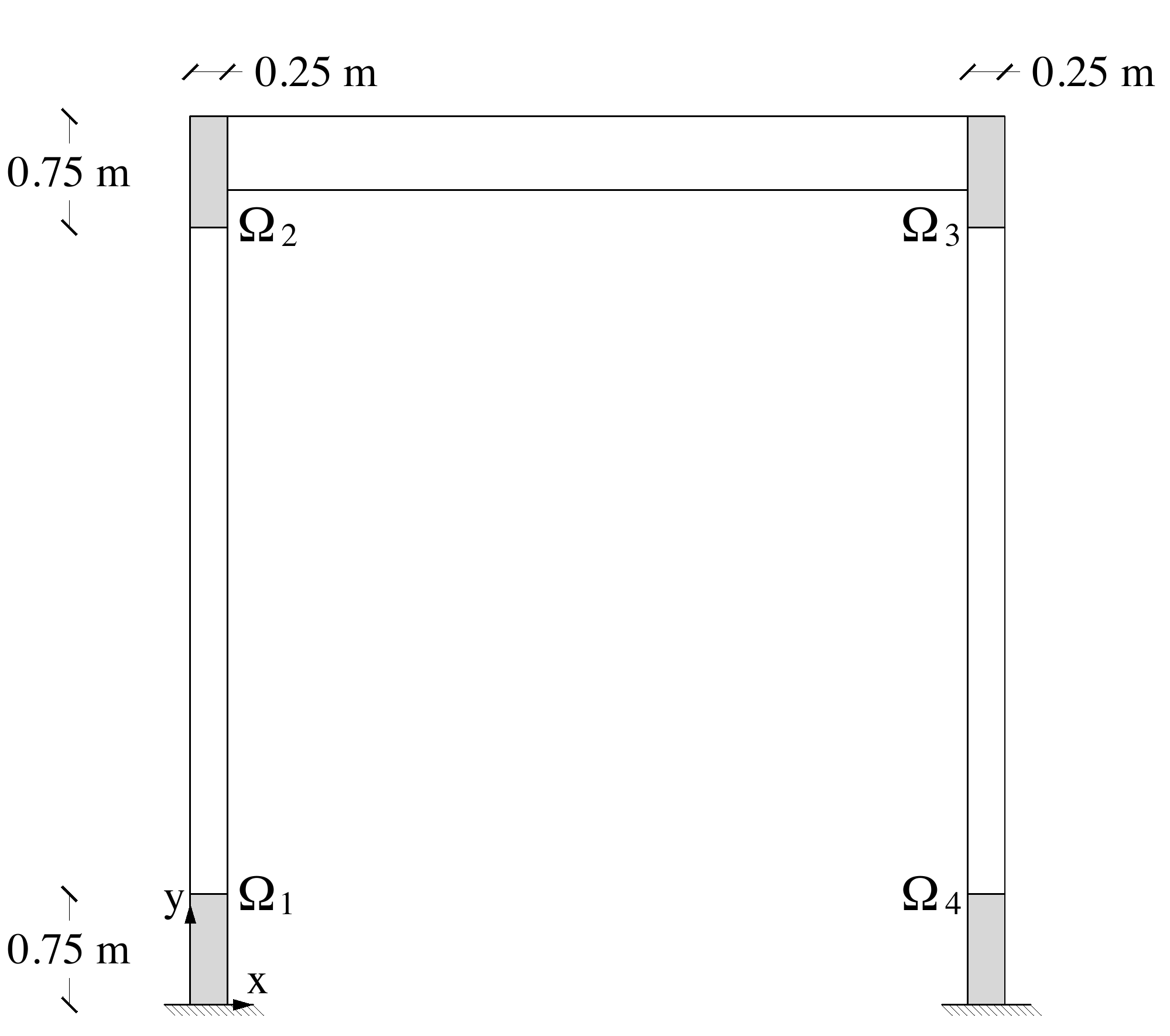}
\caption{{Portal frame: reduced-size damaged regions.}\label{fig:red_dam_scenarios}}
\end{center}
\end{figure}

\begin{figure}[h!]
\centering
\begin{minipage}{0.48\textwidth}
\begin{figure}[H]
\includegraphics[width=\textwidth]{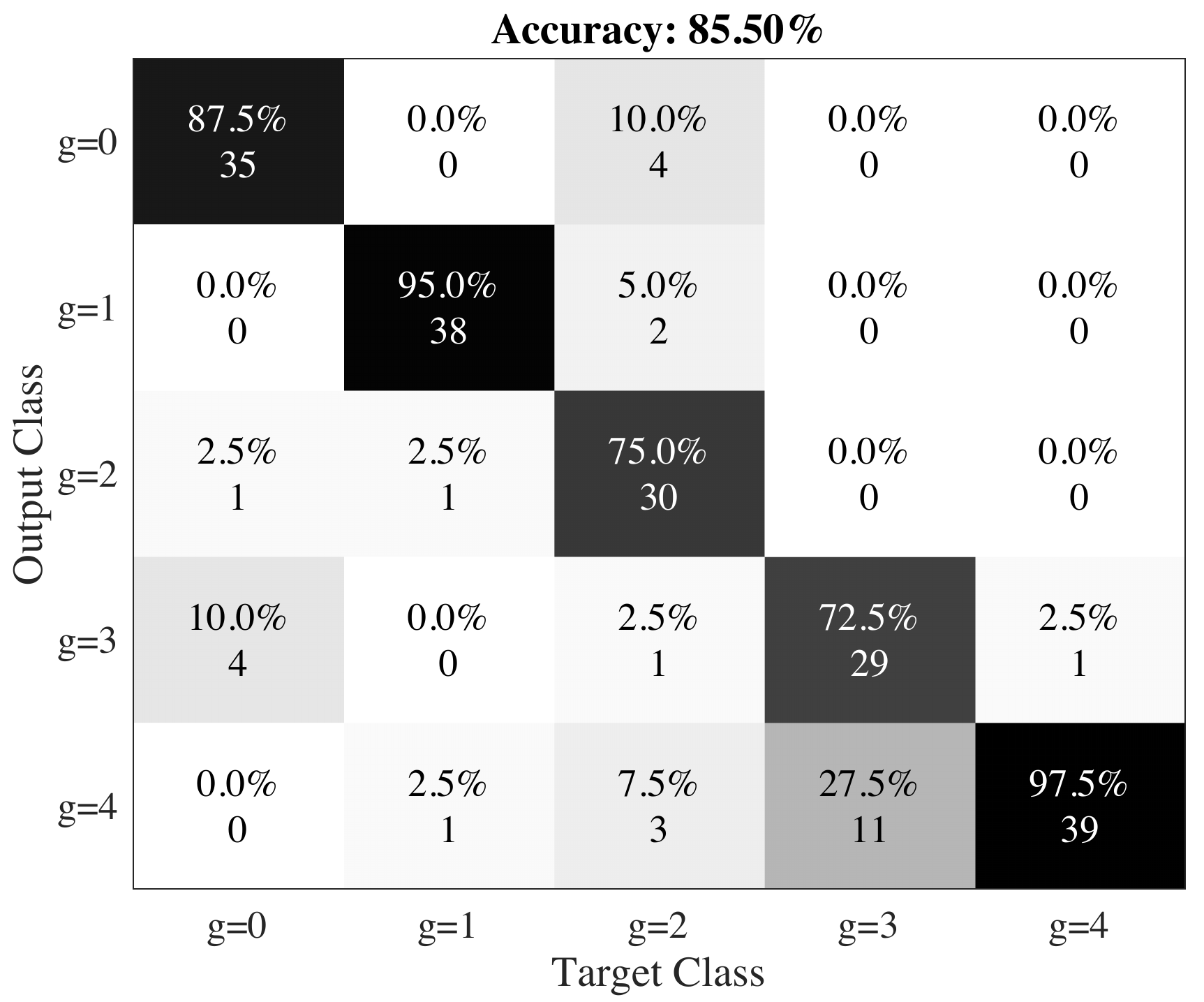}
\captionsetup{width=.9\linewidth}
\caption{{Portal frame: FCN testing. Confusion matrix for the reduced damage scenarios.}\label{fig:fig27}}
\end{figure}
\end{minipage}
\hspace{0.5cm}
\begin{minipage}{0.24\textwidth}
\begin{table}[H]
      \centering
      \footnotesize
       \[
		\begin{array}{cc}
		\toprule
	\delta \left[\%\right] & \text{Accuracy} \\
	\midrule
		25 & 74 \\
		20 & 84 \\
		15 & 92 \\
		10 & 66 \\
		5 & 56 \\
		2 & 42 \\
		\bottomrule
		\end{array}
		\]
\captionsetup{width=1.02\linewidth}
\caption{{Portal frame: FCN testing. Classification accuracy obtained in recognizing damaged regions reduced in size and characterized by a  damage level $\delta$ in the noise-free case.\label{tab:TB_6}}}
\end{table}
\end{minipage}
\hspace{1,2cm}
\end{figure}

To assess the robustness of the proposed methodology, the classifier $\mathcal{G}$ has been tested in recognizing structural states characterized by a stiffness reduction in subdomains different from those used to train it, see Fig. \ref{fig:fig1} and compare it with Fig. \ref{fig:red_dam_scenarios}: the damaged subdomains are approximately half in size of those used to construct $\mathbf{D}$. In this case, a noise-free condition has been considered. The pseudo-experimental instances used for testing have been generated via FOM. The obtained results, summarized by the confusion matrix in Fig. \ref{fig:fig27}, confirm the robustness of our methodology. The global accuracy is practically unchanged with respect to the previous case ($85.50\%$ vs $84.50\%$) and, furthermore, the sources of the misclassification error look almost the same. The only remarkable difference is linked to a  misclassification of the damaged scenarios as undamaged; this outcome is somehow expected since,  as shown in Tab. \ref{tab:TB_6}, the damaged subdomains reduced in size do have a smaller impact on the structural response to the given loadings.

%%%%%%%%%%%%%%%%%%%%%%%%%%%%%%%%%%%%%%%%%%
\section{Case study 2 - Railway bridge}
\label{sec:bridge}

\begin{figure}[h!]
\begin{center}
\includegraphics[width=1\textwidth]{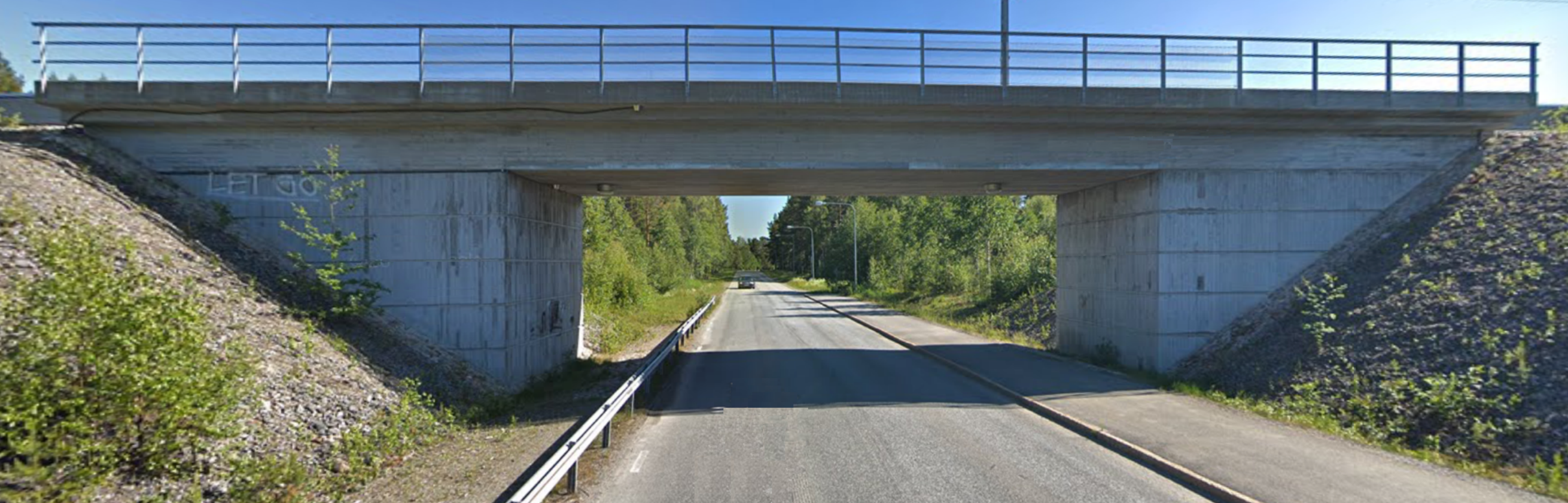}
\caption{{H{\"o}rnefors railway bridge.}\label{fig:fig28}}
\end{center}
\end{figure}

The second case study adopted to assess the performance of the proposed methodology, consists of an integral concrete portal frame railway bridge. Here, the effect of the sensor noise has been disregarded as the focus is on handling the effects of a dynamic moving load. The railway bridge, located along the Bothnia line in the urban area of H{\"o}rnefors in the northern Sweden, is depicted in Fig. \ref{fig:fig28}. The bridge has a span of $15.7 \text{ m}$, a free height of $4.7 \text{ m}$, a width of $5.9 \text{ m}$ (edge beams excluded) and it does not have any expansion joint or supporting device in between the deck and the abutments. The deck has a thickness of $0.5 \text{ m}$, whilst the frame walls have a thickness of $0.7 \text{ m}$; the wing walls, stretching out in longitudinal direction up to $8 \text{ m}$ at the top, have a thickness of $0.8 \text{ m}$. The foundation system consists in a couple of  plates connected by two stay beams and supported by pile groups. The bridge superstructure consists of a single ballasted track resting on sleepers spaced $0.65 \text{ m}$ apart, while the ballast layer is assumed to have a depth of $0.6 \text{ m}$  and a width of $4.3\text{ m}$.
\begin{figure}[h!]
\begin{center}
\includegraphics[width=0.95\textwidth]{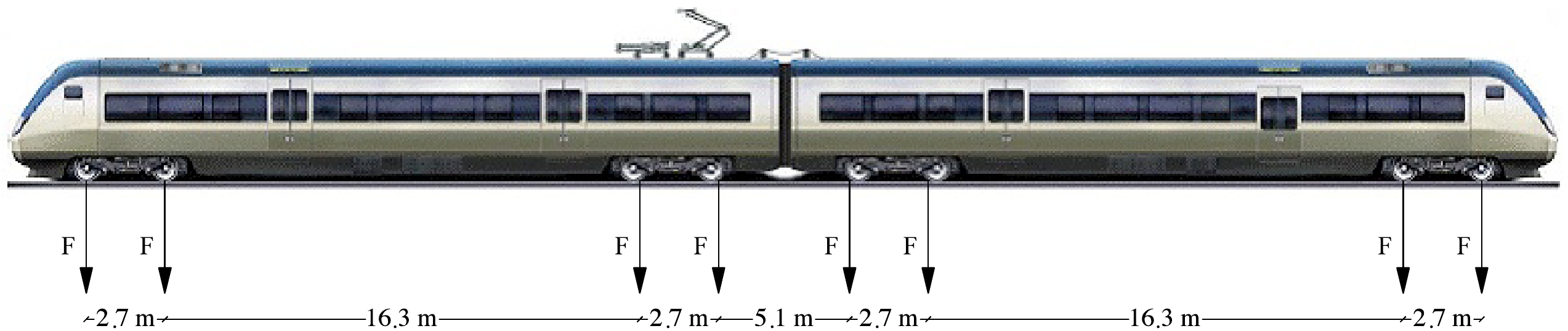}
\caption{{Gr{\"o}na T{\r a}get train type (adapted from \cite{thesis:kth2}).}\label{fig:fig29}}
\end{center}
\end{figure}   
\begin{figure}[h!]
\begin{center}
\includegraphics[width=0.72\textwidth]{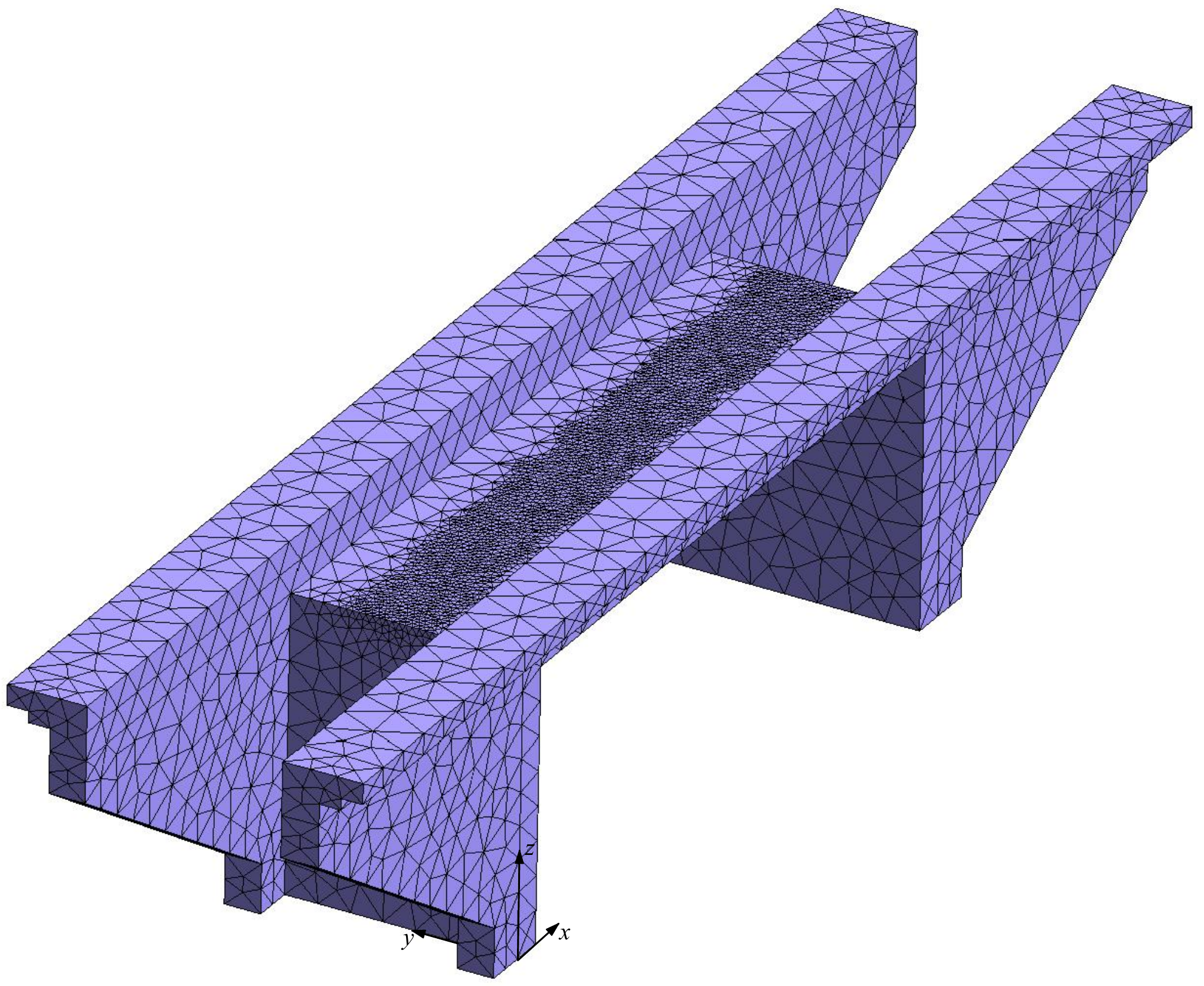}
\caption{{FE discretization of the H{\"o}rnefors railway bridge.}\label{fig:fig30}}
\end{center}
\end{figure}
        
The structure is loaded by the passage of trains of type Gr{\"o}na T{\r a}get (Fig. \ref{fig:fig29}), composed of two wagons, in transit with a speed $\gamma$ ranging between $160 \text{ km}/\text{h}$ and $215 \text{ km}/\text{h}$, and having in total $8$ axles. All the geometrical and mechanical data used to model the structure and the loads have been taken from \cite{thesis:kth2,thesis:kth3}, where the relevant soil-structure interaction was studied.

\subsection{Railway bridge - FOM}

The structure has been discretized with $15,075$ four node tetrahedral elements, as shown in Fig. \ref{fig:fig30}, with a total of $15,300$ dofs. To properly account for the acting loads, the characteristic size of the elements has been set to  $0.15 \text{ m}$ for the deck, while it has been set to  $0.80 \text{ m}$ elsewhere. The bridge has been assumed to be perfectly clamped at the bases. The adopted mechanical properties are those of a concrete class C35/45: $E=34 \text{ GPa}$, $\nu= 0.2$, $\rho=2500 \text{ kg}/\text{m}^{3}$. The ballast layer, whose density is $\rho=1800 \text{ kg}/\text{m}^{3}$, has been accounted for by modifying the density of  concrete of the deck and the edge beams, thus providing the additional mass resting on the deck. Time discretization has been performed by partitioning $(0,T)$ with subintervals of size  $2.5 \cdot 10^{-3} \text{ s}$, set to account for the maximum train speed of $215 \text{ km}/\text{h}$ with the element size of $0.15 \text{ m}$. The embankments have been modelled through distributed springs over the lateral surfaces in contact with the ground: this is equivalent to adopting a Robin mixed boundary condition (with elastic coefficient $a_{Robin}=10^{8} \text{ N}/\text{m}^{3}$) in the numerical model.

\begin{figure}[h!]
\centering
\begin{minipage}[b]{0.5\textwidth}
\centering
\includegraphics[width=0.85\textwidth]{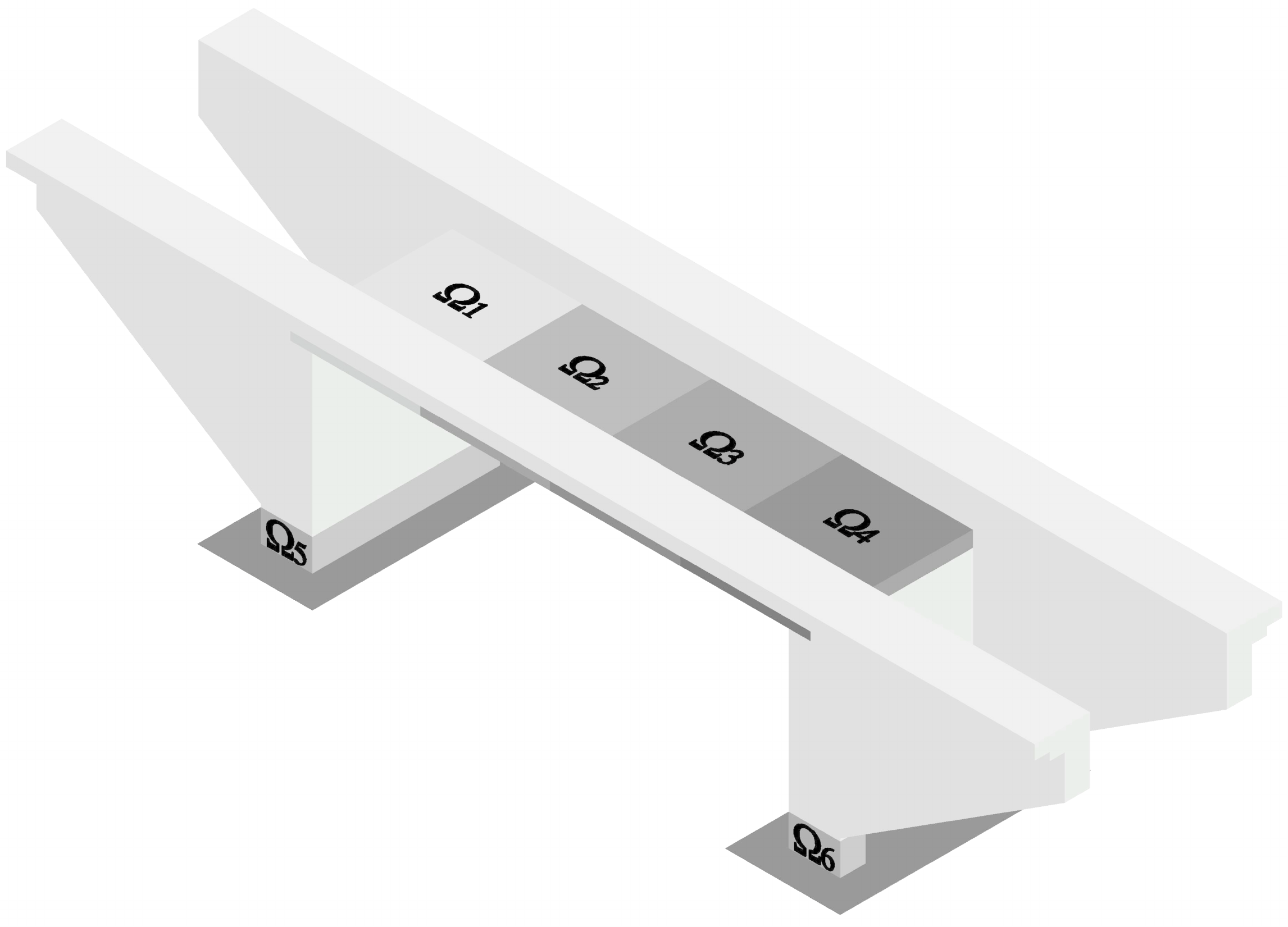}
\captionsetup{width=1\linewidth}
\caption{{Railway bridge: considered subdomains $\Omega_1,\ldots,\Omega_6$. The \textit{g}-th damage scenario refers to a localised stiffness reduction in the  \textit{g}-th subdomain.}\label{fig:fig35}}
\end{minipage} 
\hspace{0.2cm}
\begin{minipage}[b]{0.45\textwidth}
\centering
\includegraphics[width=\textwidth]{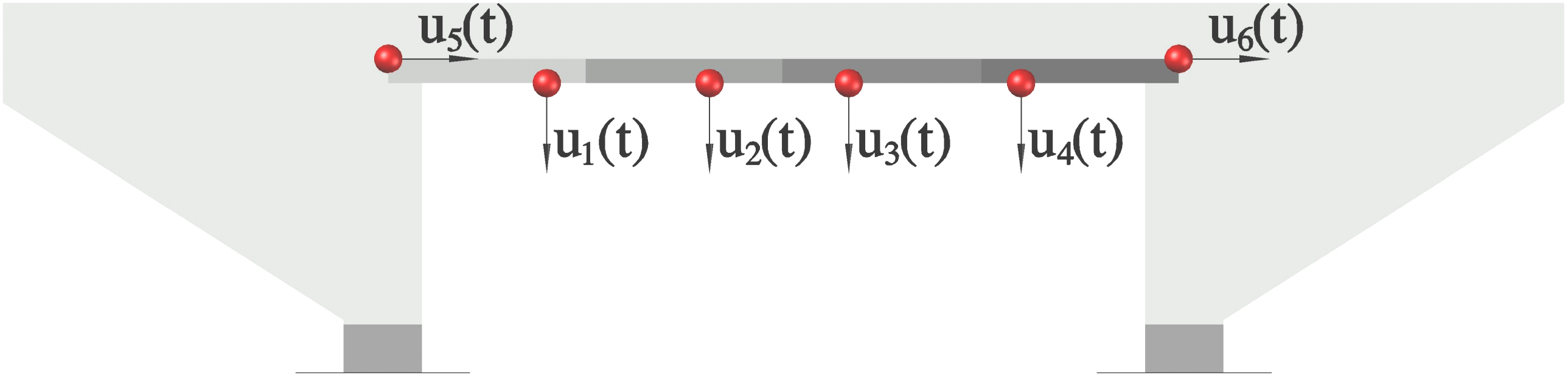}
\captionsetup{width=.8\linewidth}
\caption{{Railway bridge: monitoring system.}\label{fig:fig36}}
\end{minipage}
\end{figure}

Six damaged structural states $g\in\lbrace 1,2,3,4,5,6 \rbrace$,  schematically represented in Fig. \ref{fig:fig35}, have been considered further to the undamaged state $g=0$: the damaged states feature a stiffness reduction in the corresponding subdomain $\Omega_{1},\dots,\Omega_{6}$. Each time the mechanical problem has been solved, the damage scenario $g$ and the damage level $\delta$ have been respectively sampled via LHS from the uniform discrete pdf $\mathcal{U}_g\left(0,\dots,6\right)$ and from the uniform continuous  pdf $\mathcal{U}_{\delta}\left(5\%,25\%\right)$, respectively.
The extreme values of pdf $\mathcal{U}_{\delta}$ have been selected in order to assess if small damage events can be detected, distinguished from those characterized by a much larger reduction of the local mechanical properties, and also localized in real-life applications. This approach is obviously intended to work at the structural level, and disregard the features of microcracking patterns in the concrete structure, which are wholly measured through the adopted damage indices.  Such an approach was already adopted in the context of SHM is several studies; without any aim to provide an exhaustive account of the literature, readers are referred to, e.g. \cite{art:Farrar_I40_bridge,art:Pandey94} and also standards like \cite{code:EC2}. \\ 

The monitoring system has been assumed to be composed of $N_{0}=6$ sensors, placed as shown in Fig. \ref{fig:fig36}, and recording the vertical displacements of the deck and the horizontal displacements at the top of the frame walls. All the signals have been recorded with a sampling frequency of $400 \text{ Hz}$ within a monitoring window $T=1.5\text{ s}$, which allows the train to completely cross the bridge even if traveling at the lowest speed.

\smallskip

\smallskip

As for the train, the convoy velocity $\gamma$ and the mass $\beta$ carried by a single axle (or equivalently the load $F$ released by the single axle to the rails) have been modelled as random variables with uniform pdf $\mathcal{U}_{\gamma}\left(160,215\right)$ $\text{km}/\text{h}$ and $\mathcal{U}_{\beta}\left(16,22\right) \text{ ton}$. The load is transmitted from the rails to $25$ sleepers that cover the entire deck; the compressive distributed load  under the sleepers is then transmitted on its own to the ballast layer with a slope $4:1$ according to Eurocode 1 \cite{code:EC1}, so that the loaded surface amounts to $0.55 \text{ m} \times 2.1 \text{ m}$. The maximum compressive load value that the train can generate on the sleepers  is accordingly $p_{max}=\frac{F}{(0.55\cdot 2.1) \text{ m}^2}$; the moving load system is finally given as $P(t,\gamma,x)=\sum_{\xi}p^{\xi}(t,\gamma,x)$, with $\xi=1,\dots,25$, and \cite{code:EC1}
\begin{equation*}
p^{\xi}(t,\gamma,x)=\sum_{\alpha}p_{max}\cdot A^{\xi}(x)\cdot A^{\xi}_{\alpha}(t,\gamma) \quad \alpha=1,\dots,8.
\end{equation*}
where $A^{\xi}(x)$ is the space activation function, and $A^{\xi}_{\alpha}(t,\gamma)$ is the time modulation function related to the $\alpha$-th axle.  

The space activation function $A^{\xi}(x)$ accounts for the load in correspondence of the $\xi$-th sleeper, and is given by
\begin{equation*}
A^{\xi}(x)=H\left(x-\left(x_{\xi}-\frac{0.55\ m}{2}\right)\right)-H\left(x-\left(x_{\xi}+\frac{0.55\ m}{2}\right)\right),\phantom{-}\quad \xi=1,\dots,25 \quad ,
\label{eq:PO_1}
\end{equation*}
where $x_{\xi}$ is the abscissa of the center of gravity of the $\xi$-th sleeper, and $H(\cdot)$ is the Heavyside function.

\begin{figure}[h!]
\centering
\begin{minipage}[t]{0.48\textwidth}
\centering
    \includegraphics[width=0.3\textwidth]{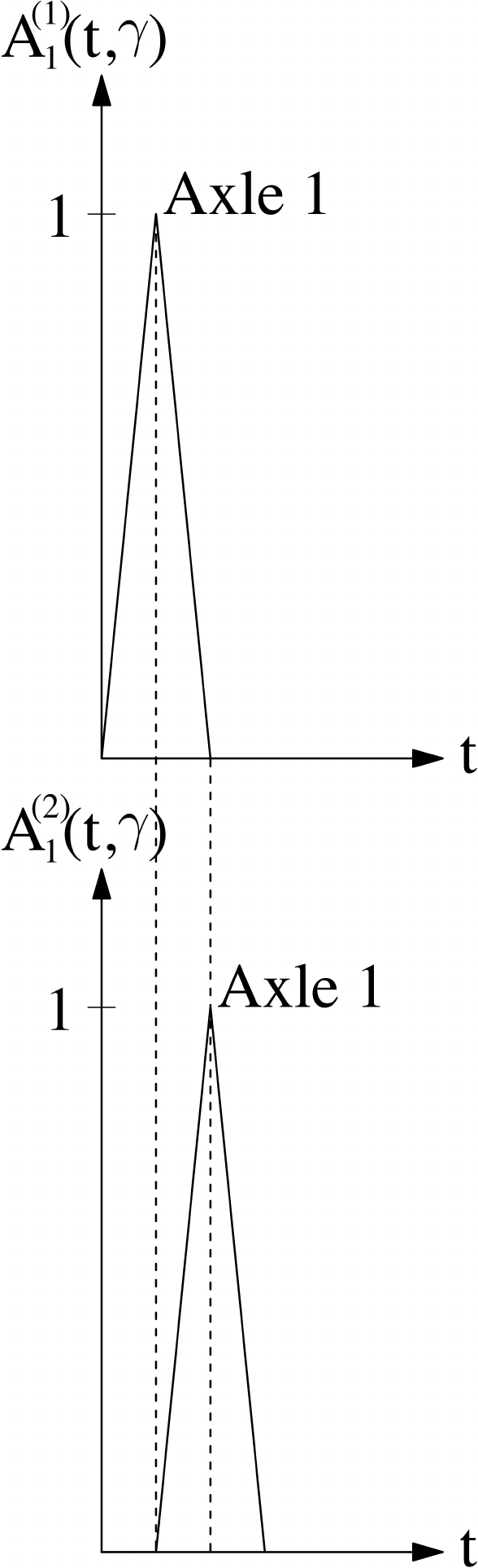}
  \captionsetup{width=.78\linewidth}
\caption{{Railway bridge: time modulation function. Example related to the $1^{\text{st}}$ axle on the $1^{\text{st}}$ and $2^{\text{nd}}$ sleepers.}\label{fig:fig38}}
\end{minipage} 
\hspace{0.2cm}
\begin{minipage}[t]{0.48\textwidth}
\centering
\includegraphics[width=\textwidth]{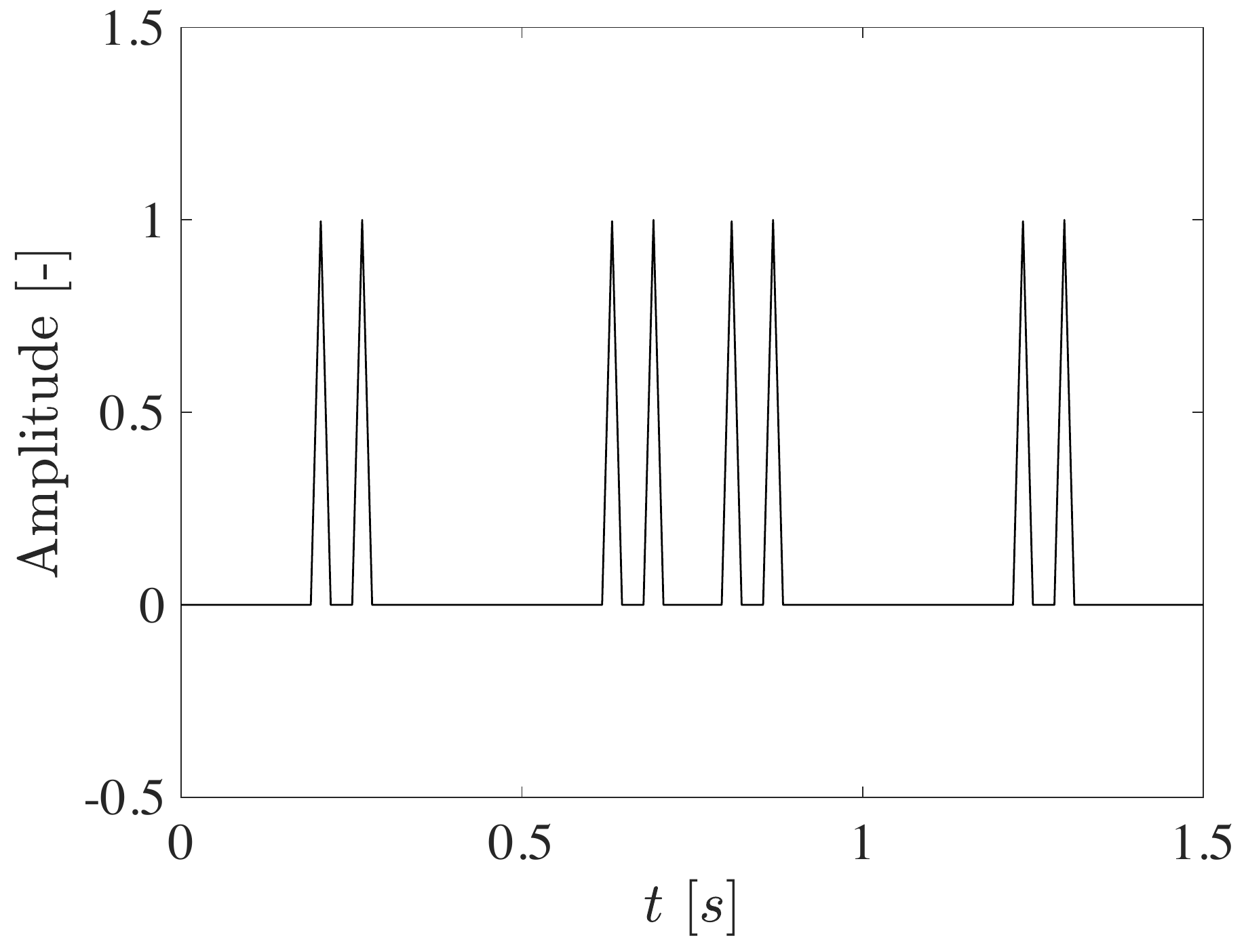}
\captionsetup{width=.95\linewidth}
\caption{{Railway bridge: time modulation function. 8 axles passage on the midspan sleeper, at the speed of $160 \text{ km}/\text{h}$.}\label{fig:fig39}}
\end{minipage}
\end{figure}

The time modulating function $A^{\xi}_{\alpha}(t,\gamma)$ allows instead to modulate the pressure value as a function of time and axle speed according to 
\begin{equation}
A^{\xi}_{\alpha}(t,\gamma)=\left[H\left(t-\frac{x_{\xi-1}+x^0_{\alpha}}{\gamma}\right)-H\left(t-\frac{x_{\xi+1}+x^0_{\alpha}}{\gamma}\right)\right]\cdot\left(1-\frac{\abs{t-\frac{x_{\xi}+x^0_{\alpha}}{\gamma}}}{\frac{0.65\ m}{\gamma}}\right), \ \alpha=1,\dots,8,
\label{eq:time_mod_funct}
\end{equation}
where $x^0_{\alpha}$ is the position of the $\alpha$-th axle at time $t=0$. Fig. \ref{fig:fig38} shows that the maximum value of the pressure on a sleeper is attained when the axle is crossing its axis, and it becomes null when the axle is crossing the axis of the previous or next sleeper, with a linear variation in between. In Fig. \ref{fig:fig39},  the time modulation function of the midspan sleeper for a train speed $\gamma=160 \text{ km}/\text{h}$ is  reported, to show the characteristic history of the external loading for this type of structural systems. 

\subsection{Railway bridge - ROM}
The simulations discussed in what follows have been run on a PC featuring an Intel (R) Core\texttrademark, i7-2600 CPU @ 3.4 GHz, with a 64 bit operating system and 16 GB RAM. The number of snapshots used to construct the ROM has been set to $S=35$. One could argue that collecting a larger number of snapshots would have further enhanced the representativeness of the ROM, but we have judged the employed number of snapshots as a good trade-off with the high computational cost of each FOM evaluation, as the snapshots collection must be carried out for the entire monitoring time window in order to fully catch the effects of moving loads. Each FOM simulation has required a computing time of about $7$ hours, and the ROM construction has required about $10$ days. 

\begin{figure}[h!]
\centering
\begin{minipage}[t]{0.48\textwidth}
\centering
\begin{tikzpicture}
  \node (POD_time_graph_1)  {\includegraphics[scale=0.35]{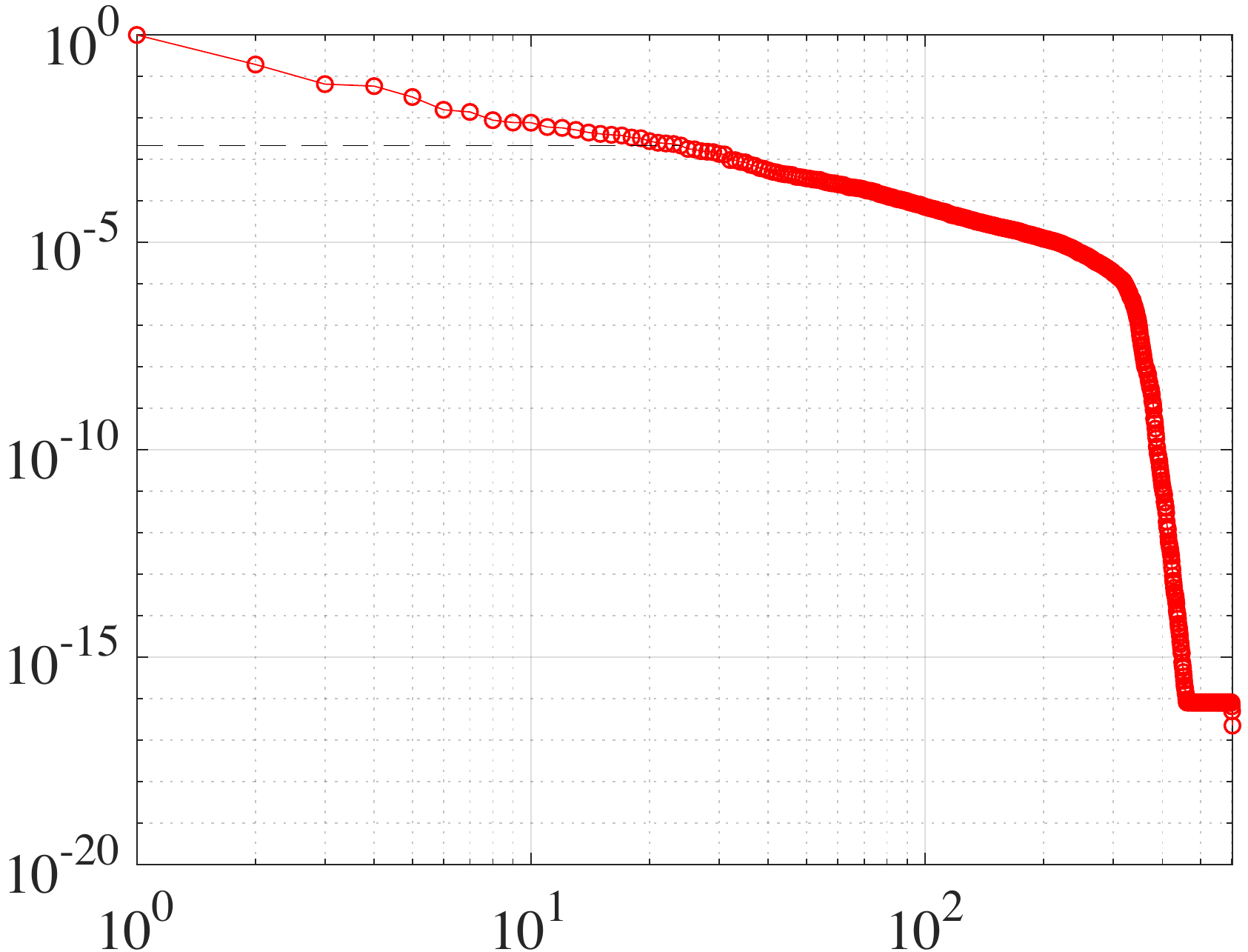}};
  \node[below=of POD_time_graph_1, node distance=0cm, yshift=1cm,font=\color{black}] {Number of snapshots};
  \node[left=of POD_time_graph_1, node distance=0cm, rotate=90, anchor=center,yshift=-0.9cm,font=\color{black}] {$\sigma^{T}_s / \sigma^{T}_1$};
 \end{tikzpicture}
\captionsetup{width=.9\linewidth}
\caption{{Railway bridge: POD in time. Descent of the singular values $\sigma^{T}_s$ normalised with respect to $\sigma^{T}_1$.}\label{fig:PODtimebridge}}
\end{minipage} 
\hspace{0.05cm}
\begin{minipage}[t]{0.48\textwidth}
\centering
\begin{tikzpicture}
  \node (POD_param_graph_1)  {\includegraphics[scale=0.35]{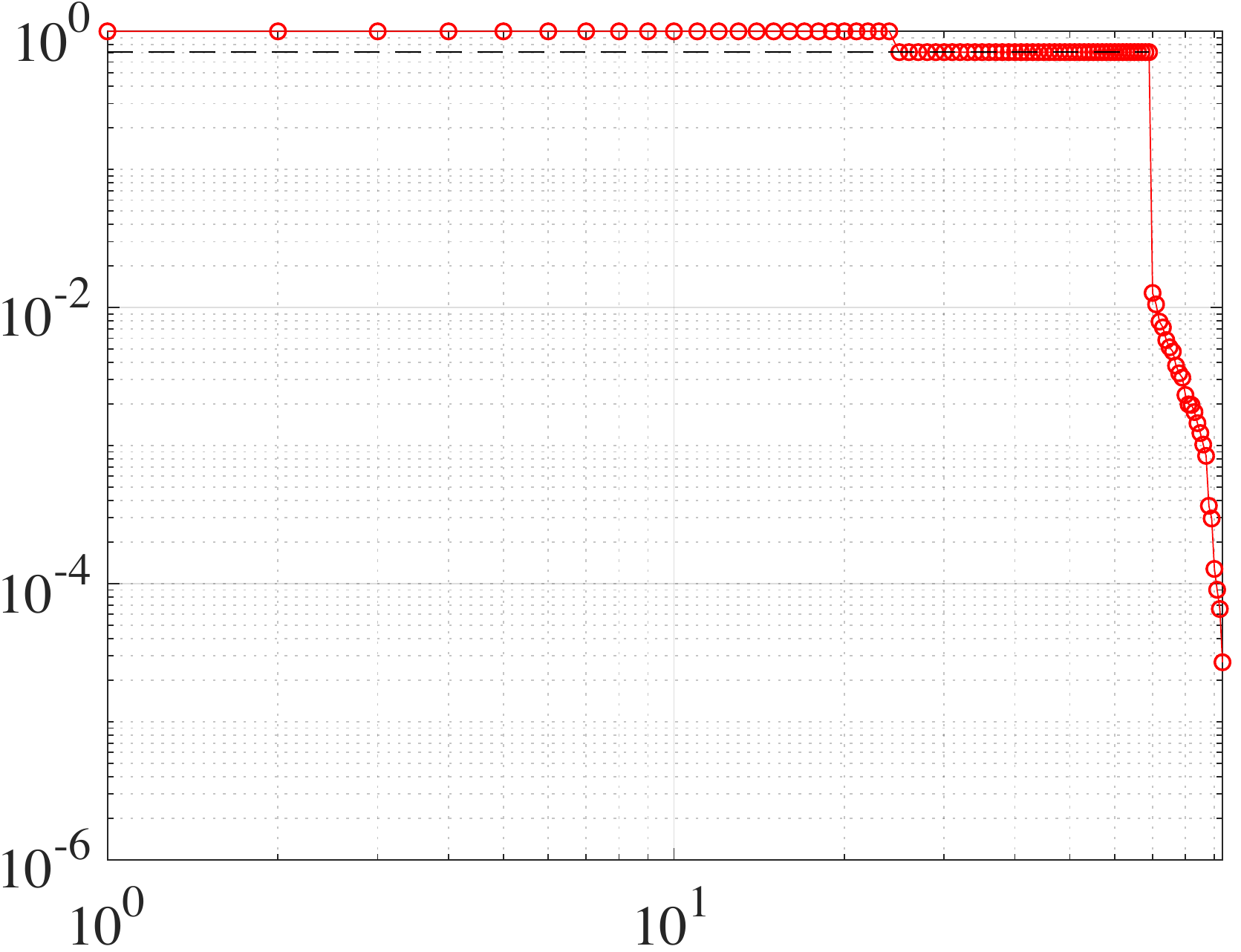}};
  \node[below=of POD_param_graph_1, node distance=0cm, yshift=1cm,font=\color{black}] {Number of snapshots};
  \node[left=of POD_param_graph_1, node distance=0cm, rotate=90, anchor=center,yshift=-0.9cm,font=\color{black}] {$\sigma^{g\boldsymbol{\eta}}_s /\sigma^{g\boldsymbol{\eta}}_1$};
 \end{tikzpicture}
\captionsetup{width=.9\linewidth}
\caption{{Railway bridge: POD over parameters. Descent of the singular values $\sigma^{g\boldsymbol{\eta}}_s$ normalised with respect to $\sigma^{g\boldsymbol{\eta}}_1$.}\label{fig:PODparambridge}}
\end{minipage}
\end{figure}

To attain high approximation capacity, the error tolerance has been set to $\varepsilon_{tol}=5\cdot10^{-3}$. In Figs. \ref{fig:PODtimebridge} and \ref{fig:PODparambridge}, the normalized singular values $\sigma^{T}_s/\sigma^{T}_1$ and $\sigma^{g\boldsymbol{\eta}}_s /\sigma^{g\boldsymbol{\eta}}_1$ are respectively shown for the POD in time and over the parametric space. Confirming what previously stated regarding the need to collect snapshots during the entire monitoring time window, in Fig. \ref{fig:PODtimebridge} the singular values in time are shown to decay only in the final part of the graph, so as the reconstruction error $\varepsilon$ does. 

\begin{figure}[h!]
\captionsetup[subfigure]{justification=centering}
%\begin{framed}
\centering
%\vspace{-0.4 cm}
\subfloat[[I POD basis.][\label{fig:fig49a}]{\includegraphics[scale=0.125]{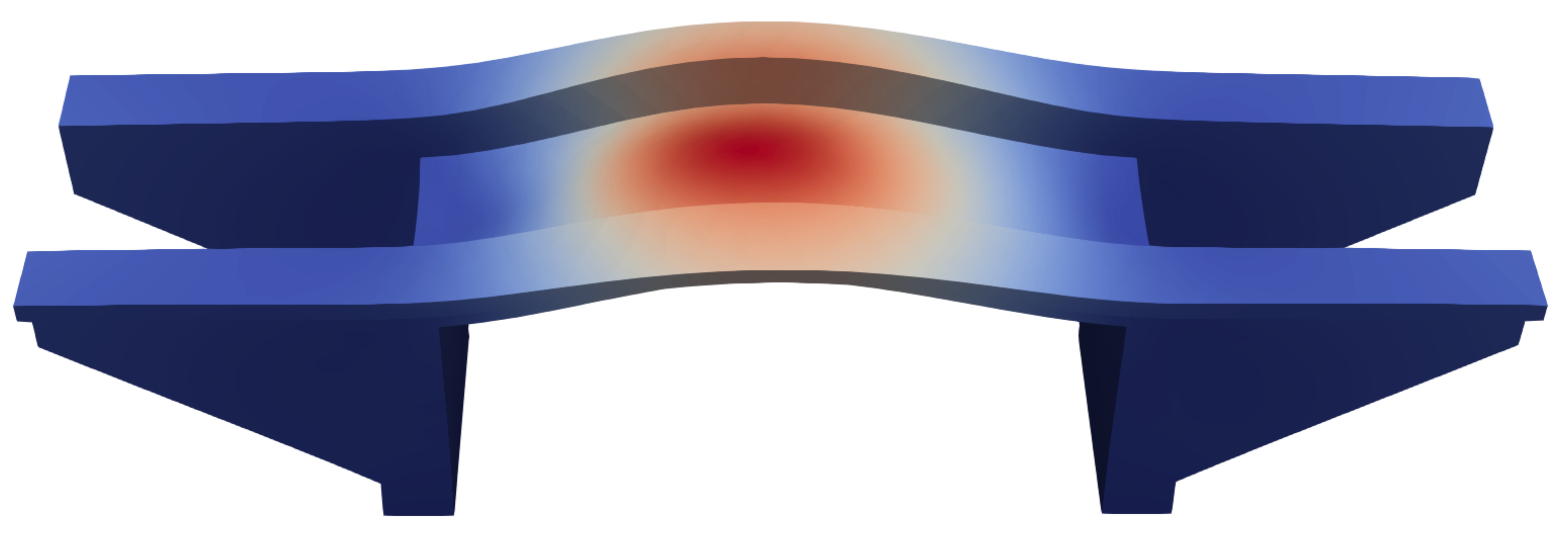}} $~$
\subfloat[[II POD basis.][\label{fig:fig50a}]{\includegraphics[scale=0.125]{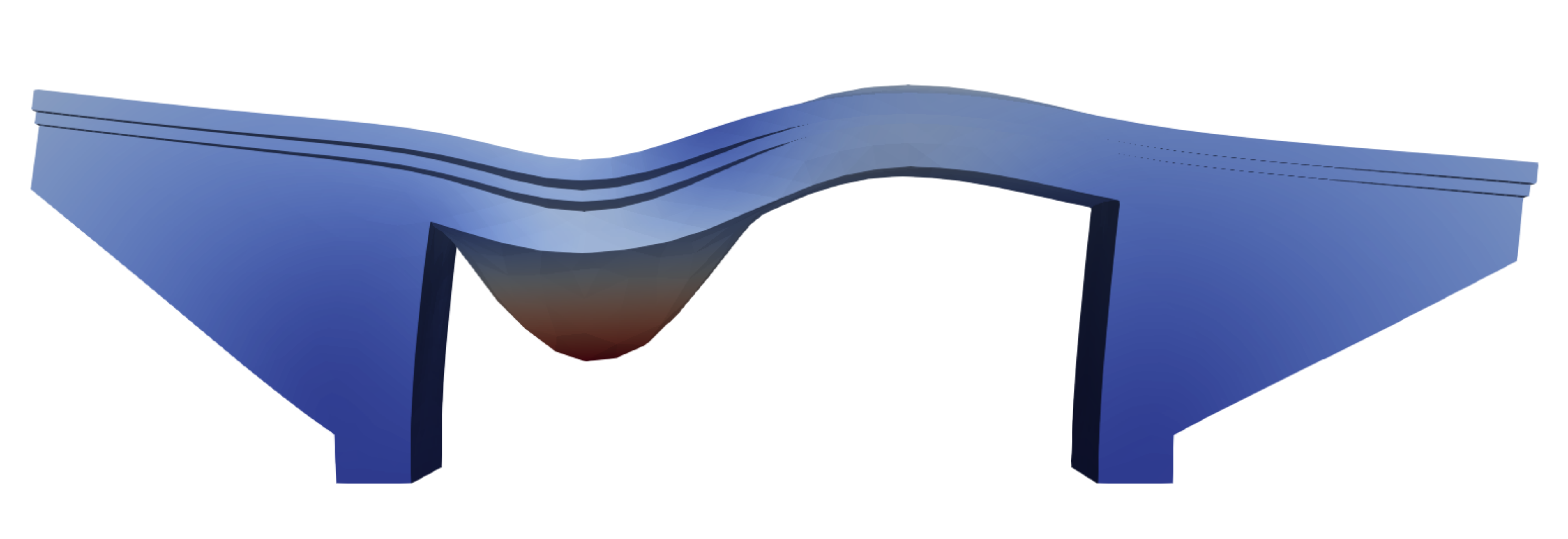}}\\
\subfloat[[III POD basis.][\label{fig:fig51}]{\includegraphics[scale=0.125]{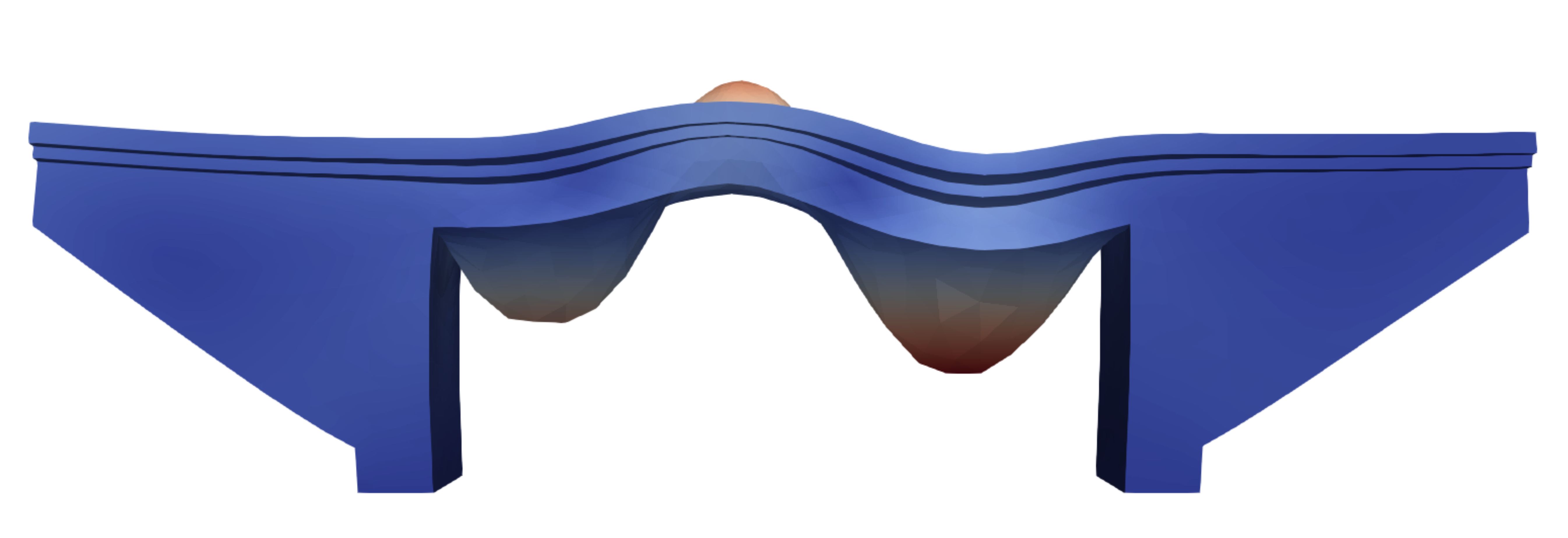}}$~$
\subfloat[[IV POD basis.][\label{fig:fig52}]{\includegraphics[scale=0.125]{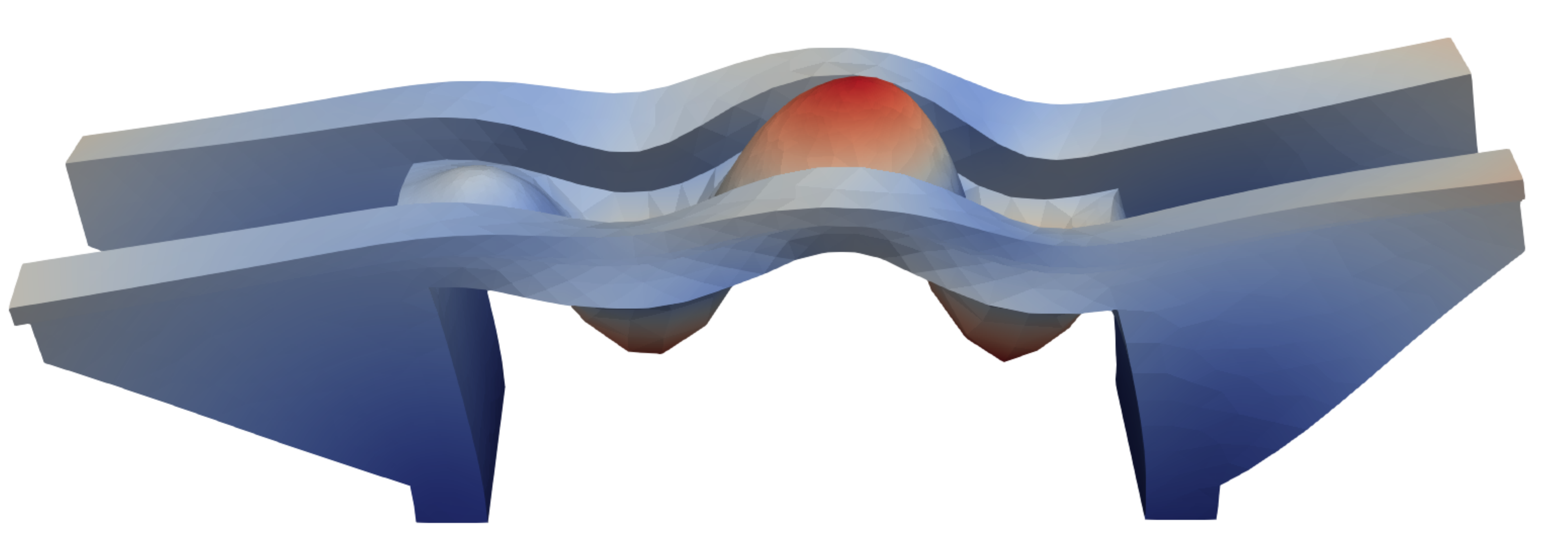}}\\
\caption{\small Railway bridge: POD bases.\label{fig:POD_basis_bridge}}
%\end{framed}
\end{figure}

\begin{figure}[h!]
\centering
\includegraphics[width=0.6\textwidth]{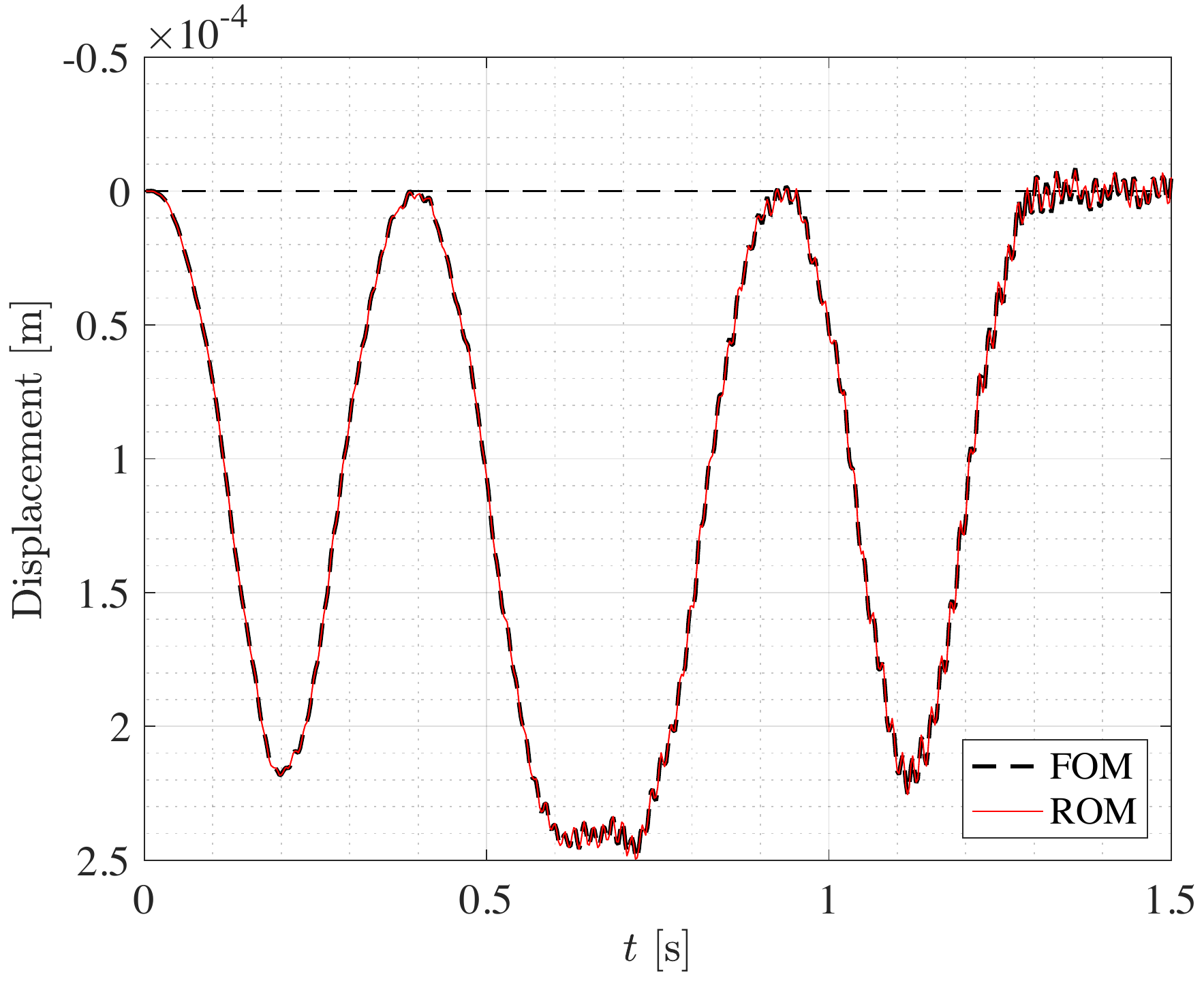}
\captionsetup{width=.75\linewidth}
\caption{{Railway bridge: comparison of the vertical displacement time histories at midspan obtained through the FOM and the ROM ($g=2$, $\delta=0.08$, $\gamma=180 \text{ km}/\text{h}$, $\beta=17.325 \text{ ton}$).}\label{fig:fig46}}
\end{figure}

The computational gain obtained with the use of the ROM is even more remarkable than in the previous case study, due to the higher computational complexity of this structure: we have moved from $M=15,300$ dofs of the FOM to only $W=69$ POD bases. The first 4 POD bases of the bridge are reported in Fig. \ref{fig:POD_basis_bridge}, to show how structural dynamics has been accounted for in the POD-based classification task. These POD bases appear very different if compared with the classical mode shapes; indeed, except for the first one, they do not present symmetries and, due to the peculiarity of the applied load, they are mainly active close to the sleepers area. The computational time required by each simulation has decreased from the aforementioned $7$ hours for a single FOM solution, to $60$ seconds for a ROM solution, with a speed-up of  $420$.
In terms of results, in Fig. \ref{fig:fig46} the vertical displacement at midspan is reported for $g=2$, $\delta=0.08$, $\gamma=180 \text{ km}/\text{h}$ and $\beta=17.325 \text{ ton}$, so as to assess the ROM accuracy: a noteworthy good approximation capacity is indeed achieved by the ROM, whose response is perfectly superposed to the FOM one. The enrichment in high frequency components of the vertical displacement along the time axis is due to the sequential passage of the axles over the sleepers, and to the absence of damping in the model.

\subsection{Railway bridge - Classification outcomes}

\begin{figure}[h!]
\centering
\begin{minipage}[c]{0.48\textwidth}
\centering
\includegraphics[width=1.05\textwidth]{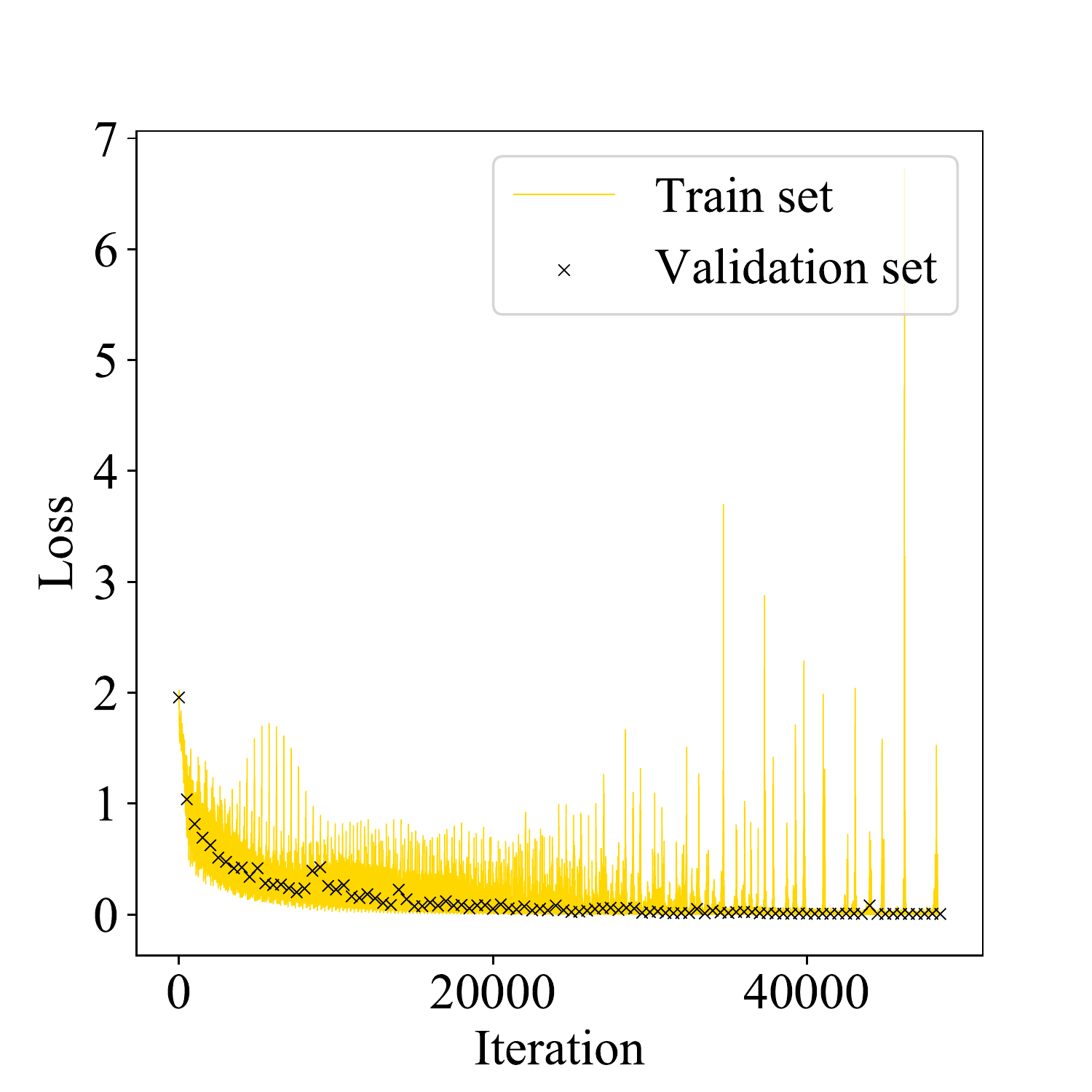}
\captionsetup{width=.9\linewidth}
\caption{{Railway bridge: FCN training. Loss function evolution on the training and validation sets.}\label{fig:bridge_loss_6_gdl}}
\end{minipage} 
\hspace{0.2cm}
\begin{minipage}[c]{0.48\textwidth}
\centering
\includegraphics[width=1.05\textwidth]{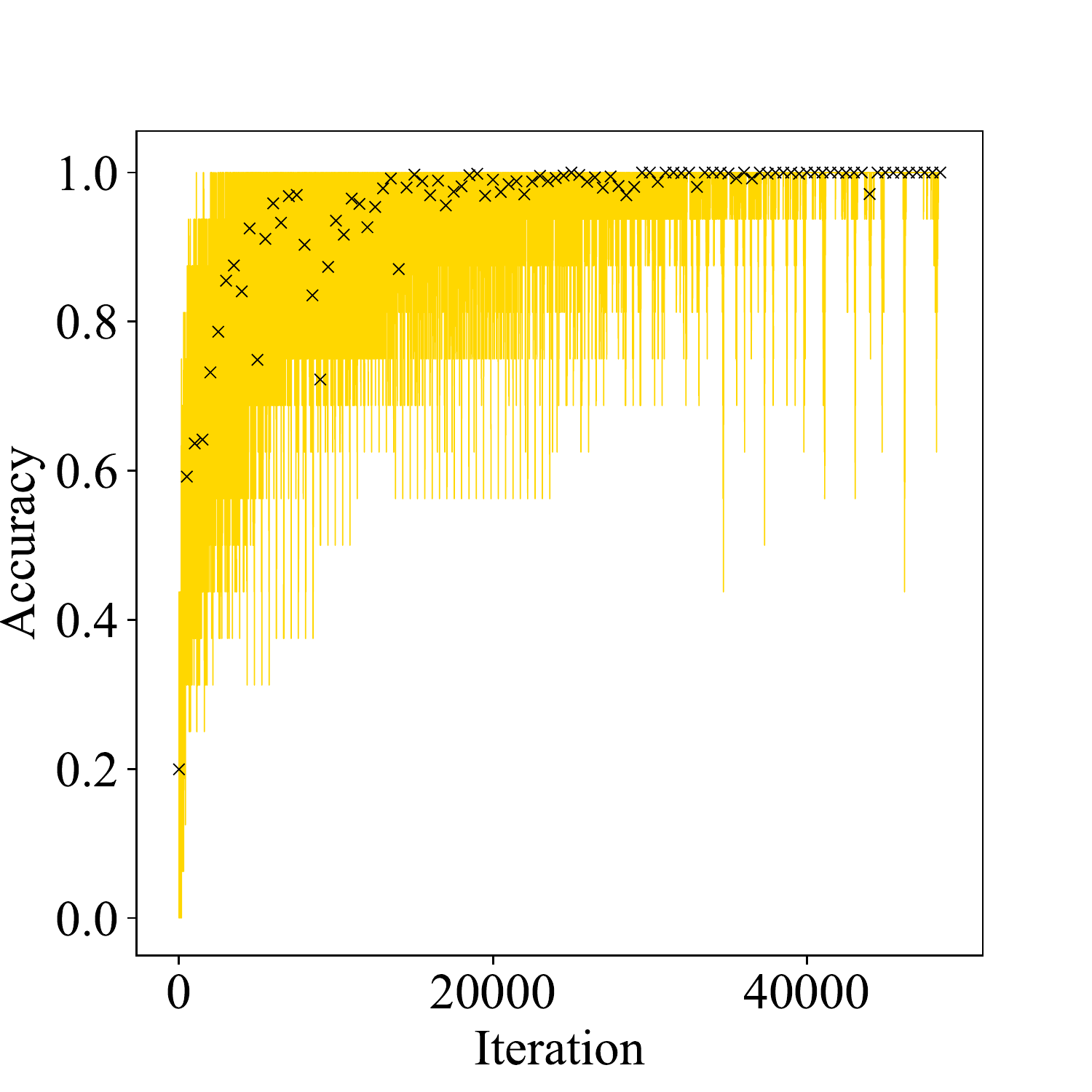}
\captionsetup{width=.9\linewidth}
\caption{{Railway bridge: FCN training. Global accuracy evolution on the training and validation sets.}\label{fig:bridge_acc_6_gdl}}
\end{minipage}
\end{figure}

The classifier $\mathcal{G}$ has been trained and validated on $I_{tr}+I_{val}=10,000$ instances, with a ratio $75:25$, for $1000$ epochs. In Figs. \ref{fig:bridge_loss_6_gdl} and \ref{fig:bridge_acc_6_gdl} the evolutions of the loss and of the accuracy functions during training are shown: a classification accuracy of $100\%$ has been  attained on both the training and validation sets.
All the damage scenarios have been therefore perfectly recognized and classified. \\ 

\begin{figure}[h!]
\centering
\begin{minipage}[c]{0.48\textwidth}
\centering
\includegraphics[width=1.05\textwidth]{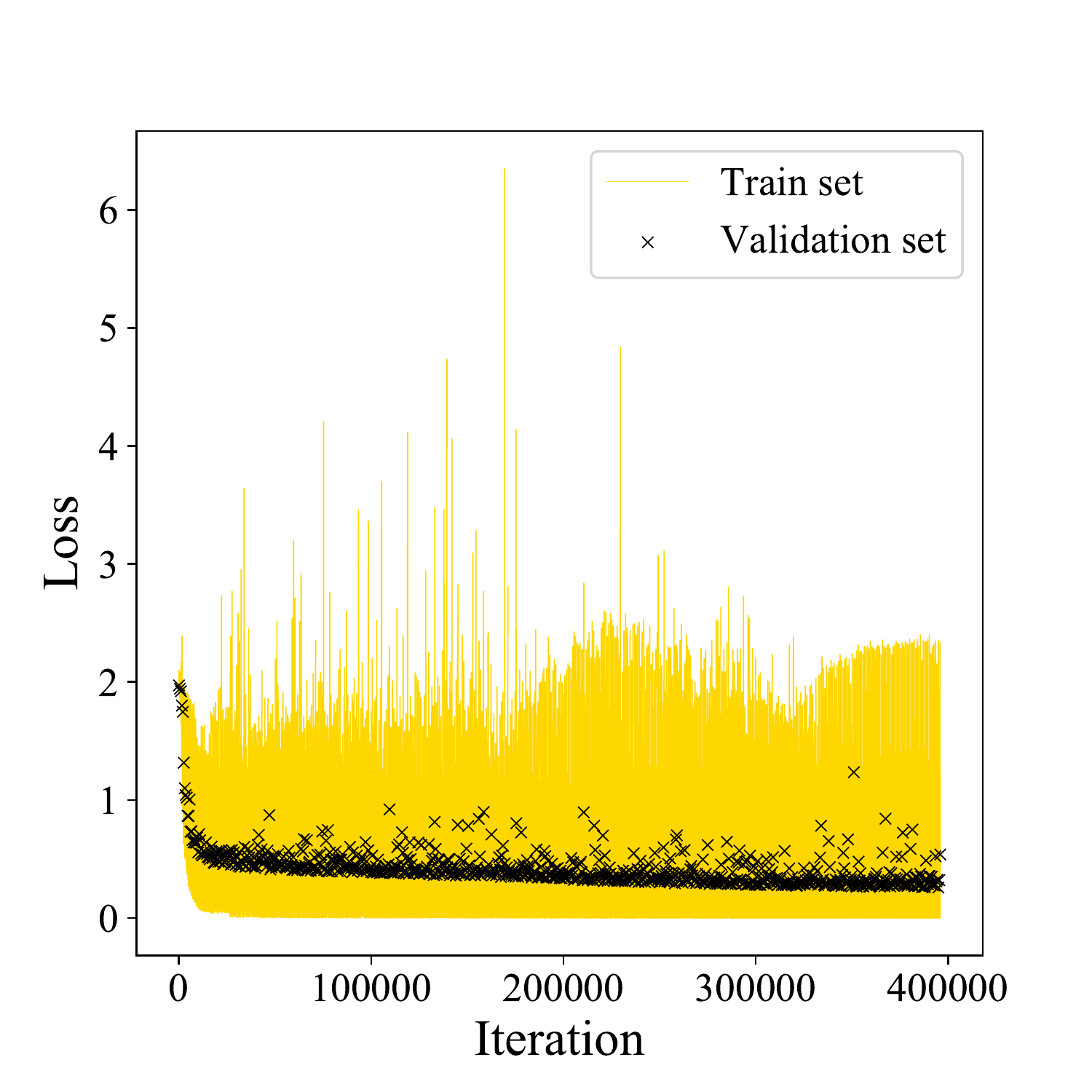}
\captionsetup{width=.9\linewidth}
\caption{{Railway bridge, four sensor monitoring system: FCN training. Loss function evolution on the training and validation sets.}\label{fig:fig47}}
\end{minipage} 
\hspace{0.2cm}
\begin{minipage}[c]{0.48\textwidth}
\centering
\includegraphics[width=1.05\textwidth]{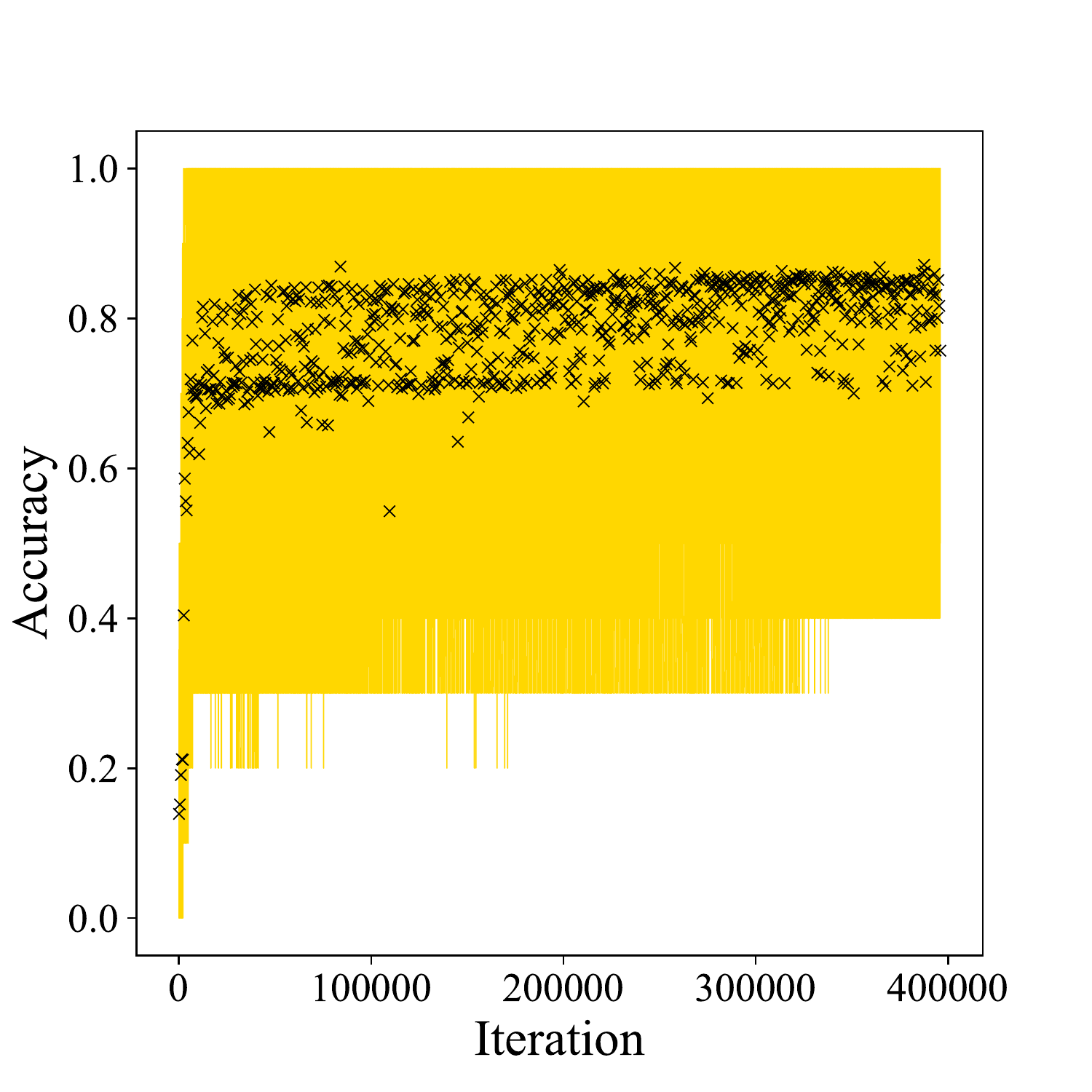}
\captionsetup{width=.9\linewidth}
\caption{{Railway bridge, four sensor monitoring system: FCN training. Global accuracy evolution on the training and validation sets.}\label{fig:fig48}}
\end{minipage}
\end{figure}

\begin{figure}[h!]
\centering
\begin{minipage}[t]{0.48\textwidth}
\centering
\includegraphics[width=\textwidth]{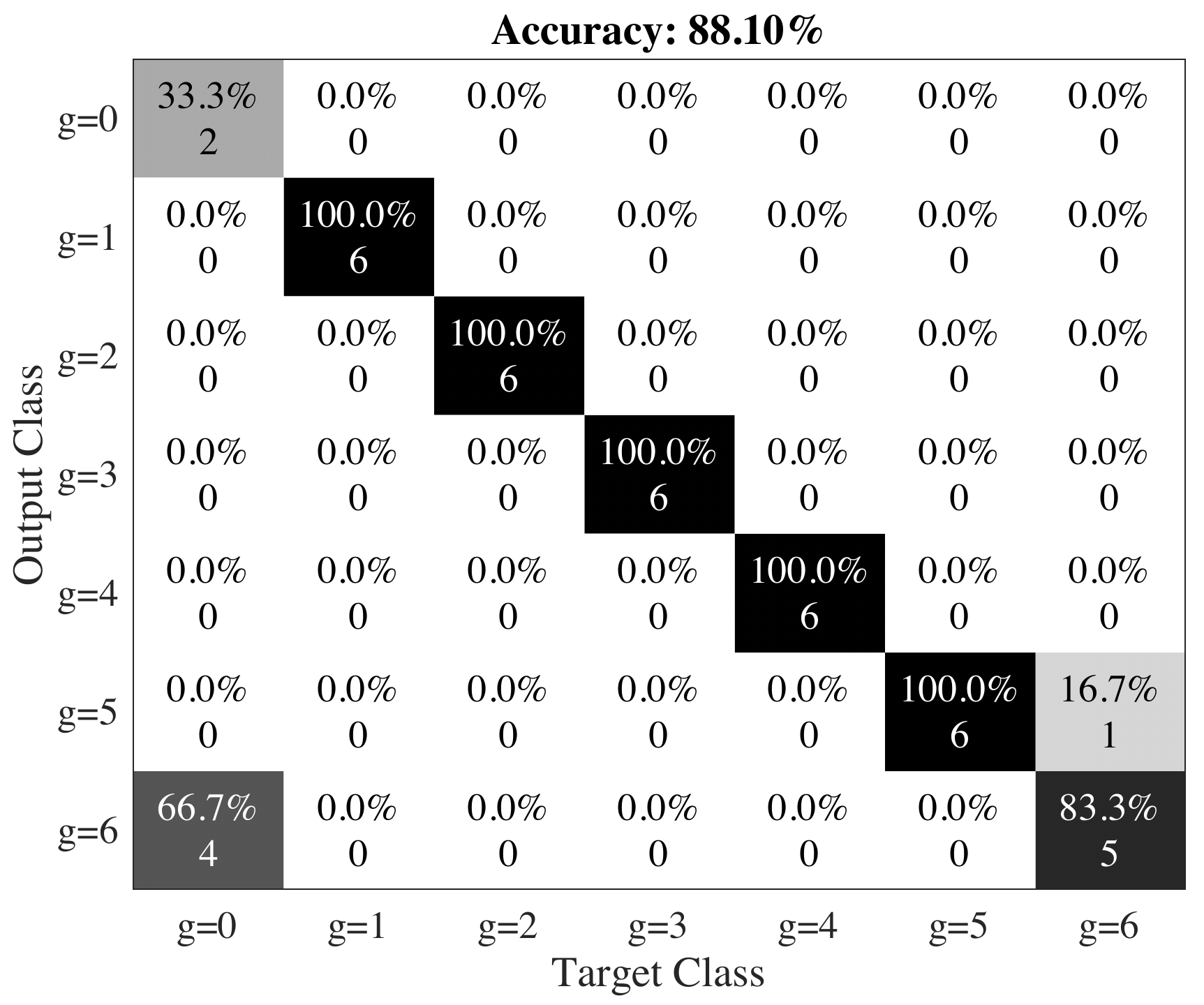}
\captionsetup{width=.9\linewidth}
\caption{{Railway bridge, four sensor monitoring system: FCN testing. Confusion matrix.}\label{fig:fig49}}
\end{minipage} 
\hspace{0.2cm}
\begin{minipage}[t]{0.48\textwidth}
\centering
\includegraphics[width=\textwidth]{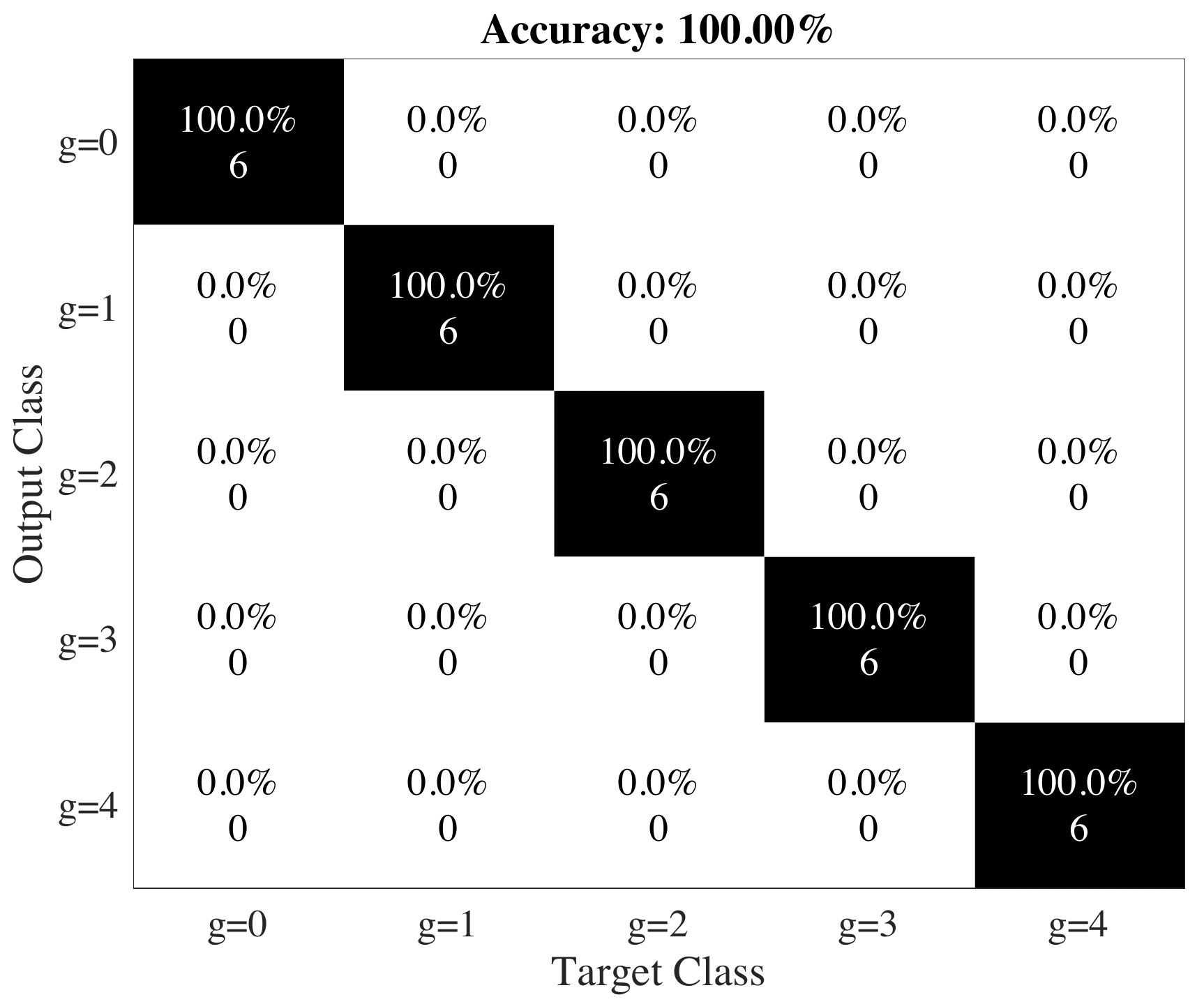}
\captionsetup{width=.9\linewidth}
\caption{{Railway bridge, four sensor monitoring system: FCN testing. Confusion matrix with damage scenarios involving only the deck.}\label{fig:fig50}}
\end{minipage}
\end{figure}

To show the paramount importance of an appropriate deployment of the sensors to measure the structural response to the external loading, a further analysis has been run by neglecting the horizontal recordings  $u_5$ and $u_6$ in Fig. \ref{fig:fig36}, hence with only information relevant to the $u_1-u_4$ time series processed by the classifier.
The relevant loss and accuracy evolutions shown in Figs. \ref{fig:fig47} and \ref{fig:fig48} highlight that $\mathcal{G}$ is not able now to  recognize all the processed instances correctly. The presence of a systematic classification error is testified also by the confusion matrix in Fig. \ref{fig:fig49}, in which the generalization capabilities of $\mathcal{G}$ are evaluated against a test set consisting of 42 pseudo-experimental instances simulated with the FOM: a global accuracy of only $88.10\%$ has been obtained. 
It has emerged that the undamaged scenario is prone to be misclassified as a structural state featuring $g=6$.
Such misclassification between scenarios $g=0$ and $g=6$ is thus due to the missed horizontal dofs  in the monitoring system. To further prove this claim, additional tests have been performed on a reduced dataset, characterized by removing scenarios $g=5,6$; results in Fig. \ref{fig:fig50} show how the performance of the classifier returns back to feature  a $100\%$ accuracy. The reported overall performances attained by $\mathcal{G}$ are considered good, especially in view of the high complexity of this  example resembling a real monitoring problem. 
  
%%%%%%%%%%%%%%%%%%%%%%%%%%%%%%%%%%%%%%%%%%%%%%%%%%%%%%%%%%
%bisogna richiamare e commentare le seguenti figure relative al: sistema di sensori con sei accelerometri; loss e accuracy nel caso di utilizzo di sei sensori

%%%%%%%%%%%%%%%%%%%%%%%%%%%%%%%%%%%%%%%%%%%%%%%%%%%%%%%%%%

\section{Conclusion}
\label{sec:conclusion}

In this work, we have proposed a neural network-based classifier, featuring a fully convolutional network architecture, to move towards online damage localization within a smart structural health monitoring framework. The classifier processes the vibration measurements, recorded by a sensor network deployed over the  structure, to identify the current structural state. To overcome the lack of experimental data for civil applications, we have exploited physics-based numerical modeling in order to build offline a large training set of structural responses,  accounting for relevant damage scenarios and operational conditions. A parametric model order reduction technique has been next adopted to replace high fidelity, time consuming finite element simulations and  speedup the dataset generation. The classifier leverages on the convolutional layers capabilities to automatically extract useful, damage sensitive features from raw data and  learn the functional link between such features and the corresponding structural states. The  obtained results have confirmed the high potential of the simulation based classification approach to structural health monitoring and of the combined use of parametric model order reduction techniques and deep learning. 

In both of the proposed case studies, the global accuracy of classification never falls below $85\%$ , regardless of whether acceleration or displacement measurements are handled. Indeed, the method has proven to be extremely robust in exploring a large parametric dependency and even in recognizing damaged scenarios significantly different from those observed during the training phase.

Tests have been carried out by adding to the response of the considered digital twins a white noise corruption of varying amplitude, which has been assumed representative of micro-electro mechanical system accelerometer self-noise, and by exploiting different reduced order models of increasing fidelity to build the training datasets. The classification outcomes have shown a slightly decreased global accuracy, featuring a minimum of $78 \%$ in the presence of a highly noisy signal (SNR$=20$), testifying that the procedure is also rather noise tolerant. Results obtained by exploiting reduced order models generated with a varying value of the error tolerance, have also provided a scheme to assess the effect of the reduced order modeling technique on the classifier performances.

In future works, varying environmental conditions and different excitation sources, such as wind action and low intensity seismicity will be allowed for, by further enlarging the parametric space exploited in the dataset construction. To handle the resulting nonaffine  dependency of the numerical arrays on the parametric space, hyper-reduction techniques are going to be exploited. To cope with the need of an optimal sensor placement, we aim to introduce a  sensor placement approach to maximize the information effectiveness for the classification task. Further examples are currently under study, in order to  validate the offered methodology against suitable experimental settings.
 
\Urlmuskip=0mu plus 1mu\relax
\bibliography{biblio}
%\biboptions{sort&compress}
\bibliographystyle{ieeetr}

\end{document}